\documentclass[pdflatex,sn-mathphys]{sn-jnl}

\jyear{2026}

\usepackage{amsmath,amssymb,amsfonts}
\usepackage{enumitem}
\usepackage{graphicx}
\usepackage{longtable}
\usepackage{booktabs}
\usepackage{array}

\newcommand{\runinhead}[1]{\par\medskip\noindent\textbf{#1}\enspace}
\makeatletter
\newcommand{\corrauth}[1]{\g@addto@macro\@equalconttext{\par *#1}}
\newcommand{\authornoteline}[1]{\g@addto@macro\@equalconttext{\par\vspace{12pt} #1}}
\let\gento@printabstract\printabstract
\renewcommand{\printabstract}{\clearpage\gento@printabstract}
\makeatother

\begin{document}

\title[Unified generative design of architected metamaterials]{Steering topology distributions for unified generative design of architected metamaterials}

\author[1,3]{\fnm{Haolin} \sur{Li}}
\equalcont{These authors contributed equally to this work.}

\author[2,8]{\fnm{Yuyang} \sur{Miao}}
\equalcont{These authors contributed equally to this work.}

\author[4]{\fnm{Menglei} \sur{Li}}

\author[5]{\fnm{Jinshuai} \sur{Bai}}

\author*[2]{\fnm{Liyuan} \sur{Wang}}

\author[4]{\fnm{Xin} \sur{Liu}}

\author[9]{\fnm{Bo} \sur{Gao}}

\author[6]{\fnm{Jiantao} \sur{Liu}}

\author[8]{\fnm{Danilo} \sur{Mandic}}

\author[1]{\fnm{Zahra} \sur{Sharif Khodaei}}

\author[1]{\fnm{M. H.} \sur{Aliabadi}}

\author*[3,7]{\fnm{Weiqiu} \sur{Chen}}

\affil[1]{Department of Aeronautics, Imperial College London, London, UK}
\affil[2]{Department of Psychological and Cognitive Sciences, Tsinghua University, Beijing, China}
\affil[3]{Department of Engineering Mechanics, Zhejiang University, Hangzhou, China}
\affil[4]{National Key Laboratory of Science and Technology on Advanced Composites in Special Environments, Harbin Institute of Technology, Harbin, China}
\affil[5]{Institute of Biomechanics and Medical Engineering, Department of Engineering Mechanics, Tsinghua University, Beijing, China}
\affil[6]{School of Mechanical Engineering, Southwest Jiaotong University, Chengdu, China}
\affil[7]{Faculty of Mechanical Engineering and Mechanics, Ningbo University, Ningbo, China}
\affil[8]{Department of Electrical and Electronic Engineering, Imperial College London, London, UK}
\affil[9]{Electrical and Computer Engineering, Carnegie Mellon University, Pittsburgh, USA}
\corrauth{Corresponding authors.}

\authornoteline{liyuanwang@tsinghua.edu.cn; chenwq@zju.edu.cn}

\abstract{
Architected metamaterials derive their functions from structure, creating vast opportunities to program physical responses through topology design. However, existing design methods are often tailored to individual design problems, making limited use of topology knowledge for effective and broadly applicable design as objectives, constraints, and physical functions change. Here we introduce Generative Topology Optimization (GenTO), a unified framework that turns a learned topology prior into a reusable design engine. GenTO trains a diffusion model on a large full-order topology dataset and then iteratively steers the resulting topology distribution toward task-specific high-performing regions using user-defined physical objectives and constraints. This shifts the object of optimization from a single structure to a task-adapted topology distribution. Across topology design problems spanning thermal extremization, multi-objective morphology control, property-targeted auxetic design, and vibration transmission design, GenTO reuses pretrained topology priors for heterogeneous tasks, preserves structural diversity, and reaches high-performing solutions supported by numerical benchmarks and experimental validation. These results establish reusable topology knowledge as a unified principle for effective and scalable architected metamaterial design.
}


\maketitle

\section{Introduction}

Architected metamaterials are engineered materials whose effective properties are determined primarily by designed internal architecture \cite{kadic20193d,berger2017mechanical,zhao2025modular,fang2025large}. The structural origin of function makes them a powerful platform for programming thermal, mechanical, acoustic, and other physical responses, including behaviors that are difficult to realize in conventional natural materials. In such systems, the design of internal connectivity and morphology directly governs how architecture gives rise to emergent structural behavior \cite{bertoldi2017flexible,xin2020topological,liu2018topological}. Knowledge of how structural organization shapes function, referred to here as \textbf{topology knowledge}, therefore constitutes a critical basis for metamaterial design. As objectives, constraints, and physical functions continue to diversify, a central challenge is to identify high-performing structures for individual tasks while making topology knowledge reusable across different design problems.

A prevalent design paradigm revolves around the optimization of individual structures. Established computational approaches, originally developed for structural optimization, have been widely adapted to metamaterial design \cite{yu2018mechanical,zheng2021topology,diaz2010topology,li2019topology}. Whether through classical topology optimization and its variants or through gradient-free black-box search, these methods can in principle be redeployed to new design problems once objectives and constraints are specified~\cite{sigmund200199,jeong2025advanced,jeong2023complete,jeong2023physics,dong2022structural,cerniauskas2024machine,wang2021parameter}. However, each new problem still requires searching over candidate structures \textbf{from scratch}, a process that often becomes difficult, inefficient, or unstable when the design landscape is highly non-convex, the objective is discrete or non-differentiable, or evaluation relies on expensive black-box simulations or experiments \cite{ji2026designing,fotopoulos2024review,chen2025nonlocal,zangeneh2021analogue,zhang2023diffusion}. These methods therefore remain adaptable in problem formulation, yet they do not convert accumulated design experience into a reusable structural prior.

Generative design offers another complementary route by learning statistical regularities in topology space from large topology datasets, thereby providing priors over admissible structures and showing promise across a range of inverse-design settings \cite{mao2020designing,zheng2021controllable,challapalli2021inverse,chiang2023generating,zheng2023unifying,bastek2023inverse,liu2024generative,kim2024arithmetic,ma2019probabilistic,wang2020deep,du2026topogenmeta,yang2026guided,dedoncker2025generative,apaza2024transfer,abumullla2025curved,wang2026bicontinuous,xie2026physics}. Recent studies further treat learned generators as both fixed conditional samplers and models that can be transferred or adapted to new tasks \cite{apaza2024transfer,yang2026guided,xue2025mind}, pointing to topology design around a learned distribution as a promising direction. Yet most generative design methods remain tied to prescribed property targets, fixed conditional interfaces, or reduced design spaces, leaving the learned topology prior \textbf{rarely reusable} across heterogeneous design problems. The field therefore still lacks a general framework in which learned topology knowledge can be retained, repurposed, and repeatedly adapted as design tasks change.

\begin{figure}[htbp!]
    \centering
    \includegraphics[width=0.99\textwidth]{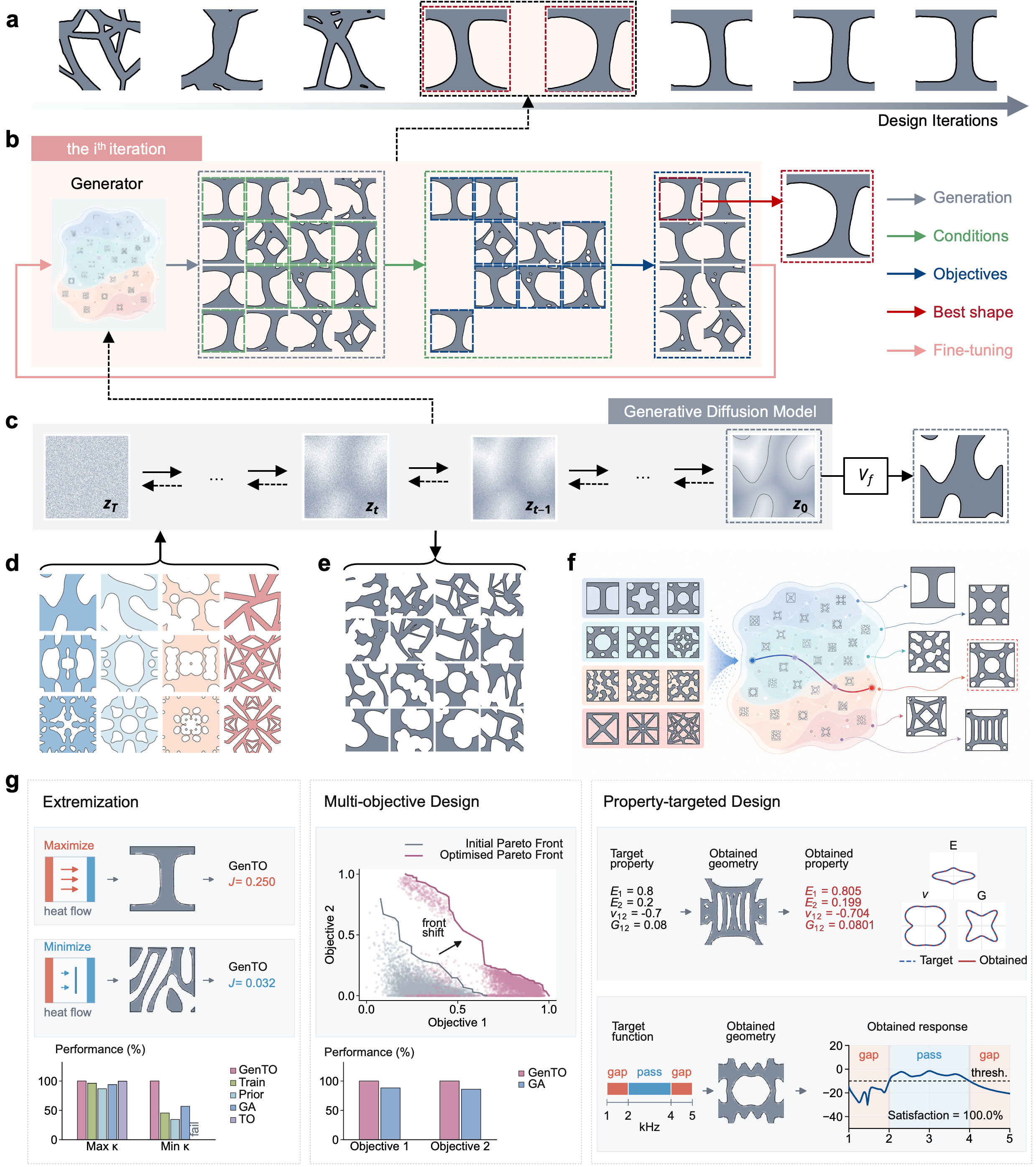}
\caption{\textbf{Overview of the GenTO framework.}
\textbf{a}, Illustration of the iterative topology design process, in which generated structures progressively evolve over design iterations toward a high-performing topology.
\textbf{b}, Schematic of the GenTO workflow. A pretrained generative model proposes candidate topologies, which are filtered by user-defined conditions and objectives; the selected best shape is then used for fine-tuning, closing the iterative loop (Supplementary Fig.~\ref{fig:SI_F1}).
\textbf{c}, Progressive generation by the underlying diffusion prior. Starting from Gaussian noise, the model denoises through intermediate states to a clean signed-distance-function (SDF) field $\mathbf{z}$, from which a binary topology at prescribed volume fraction $V_f$ is extracted (Supplementary Fig.~\ref{fig:SI_F3}).
\textbf{d}, Training dataset composed of four topology families: Gaussian Random Fields (GRF), Cahn-Hilliard Phase Fields (CHPF), Porous Structures (PR), and Truss-based Structures (TS). Each family is generated under three symmetry groups: Periodic, 4-fold, and 8-fold (Supplementary Fig.~\ref{fig:SI_F2}).
\textbf{e}, Representative samples generated from the learned joint prior, illustrating the diversity of the reusable topology distribution.
\textbf{f}, Learned topology prior obtained from the four training families. The pretrained prior is a joint distribution in which family information can mix, rather than a simple collection of separated family distributions (Methods Section~\ref{sec:methods_distribution_steering}).
\textbf{g}, Unified design capabilities enabled by GenTO. Starting from the same reusable topology prior, the same distribution-steering framework can support extremization, multi-objective optimization, property-targeted design, and function-targeted design, with the four task cards illustrating representative uses in the case studies.
}
    \label{fig:overview}
\end{figure}

In this work, we propose a unified framework for generative design of architected metamaterials, namely Generative Topology Optimization (GenTO), that turns a learned topology prior into a reusable design engine (Fig.~\ref{fig:overview}). Instead of building another task-specific conditional generator, GenTO formulates topology design as iterative steering of a pretrained topology distribution. Starting from a generative model trained on a large full-order topology dataset, GenTO progressively steers the learned prior toward high-performing regions using user-defined physical objectives and constraints through an iterative loop of candidate generation, evaluation, selection, and fine-tuning (Methods and Supplementary Section~\ref{sec:si_method_details}). The object of optimization thus shifts from a single structure to a task-adapted topology distribution. The shift decouples reusable topology knowledge from task-specific design objectives and makes it the central substrate of the design process.

We validate this formulation in topology design problems spanning thermal extremization, multi-objective morphology control, property-targeted auxetic design, and vibration transmission design. Across these diverse settings, GenTO reuses task-matched pretrained topology priors to support single-objective, multi-objective, property-targeted, and function-targeted design within one framework (Supplementary Sections~\ref{sec:si_case_details} and \ref{sec:si_additional_results}). Because optimization proceeds in distribution space instead of along a single design trajectory, GenTO preserves structural diversity, remains effective under markedly different objectives and constraints, and in several cases pushes generation beyond the support of the initial pretrained distribution, yielding high-performing out-of-distribution designs. These results establish reusable topology knowledge as a general design principle for architected metamaterials and position distribution steering as a route toward more transferable generative design.

\section{Results}
\label{sec:results}

GenTO provides a reusable design framework that adapts a learned topology prior to new objectives and constraints, rather than restarting the search from scratch for each problem. Instead of directly optimizing one topology at a time, \textbf{GenTO optimizes how topologies are sampled}. It does so by changing the generator that produces them. In each design iteration, the model generates candidate structures, keeps those that satisfy the user-defined conditions and objectives, and uses these selected structures to fine-tune the model before the next round (Fig.~\ref{fig:overview}a,b; Methods and Supplementary Section~\ref{sec:si_learning_steering}). The generator is a generative diffusion model (Fig.~\ref{fig:overview}c) that produces candidate topologies (Fig.~\ref{fig:overview}e). Before any task-specific update, the generator is pretrained on multiple topology families (Fig.~\ref{fig:overview}d, Supplementary Section~\ref{sec:si_datasets}), giving the prior $p_0$ the ability to produce a broad range of structures (Fig.~\ref{fig:overview}d,e). Repeated generation, selection, and fine-tuning convert this broad prior into a narrower task-adapted distribution. In compact form, all case studies can be written as the following task-dependent optimization problem:
\begin{equation}
\begin{aligned}
\mathbf{x}^{\star}
&\in
\operatorname*{arg\,max}_{\mathbf{x}}
\ \mathcal{J}_{\mathcal{T}}(\mathbf{x}),\\
\textrm{s.t.}\qquad
&\mathbf{x}=\mathcal{B}_{V_f^0}(\mathbf{z}),\quad
\mathbf{z}\sim p_{\mathcal{T}}^{\star}(\mathbf{z}),\quad
\phi_{\mathcal{T}}(\mathbf{x})=1,\quad
V_f(\mathbf{x})=V_f^0.
\end{aligned}
\label{eq:general_task_objective}
\end{equation}
Here, $\mathbf{z}$ denotes the generated signed-distance-function (SDF) field, and $\mathcal{B}_{V_f^0}$ converts $\mathbf{z}$ into a binary topology $\mathbf{x}$ at the prescribed volume fraction $V_f^0$. $\mathcal{J}_{\mathcal{T}}$ denotes the task-specific objective, and $\phi_{\mathcal{T}}$ collects hard feasibility constraints. In multi-objective design, $\mathcal{J}_{\mathcal{T}}$ is replaced by a vector objective and selection is performed by Pareto dominance. The task-adapted distribution $p_{\mathcal{T}}^{\star}$ is obtained by steering the pretrained prior through the GenTO loop. Case-specific objective and constraint definitions are given in Methods Section~\ref{sec:methods_case_objectives}.

The pretrained topology prior is constructed from four representative topology families chosen to span complementary classes of geometry (Fig.~\ref{fig:overview}d; Supplementary Section~\ref{sec:si_datasets} and Supplementary Fig.~\ref{fig:SI_F2}), corresponding to the family-level priors $p_{\mathrm{GRF}}$, $p_{\mathrm{CHPF}}$, $p_{\mathrm{PR}}$, and $p_{\mathrm{TS}}$: Gaussian Random Fields (GRF) yield stochastic smooth morphologies; Cahn-Hilliard Phase Fields (CHPF) produce bicontinuous patterns; Porous Structures (PR) represent heterogeneous media with randomly introduced voids; and Truss-based Structures (TS) capture skeletal load-bearing layouts. The underlying diffusion generator maps Gaussian noise to a continuous SDF field, which is then binarized at the prescribed volume fraction to produce the final topology (Fig.~\ref{fig:overview}c; Supplementary Section~\ref{sec:si_model_architecture}, Supplementary Fig.~\ref{fig:SI_F3}; Supplementary Section~\ref{sec:si_distance_field} and Supplementary Fig.~\ref{fig:SI_F4}). Representative in-distribution samples generated from the pretrained model illustrate the diversity of this learned prior across symmetry groups (Fig.~\ref{fig:overview}e). Details of the random-structure training set and prior construction are provided in Supplementary Section~\ref{sec:si_method_details}.

Building on this formulation, we examine four topology design scenarios for architected metamaterials, covering single-objective optimization, multi-objective optimization, property-targeted design, and function-targeted design. The resulting behavior appears in the final obtained designs and in distribution-level quantities such as score distributions, Pareto fronts, property-space envelopes, satisfaction distributions, and diversity maps.

\subsection{Thermal Conductivity Optimization}

\begin{figure}[htbp!]
    \centering
    \includegraphics[width=0.99\textwidth]{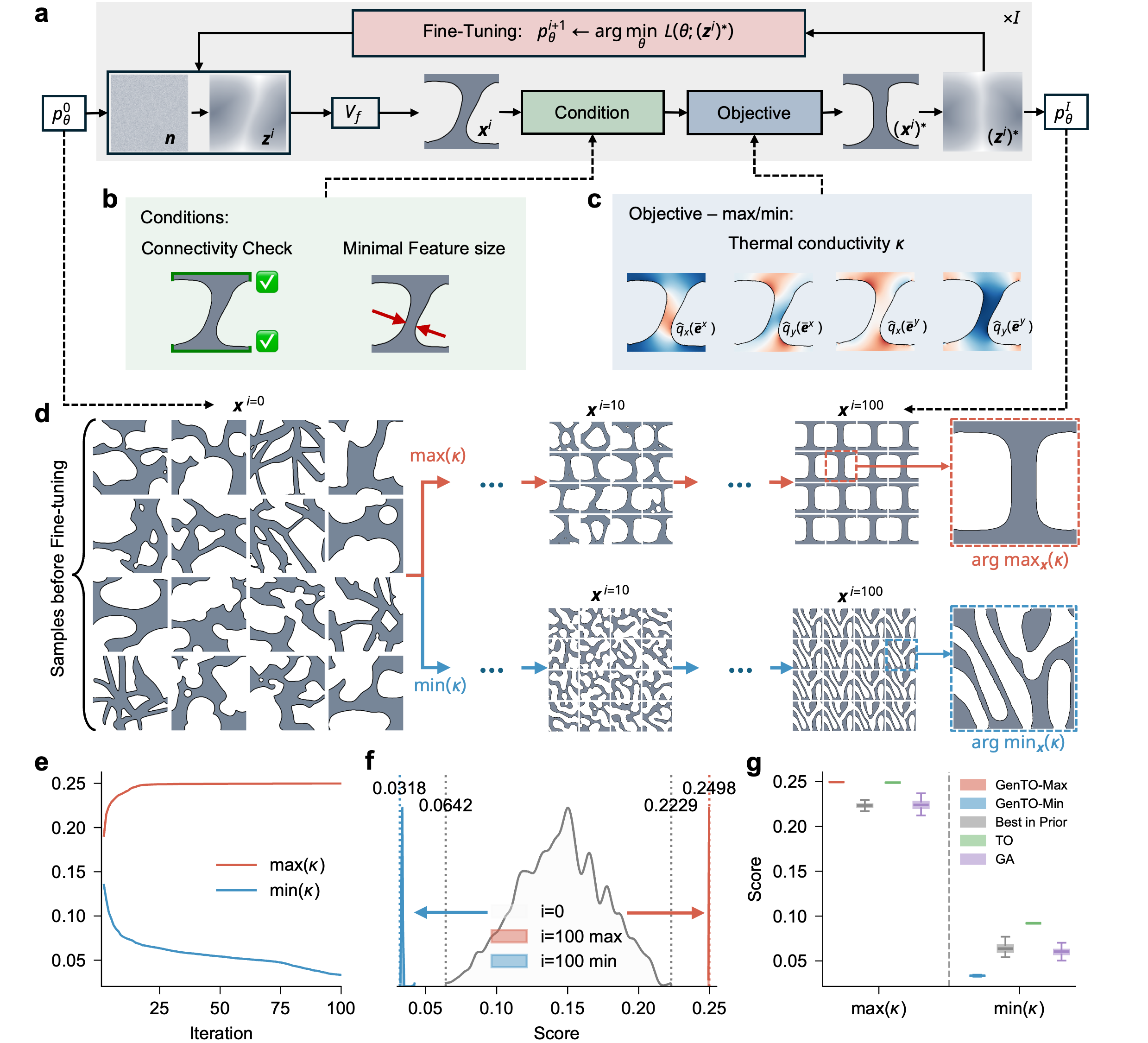}
\caption{\textbf{Thermal conductivity optimization.}
\textbf{a}, Task-specific fine-tuning loop for thermal conductivity design. A candidate is sampled from the current generator $p_i$ by denoising Gaussian noise $\boldsymbol{\epsilon}$ into an SDF field $\mathbf{z}^i$, binarized at $V_f=0.4$ to give a topology $\mathbf{x}^i$, filtered by connectivity and minimum feature-size constraints, scored by the conductivity objective, and then used to fine-tune the model through the loss $L$ for the next generator $p_{i+1}$. The best accepted topology and field are denoted $(\mathbf{x}^i)^*$ and $(\mathbf{z}^i)^*$, and the loop is repeated for $I$ design iterations.
\textbf{b}, Feasibility checks used in this task: full connectivity and minimum feature size.
\textbf{c}, Evaluation of the thermal objective to maximize or minimize the effective thermal conductivity $\kappa$ under orthogonal loading directions. The terms $\hat{q}_x(\bar{\boldsymbol{e}}^{x})$ and $\hat{q}_y(\bar{\boldsymbol{e}}^{y})$ denote normalized averaged heat-flux responses under imposed macroscopic temperature-gradient directions, where the superscripts $x$ and $y$ label the two loading directions.
\textbf{d}, Representative samples before fine-tuning and their evolution over design iterations for conductivity maximization, $\max(\kappa)$, and minimization, $\min(\kappa)$. Final obtained designs are highlighted with dashed boxes.
\textbf{e}, Convergence of the thermal conductivity score over iterations for $\max(\kappa)$ and $\min(\kappa)$.
\textbf{f}, Score distributions at the initial iteration ($i=0$) and after convergence ($i=100$), illustrating how fine-tuning shifts the generated distribution toward higher-performing regions in both tasks. The converged distributions extend beyond the initial score range, indicating out-of-distribution performance relative to the pretrained prior.
\textbf{g}, Benchmarking of GenTO against the Best in Prior baseline, topology optimization (TO), and genetic algorithm (GA) for both $\max(\kappa)$ and $\min(\kappa)$ tasks. The bars summarize normalized performance over 10 runs.}
    \label{fig:thermal_conductivity}
\end{figure}

The first case uses thermal conductivity design as a representative single-objective benchmark for GenTO (Fig.~\ref{fig:thermal_conductivity}; Methods and Supplementary Section~\ref{sec:si_case1}). The goal is to maximize or minimize the effective thermal conductivity of the metamaterial while enforcing periodic connectivity, a minimum feature size, and a fixed material volume fraction. The connectivity condition ensures that the structure forms a fully connected transport path and can therefore sustain non-zero conductivity, whereas the minimum feature-size constraint suppresses arbitrarily thin members and promotes manufacturability (Fig.~\ref{fig:thermal_conductivity}b). The conductivity objective is evaluated as the average effective response under two orthogonal loading directions, $\hat{q}_x\left(\bar{\boldsymbol{e}}^{x}\right)$ and $\hat{q}_y\left(\bar{\boldsymbol{e}}^{y}\right)$, where $\hat{\boldsymbol{q}}$ is the normalized heat-flux vector and $\bar{\boldsymbol{e}}^{x}$ and $\bar{\boldsymbol{e}}^{y}$ are the applied macroscopic temperature-gradient directions (Fig.~\ref{fig:thermal_conductivity}c), computed using FFT-based numerical homogenization~\cite{li2025enhanced}. 

After each round of feasibility screening and scoring, fine-tuning progressively steers the topology distribution toward higher-performing regions, with representative samples evolving from diverse in-distribution topologies into structures clearly adapted to the respective objectives and the corresponding conductivity scores converging steadily over iterations (Fig.~\ref{fig:thermal_conductivity}d,e; Supplementary Video~1). The accompanying score distributions show the same population-level shift for both conductivity maximization and minimization, and their extension beyond the initial score range indicates out-of-distribution performance relative to the pretrained prior (Fig.~\ref{fig:thermal_conductivity}f). Such population-level search is especially valuable for the highly non-convex minimization problem, where GenTO can explore multiple feasible regions instead of following one design trajectory.

Benchmarking against the Best in Prior baseline, topology optimization (TO), and genetic algorithms (GA) further reveals a clear asymmetry between the two objectives (Fig.~\ref{fig:thermal_conductivity}g; Supplementary Sections~\ref{sec:si_baselines} and \ref{sec:si_case1_additional}, Supplementary Figs.~\ref{fig:SI_F6}--\ref{fig:SI_F11}, and Supplementary Table~\ref{tab:SI_P5_T1}). For conductivity maximization, where the optimization landscape is comparatively well behaved, GenTO achieves performance competitive with TO. For conductivity minimization, where gradient-based optimization is more prone to poor local minima, GenTO significantly outperforms both TO and GA. The contrast shows that GenTO is especially effective for irregular non-convex landscapes in which local search can become trapped.

\subsection{Multi-objective Morphological Optimization}

\begin{figure}[htbp!]
    \centering
    \includegraphics[width=0.99\textwidth]{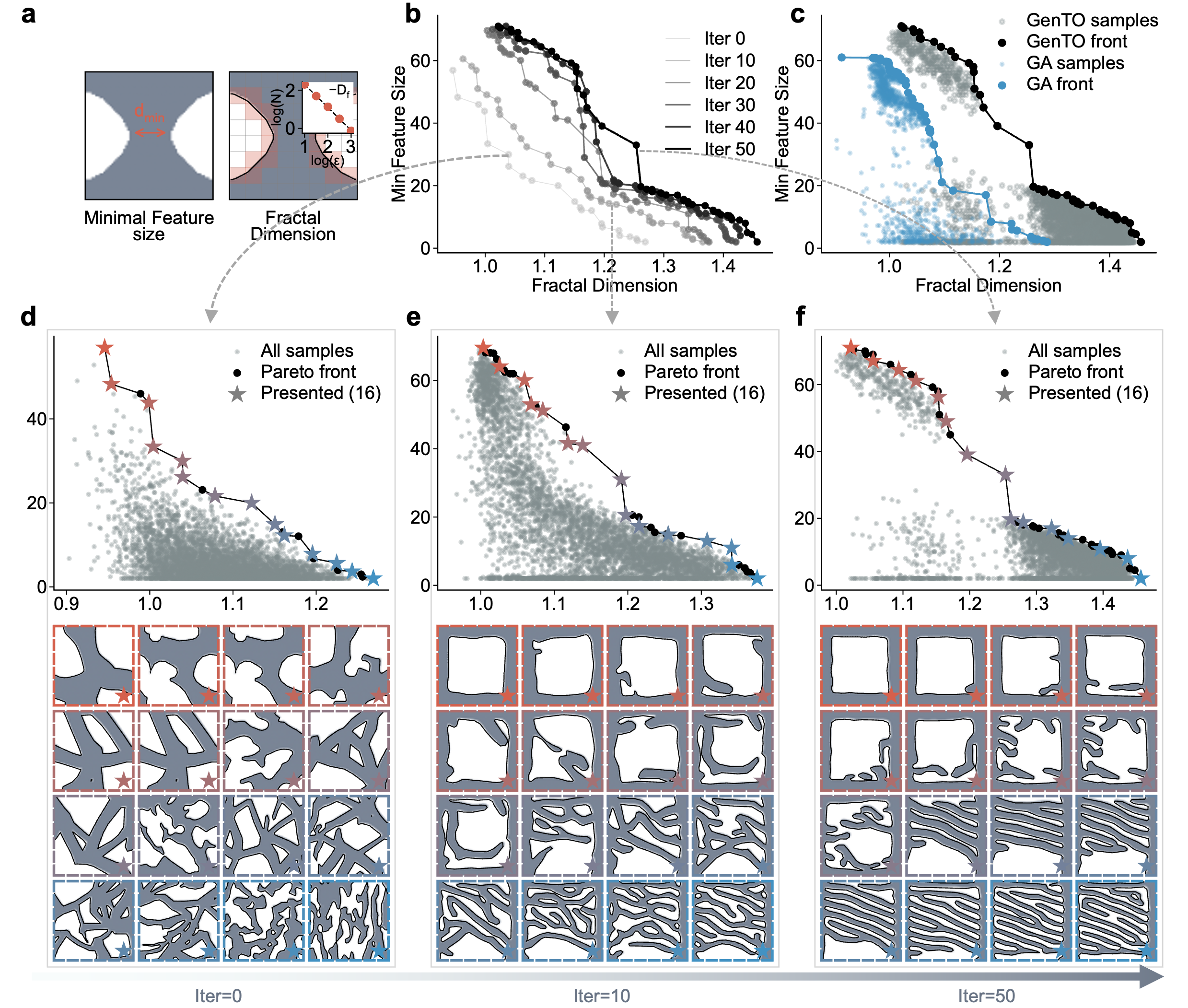}
    \caption{\textbf{Multi-objective morphological optimization.}
\textbf{a}, Two optimization objectives. The minimal feature size quantifies the smallest structural width in the topology, and the fractal dimension measures the geometric complexity of the structure boundary (Methods Section~\ref{sec:methods_case_objectives} and Supplementary Section~\ref{sec:si_case2}).
\textbf{b}, Evolution of the Pareto front in the objective space from iteration 0 to iteration 50, showing progressive outward expansion of the front.
\textbf{c}, Comparison between GenTO and a genetic algorithm (GA) in the objective space. GenTO samples and the corresponding Pareto front cover a broader and better-performing region than GA.
\textbf{d-f}, Sample distributions, Pareto fronts, and representative topology galleries at iterations 0 (\textbf{d}), 10 (\textbf{e}), and 50 (\textbf{f}), respectively. Gray dots denote all sampled designs, black points denote Pareto-front samples, and colored stars mark the representative topologies arranged from large-feature designs to high-fractal-dimension designs.}
    \label{fig:multi_objective}
\end{figure}

A second test moves from scalar optimization to multi-objective design, targeting feature-size maximization and fractal-dimension maximization of the structure surface (Fig.~\ref{fig:multi_objective}; Methods and Supplementary Section~\ref{sec:si_case2}). Larger features promote manufacturability \cite{hernandez2015effect}, whereas a higher fractal dimension favors more complex surface geometries that can enhance physical properties \cite{sanchez2013fractal}. The minimal feature size quantifies the narrowest structural width in the topology, whereas the fractal dimension measures the geometric complexity of the structure boundary (Fig.~\ref{fig:multi_objective}a).
The resulting Pareto-front evolution, sample distributions, and benchmark comparisons are presented in Fig.~\ref{fig:multi_objective}b-f. 

The two objectives are inherently in tension: structures with large, regular features tend to have smoother boundaries and lower fractal dimension, whereas geometrically complex structures naturally exhibit smaller minimum features. The trade-off is already evident in the initial sample distribution, whose Pareto front occupies only a narrow region of the objective space and reflects a limited set of trade-off solutions (Fig.~\ref{fig:multi_objective}d). Over 50 iterations, GenTO progressively expands this front outward as a whole, populating a broader and higher-performing region of the objective space while maintaining substantial structural diversity (Fig.~\ref{fig:multi_objective}b,d-f). The corresponding topology galleries show that this outward expansion is accompanied by a continuous evolution of morphology without collapse onto a few extreme solutions: as the Pareto front advances, representative designs span a wide continuum from large-feature, low-complexity structures to highly intricate high-fractal-dimension morphologies (Fig.~\ref{fig:multi_objective}d-f).

Compared with a genetic algorithm, GenTO achieves a superior Pareto front and covers a broader region of the objective space under a comparable evaluation budget (Fig.~\ref{fig:multi_objective}c; Supplementary Sections~\ref{sec:si_baselines} and \ref{sec:si_case2_additional}, Supplementary Figs.~\ref{fig:SI_F12}--\ref{fig:SI_F14}, and Supplementary Table~\ref{tab:SI_P5_T2}). The gain is therefore reflected in both the extreme objective values and the breadth and continuity of the trade-off designs recovered during optimization.

\subsection{Property-Targeted Design of Auxetic Metamaterials}

\begin{figure}[htbp!]
    \centering
    \includegraphics[width=0.99\textwidth]{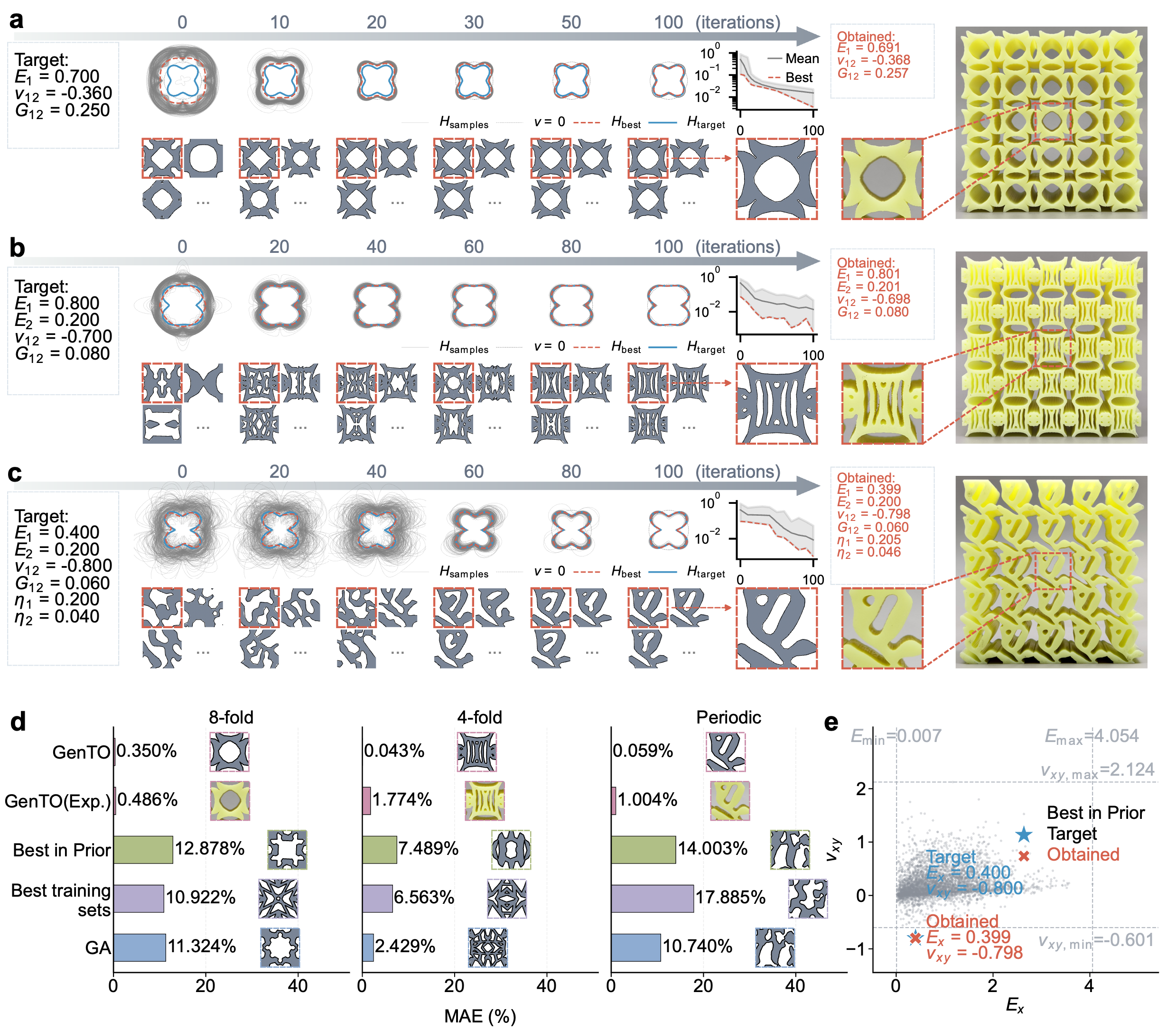}
\caption{\textbf{Property-targeted design of auxetic metamaterials.}
\textbf{a-c}, Three property-targeted design tasks with increasing constraints on the elasticity tensor. In each panel, the target-property box specifies the prescribed engineering constants, including Young's moduli $E_1$ and $E_2$, Poisson's ratio $\nu_{12}$, shear modulus $G_{12}$, and, when required, coupling parameters $\eta_1$ and $\eta_2$. The response plots show Poisson's-ratio polar envelopes over iterations as gray curves, where $H_{\mathrm{samples}}$, $H_{\mathrm{best}}$, and $H_{\mathrm{target}}$ denote sampled, best obtained, and target homogenized elastic responses, and $\nu=0$ marks the zero-Poisson's-ratio reference. The target response is shown in blue and the best obtained response in red. The topology sequences below show representative samples across design iterations, and the inset plots report the mean and minimum error during optimization. The right panels list the obtained properties and show the corresponding designed unit cell, one-cell experimental image, and 3D-printed lattice.
\textbf{d}, Quantitative comparison of constitutive-matrix mean absolute error (MAE) across three symmetry groups: 8-fold, 4-fold, and Periodic. For numerical designs, the error is computed between the target effective elasticity tensor $\mathbf{H}_{\mathrm{ES,target}}$ and the obtained tensor $\mathbf{H}_{\mathrm{ES}}$. The comparison baselines include GenTO, experimental GenTO designs, Best in Prior, the Best in training sets baseline, and genetic algorithm (GA). Representative topologies or experimental unit cells produced by the different methods are also shown.
\textbf{e}, Target and obtained designs in property space relative to the pretrained distribution and its convex hull, where $E_x$ and $E_y$ denote directional Young's moduli and $\nu_{xy}$ denotes the in-plane Poisson's ratio. The limits $E_{\min}$, $E_{\max}$, $\nu_{xy,\min}$, and $\nu_{xy,\max}$ mark the range of the pretrained distribution, illustrating the out-of-distribution design capability of GenTO.
}
\label{fig:property_targeted}
\end{figure}

The third case is a property-targeted inverse-design problem, in which the goal is to identify metamaterial topologies whose mechanical response matches a prescribed elasticity tensor $\mathbf{H}_{\mathrm{ES,target}}$ as closely as possible (Fig.~\ref{fig:property_targeted}; Methods and Supplementary Section~\ref{sec:si_case3}). Such problems are particularly challenging because the mapping from topology to properties is highly nonlinear, and the target may correspond to unconventional mechanical behavior. We focus on auxetic metamaterials, characterized by a negative Poisson's ratio, which represent a canonical and practically important class of architected structures \cite{ren2018auxetic}. The resulting optimization trajectories, benchmark comparisons, and property-space analysis are presented in Fig.~\ref{fig:property_targeted}.

Three design tasks of increasing difficulty, each targeting auxetic behavior under progressively more constrained elasticity tensors, are presented in Fig.~\ref{fig:property_targeted}a-c. In each case, the target properties are specified on the left-hand side, and the optimization proceeds toward a final design whose effective response most closely matches the prescribed tensor. Along this process, the Poisson's-ratio polar envelopes contract from a broad initial distribution toward the target response, while the accompanying topology sequences evolve from generic in-distribution samples into structures with increasingly well-defined auxetic features, ending in the best-matched design shown on the right-hand side of each panel (Supplementary Video~2). The corresponding convergence plots show steady reduction in both mean and minimum error over iterations, with the best designs at iteration 100 reaching reported constitutive-matrix mean absolute errors of approximately 0.04\%--0.35\% relative to the target tensors.

Quantitative benchmarking further shows that GenTO consistently outperforms Best in Prior, the Best in training sets baseline, and genetic algorithms across all three symmetry groups, including 8-fold, 4-fold, and Periodic designs (Fig.~\ref{fig:property_targeted}d; Supplementary Sections~\ref{sec:si_baselines} and \ref{sec:si_case3_additional}, Supplementary Figs.~\ref{fig:SI_F15}--\ref{fig:SI_F17}, and Supplementary Table~\ref{tab:SI_P5_T3}). The performance gain is accompanied by representative topologies from each method, showing that the improved property-matching accuracy is achieved through distinct and well-formed auxetic architectures, not trivial variation around the pretrained prior. Within the scope of the methods and settings considered here, these results indicate that GenTO achieves state-of-the-art accuracy for this class of property-targeted inverse-design problem.

A particularly important result emerges in property space (Fig.~\ref{fig:property_targeted}e; Supplementary Section~\ref{sec:si_case3}): both the target elasticity tensor and the final obtained design lie outside the convex hull of the pretrained distribution. GenTO therefore does more than retrieve structures already represented in the original prior; it steers the model into previously unvisited regions of the property space. For property-targeted inverse design, this capability is especially valuable, as it allows the framework to synthesize structures with mechanical responses beyond those directly accessible from any fixed pretrained distribution, substantially expanding the reachable design space.

\subsection{Vibration Transmission Design}

\begin{figure}[htbp!]
    \centering
    \includegraphics[width=0.999\textwidth]{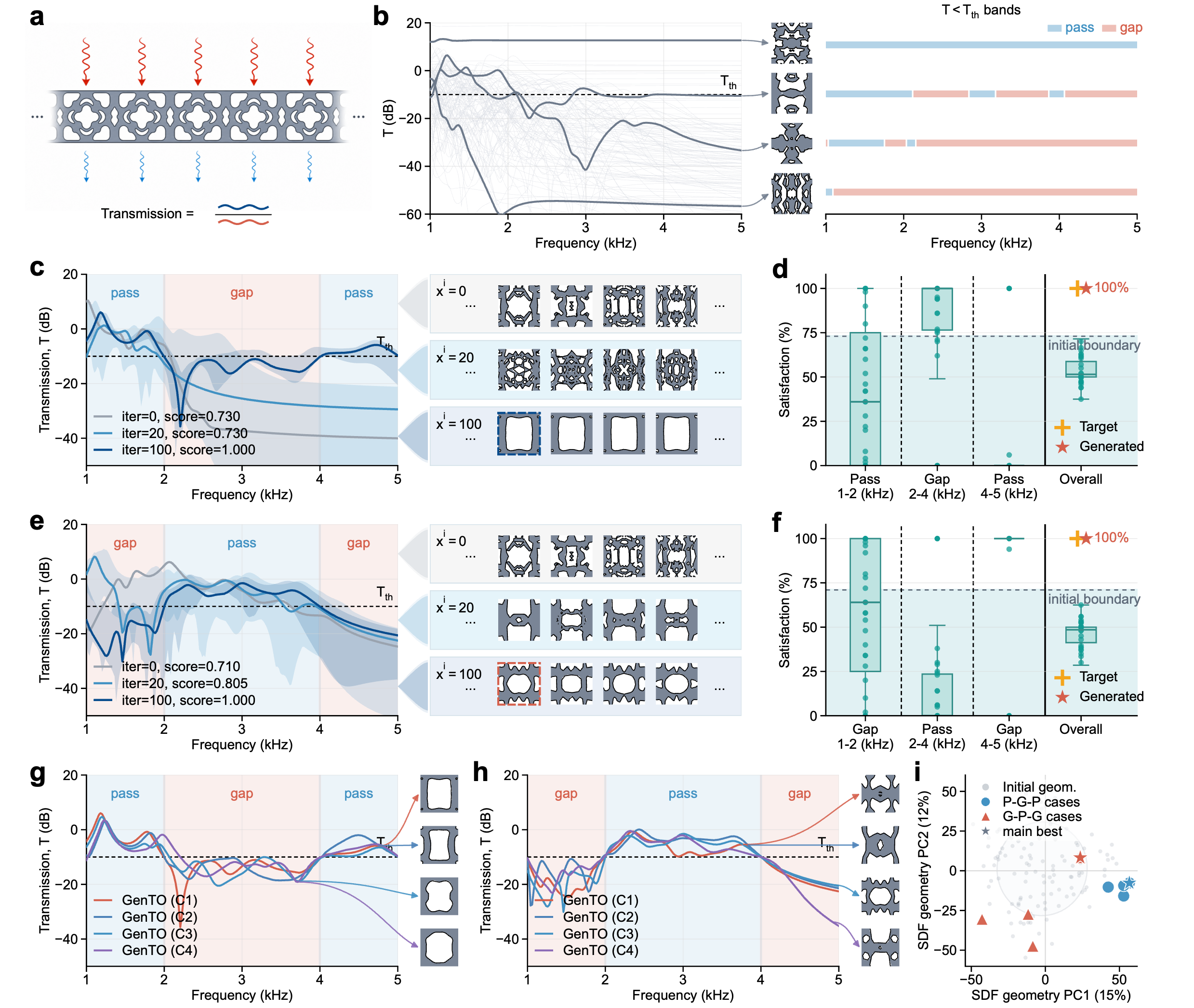}
    \caption{\textbf{Vibration transmission design.}
\textbf{a}, Schematic of the vibration transmission task and the transmission metric used for evaluation, where $T_{\mathbf{x}}(f)$ denotes the transmission spectrum of topology $\mathbf{x}$ at frequency $f$ and $T_{\mathrm{th}}$ is the pass--gap threshold.
\textbf{b}, Initial transmission spectra from candidate unit cells, four representative topologies, and the corresponding pass-gap maps.
\textbf{c}, Pass-gap-pass design with target pass bands $[1,\,2]$ and $[4,\,5]$~kHz and stop band $[2,\,4]$~kHz. The spectra at iterations 0, 20, and 100 are shown together with representative topology evolution.
\textbf{d}, Satisfaction distribution for the pass-gap-pass task across the three frequency regions and overall response. The generated design reaches 100\% overall satisfaction.
\textbf{e}, Gap-pass-gap design with target pass band $[2,\,4]$~kHz and stop bands $[1,\,2]$ and $[4,\,5]$~kHz. The spectra at iterations 0, 20, and 100 are shown together with representative topology evolution.
\textbf{f}, Satisfaction distribution for the gap-pass-gap task across the three frequency regions and overall response.
\textbf{g}, Additional pass-gap-pass optimization cases, GenTO (C1)--GenTO (C4), showing distinct high-performing transmission responses and corresponding topologies.
\textbf{h}, Additional gap-pass-gap optimization cases, GenTO (C1)--GenTO (C4), showing multiple valid topologies under the same target function.
\textbf{i}, Diversity map of generated designs and the selected solutions, showing that high-satisfaction responses can be obtained from different regions of the generated distribution.}
    \label{fig:vibration_transmission}
\end{figure}

The final example turns to a qualitatively different class of problem: function-targeted design in the frequency domain (Fig.~\ref{fig:vibration_transmission}; Methods and Supplementary Section~\ref{sec:si_case4}). The goal is to design metamaterial structures that realize prescribed band-gap or pass-band responses, a capability with direct relevance to vibration isolation, waveguiding, and acoustic metamaterial engineering \cite{cummer2016controlling,gao2022acoustic,an20203d}. Unlike the previous cases, the design target is an entire transmission curve, and the resulting threshold-based satisfaction score is non-differentiable with respect to the structural variables, making the problem particularly challenging for gradient-based methods. For each candidate topology, the transmission spectrum $T_{\mathbf{x}}(f)$ is evaluated over the frequency range of interest, and the score rewards high transmission within the target pass bands and low transmission within the target stop bands (Fig.~\ref{fig:vibration_transmission}a). Two target configurations are considered: a dual pass band $\Omega_{\mathrm{p}} = [1,\,2]\cup[4,\,5]$~kHz and a single pass band $\Omega_{\mathrm{p}} = [2,\,4]$~kHz.

The resulting optimization dynamics are summarized in Fig.~\ref{fig:vibration_transmission}. The initial spectra already show that different unit-cell geometries can realize very different pass--gap patterns, motivating a distribution-level search instead of direct inversion from a target spectrum (Fig.~\ref{fig:vibration_transmission}b). For the pass--gap--pass target, the spectra at iterations 0, 20, and 100 move toward high transmission in the two pass bands and low transmission in the central gap band, while the accompanying topology sequence shows the corresponding structural adaptation (Fig.~\ref{fig:vibration_transmission}c). The region-wise satisfaction distribution confirms that the generated design approaches the target response both within individual frequency regions and overall (Fig.~\ref{fig:vibration_transmission}d). The gap--pass--gap target exhibits the complementary behavior, with transmission concentrated in the central pass band and suppressed in the two outer gaps (Fig.~\ref{fig:vibration_transmission}e,f; Supplementary Video~3). Additional optimization runs, denoted GenTO (C1)--GenTO (C4), further show that the same target function can admit multiple different high-performing topologies in both target settings (Fig.~\ref{fig:vibration_transmission}g,h; Supplementary Section~\ref{sec:si_case4_additional} and Supplementary Figs.~\ref{fig:SI_F19} and \ref{fig:SI_F20}), and the diversity map shows that these solutions occupy different regions of the generated distribution (Fig.~\ref{fig:vibration_transmission}i). The single-pass-band cases are more widely separated in geometry space, whereas the dual-pass-band cases are more clustered, suggesting that the former admits more geometrically distinct high-satisfaction basins.

The example demonstrates that GenTO can handle a non-differentiable objective: the score enters only through evaluated transmission responses, without requiring gradients or an explicit analytical description of the optimization landscape. The final designs achieve target-satisfaction scores close to 100.0\%. Static-reference comparisons with the best training-set and fixed-prior samples are provided in Supplementary Sections~\ref{sec:si_baselines} and \ref{sec:si_case4_additional}, and Supplementary Table~\ref{tab:SI_P5_T4}.

\subsection{Theoretical Interpretations and Underlying Mechanisms}

\begin{figure}[t]
    \centering
    \includegraphics[width=0.999\textwidth]{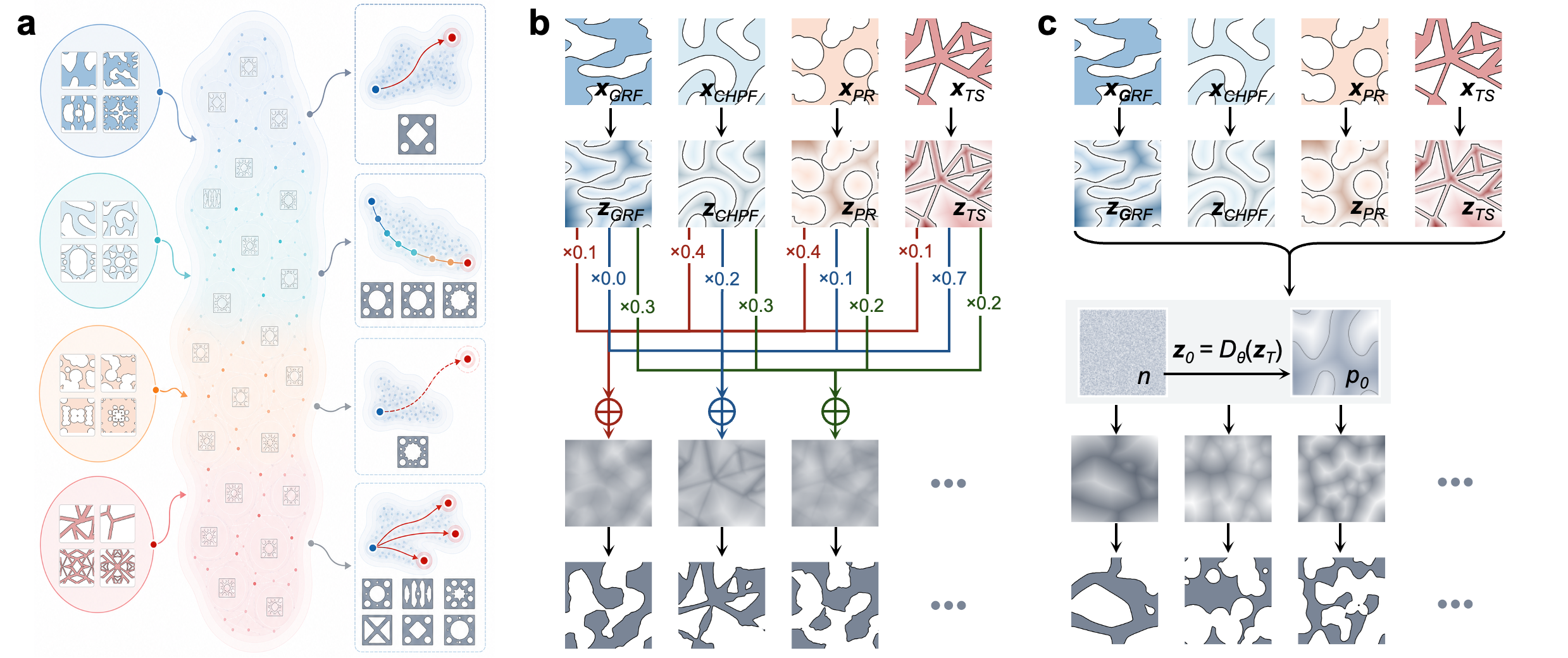}
    \caption{\textbf{Signed-distance-function representation for joint topology generation.}
\textbf{a}, Conceptual view of the joint topology distribution learned from multiple structure families. The learned prior covers different morphology classes and supports several distribution-steering behaviors, including optimization toward an optimum, Pareto-front migration, out-of-distribution targeting, and diverse inverse-design solutions.
\textbf{b}, Signed-distance-function (SDF) representation of topology samples. Binary structures from Gaussian Random Fields, Cahn-Hilliard Phase Fields, Porous Structures, and Truss-based Structures are denoted $\mathbf{x}_{\mathrm{GRF}}$, $\mathbf{x}_{\mathrm{CHPF}}$, $\mathbf{x}_{\mathrm{PR}}$, and $\mathbf{x}_{\mathrm{TS}}$, and their SDF fields are denoted $\mathbf{z}_{\mathrm{GRF}}$, $\mathbf{z}_{\mathrm{CHPF}}$, $\mathbf{z}_{\mathrm{PR}}$, and $\mathbf{z}_{\mathrm{TS}}$. Linear superposition of these SDF fields with the shown weights can generate new topologies, showing that the SDF space is suitable for cross-family structure generation.
\textbf{c}, GenTO learns this SDF distribution with a diffusion model. Starting from Gaussian noise $\boldsymbol{\epsilon}$, the learned denoiser/generator $D_{\theta}$ maps the noisy field $\mathbf{z}_{T}$ to a clean SDF field $\mathbf{z}_{0}$, written as $\mathbf{z}_{0}=D_{\theta}(\mathbf{z}_{T})$, and the resulting prior $p_{0}$ generates SDF samples that are converted into binary topologies.}
    \label{fig:summary_framework}
\end{figure}

Across the four tasks above, the common element is a reusable topology representation that can be steered by different objectives. Fig.~\ref{fig:summary_framework} summarizes this representation-level mechanism. The pretrained prior is not a collection of isolated topology families, but a joint distribution over multiple morphology classes that can be steered toward different task-dependent regions (Fig.~\ref{fig:summary_framework}a; Supplementary Section~\ref{sec:si_datasets}). The joint prior supports scalar optimization, Pareto-front migration, out-of-distribution targeting, and diverse inverse-design solutions within one procedure. A key ingredient is the signed-distance function (SDF) representation: after binary topologies are mapped to smooth SDF fields, even simple linear superposition in this space can yield coherent new structures instead of broken pixel-level mixtures (Fig.~\ref{fig:summary_framework}b; Supplementary Section~\ref{sec:si_distance_field} and Supplementary Fig.~\ref{fig:SI_F4}). GenTO therefore trains the diffusion model directly on SDF fields and converts generated fields back to binary topologies only after sampling, giving the model a smoother representation in which cross-family generation and distribution steering are easier to realize (Fig.~\ref{fig:summary_framework}c; Methods and Supplementary Section~\ref{sec:si_learning_steering}).

\section{Discussion}
\label{sec:discussion}

GenTO introduces a reusable design paradigm for architected metamaterials, centered on a learned topology prior that can be adapted across tasks. Instead of reconstructing the design process for each new objective, constraint, or physical function, a task-matched prior is steered toward high-performing regions, making topology knowledge reusable across heterogeneous design problems.

Across the full set of experiments, GenTO delivers strong performance under markedly different design regimes (Supplementary Section~\ref{sec:si_additional_results}). For fair comparison, the population-based baselines were evaluated under the same evaluation budget as GenTO, while trajectory-based topology optimization was treated separately (Supplementary Section~\ref{sec:si_discussion_efficiency} and Supplementary Tables~\ref{tab:SI_eval_budget_definition} and \ref{tab:SI_eval_budget}). The distribution-level formulation preserves structural diversity while concentrating generation around task-specific high-performing regions, and in several cases expands the reachable design space beyond the support of the initial pretrained prior. 

At a practical level, this formulation opens a more effective and more generalizable route for architected metamaterial design in application-driven settings. The ability to adapt a learned topology prior across changing objectives and constraints is especially relevant for multifunctional architected systems, where different performance requirements must be explored within the same structural design space, e.g., lightweight load-bearing performance, manufacturability, vibration or acoustic control, and programmable mechanical response \cite{de2007topological,yeo2019structurally,surjadi2019mechanical}. Such flexibility is important for areas in which architected materials are increasingly being considered, including aerospace and automotive structures \cite{schaedler2011ultralight,zheng2014ultralight}, biomedical implants and tissue-engineering scaffolds \cite{hollister2005porous,zadpoor2020meta}, vibration and wave-control devices \cite{cummer2016controlling,assouar2018acoustic}, and multi-purpose components in robotics and electronics \cite{xia2022responsive,he2024programmable}. In these settings, GenTO offers a way to retain and reuse accumulated structural knowledge while continuing to adapt the design process to new performance targets.

From a materials-design perspective, the key point is the separation between learning structural space and optimizing task-specific physical objectives. The prior can be enriched as new structural classes become available, while user-defined objectives determine how the design process moves within or beyond that space. Such modularity suggests that what is transferred is a general description of design space itself, rather than a task-specific inverse map.

The perspective also connects GenTO to an emerging direction in AI for Science. Recent advances such as AlphaFold and its successors have shown how learned structural priors can transform scientific workflows by providing reusable representations for downstream analysis, discovery, and design \cite{jumper2021highly,abramson2024accurate}. Similar developments in generative molecular design, biomolecular modeling, and materials foundation models suggest that this is not an isolated pattern, but an emerging principle of scientific AI \cite{gomezbombarelli2018automatic,watson2023denovo,merchant2023scaling}. GenTO exemplifies this principle in topology design: a pretrained generative model supplies a reusable structural prior, and scientific objectives steer that prior toward new regions of design space. Reusable generative priors may therefore become a general design primitive across scientific domains wherever structured spaces must be adapted to changing downstream objectives. 

GenTO marks an initial step toward reusable generative design, and several limitations remain at the current stage. The expressive range of the framework is still shaped by the coverage and richness of the pretrained topology prior, and will broaden as larger and more diverse structural datasets become available. Repeated test-time evaluation and fine-tuning can also be computationally demanding, motivating more efficient optimization schemes and lighter adaptation strategies. In addition, the present study focuses on two-dimensional architected systems under relatively controlled constraints, providing a foundation for extension to more realistic design settings but not defining the boundary of the framework.

Several further directions follow naturally from the formulation itself. A more rigorous theory of distribution steering is needed, particularly for its out-of-distribution behavior, its dependence on prior coverage, and its interaction with objective landscapes and model updates. Equally important is the extension of the framework to more realistic settings, including three-dimensional topologies, multimaterial systems, stronger manufacturing constraints, coupled multiphysics problems, robust design, uncertainty-aware design, and more complex function-valued targets in space and time \cite{zegard2016bridging,bayat2023holistic}. Making these extensions practical will also require more efficient use of physical evaluations. Future implementations could reduce the required evaluation budget through surrogate-assisted screening, active learning, or uncertainty-guided candidate selection, while retaining the same distribution-steering framework.

Looking ahead, reusable structural priors may provide a practical foundation for generative design in architected metamaterials and beyond. As prior learning and distribution steering are extended to richer structural spaces and more realistic settings, generative models may increasingly support the continual reuse of accumulated design knowledge across evolving scientific and engineering objectives.

\section{Methods}
\label{sec:methods}
GenTO consists of an offline prior-learning stage and an online distribution-steering stage, as summarized in Fig.~\ref{fig:overview}. Instead of solving each design task by retraining a task-specific conditional generator, GenTO starts from a reusable pretrained topology prior and then iteratively shifts this prior toward a user-defined high-performing region of the design space. Additional implementation details are provided in Supplementary Section~\ref{sec:si_method_details}.

\subsection{GenTO Framework}
\label{sec:methods_gento_framework}
GenTO consists of an offline stage that learns a reusable prior over representative topology classes; and an online stage that adapts this prior to downstream design tasks through iterative sampling, screening, and fine-tuning.

\runinhead{Offline Prior Learning.}
Let $\mathcal{D}_{x}=\{\mathbf{x}^{(n)}\}_{n=1}^{N}$ denote the training set of binary periodic unit-cell topologies collected from four topology families: Gaussian Random Fields, Cahn-Hilliard Phase Fields, Porous Structures, and Truss-based Structures. These four families were chosen to cover complementary morphology generators, including stochastic smooth patterns, phase-separated patterns, porous media, and truss-like load-bearing layouts. Other representative families can be incorporated in GenTO in a similar way. Each topology family is used to generate a large number of samples for three symmetry types: Periodic, 4-fold, and 8-fold (Supplementary Section~\ref{sec:si_datasets} and Supplementary Fig.~\ref{fig:SI_F2}). These three symmetry types were chosen to cover common periodic and rotationally symmetric unit-cell layouts while keeping the training set compact. Instead of directly learning on binary images, each topology is converted into a signed-distance function (SDF) map
\begin{equation}
\mathbf{z}^{(n)}
= \operatorname{SDF}\!\left(\mathbf{x}^{(n)}\right),
\end{equation}
where $\operatorname{SDF}(\cdot)$ denotes the signed-distance function~\cite{osher1988fronts}, such that $\mathbf{z}^{(n)}>0$ in the material phase and $\mathbf{z}^{(n)}<0$ in the void phase. In the Python implementation, the two unsigned distance maps used to construct this SDF are obtained with a Euclidean distance transform (EDT) routine. This SDF representation converts a discontinuous binary topology into a smooth continuous field, which is better suited for generative modeling and later makes interpolation between topologies meaningful (Supplementary Section~\ref{sec:si_distance_field}). Denoting the resulting continuous-field dataset by $\mathcal{D}_{z}=\{\mathbf{z}^{(n)}\}_{n=1}^{N}$, a diffusion model parameterized by $\theta$ is pretrained on $\mathcal{D}_{z}$ to learn a broad prior $p_{0}(\mathbf{z})\approx p_{\mathrm{data}}(\mathbf{z})$~\cite{ho2020denoising} for each symmetry type. Diffusion models learn a data distribution by gradually corrupting training samples with Gaussian noise in a forward process and then learning the reverse denoising process that reconstructs clean samples from noisy ones. This choice is well suited to broad multimodal geometry distributions in the continuous SDF representation.
For a clean SDF field $\mathbf{z}_{0}$, the forward diffusion process is
\begin{equation}
\mathbf{z}_{t}=\sqrt{\bar{\alpha}_{t}}\,\mathbf{z}_{0}+\sqrt{1-\bar{\alpha}_{t}}\,\boldsymbol{\epsilon},
\qquad
\boldsymbol{\epsilon}\sim\mathcal{N}(\mathbf{0},\mathbf{I}),
\end{equation}
where $t$ is the diffusion step, $\bar{\alpha}_{t}$ is the cumulative noise schedule, and $\mathbf{I}$ is the identity covariance. The forward process progressively destroys geometric detail as $t$ increases, whereas generation follows the learned reverse process from Gaussian noise back to a clean topology field (Supplementary Fig.~\ref{fig:SI_F3}). The denoising network $f_{\theta}(\mathbf{z}_{t},t)$ takes a noisy field and its diffusion step as input and is trained using a $\mathbf{z}_{0}$-prediction objective
\begin{equation}
\min_{\theta}\;
\mathbb{E}_{\mathbf{z}_{0}\sim\mathcal{D}_{z},\,t,\,\boldsymbol{\epsilon}}
\left[
\left\|f_{\theta}(\mathbf{z}_{t},t)-\mathbf{z}_{0}\right\|_{2}^{2}
\right],
\end{equation}
where $\mathbb{E}[\cdot]$ denotes expectation over clean training fields, diffusion steps, and Gaussian noise realizations, and $\|\cdot\|_2$ denotes the Euclidean norm.
At inference, the denoised SDF field $\hat{\mathbf{z}}_{0}$ is converted into a binary topology at the prescribed volume fraction $V_{f}$ by sample-wise quantile thresholding,
\begin{equation}
\mathcal{B}_{V_{f}}(\hat{\mathbf{z}}_{0})
=
\mathbf{1}\!\left[
\hat{\mathbf{z}}_{0}>q_{1-V_{f}}(\hat{\mathbf{z}}_{0})
\right],
\end{equation}
which yields the binary structure used for evaluation. Here, $q_{1-V_{f}}(\hat{\mathbf{z}}_{0})$ denotes the sample-wise $(1-V_f)$-quantile of the predicted field values, and $\mathbf{1}[\cdot]$ denotes the indicator function. Using an SDF representation keeps the learned geometry field continuous, while the quantile threshold ensures that the generated structure exactly matches the prescribed volume fraction. The joint training distribution, the SDF representation, and the resulting reusable diffusion prior are schematically illustrated in Fig.~\ref{fig:summary_framework}a-c. This pretrained model defines the initial generative prior from which all subsequent design iterations start. The generation process and model details are described in Supplementary Section~\ref{sec:si_model_architecture}.
Throughout the design loop, the binary topology is therefore written as $\mathbf{x}=\mathcal{B}_{V_f}(\mathbf{z})$.

\runinhead{Online Design as Distribution Steering.}
During the online stage, GenTO adapts the pretrained prior to each downstream design task.
The same update loop is reused across tasks: candidate topologies are sampled from the current distribution, evaluated by task-specific criteria, screened by feasibility and performance, and then used for fine-tuning so that probability mass gradually shifts toward the target region (Supplementary Section~\ref{sec:si_learning_steering}). For a user-defined task, let $r(\mathbf{x})$ denote the scalar design score and let $\phi_{\mathrm{task}}(\mathbf{x})\in\{0,1\}$ denote a feasibility indicator collecting all task-specific constraints, such as connectivity, symmetry, target volume fraction, or minimum feature-size requirements. For scalar tasks, $r(\mathbf{x})$ is the implementation-level form of the task objective $\mathcal{J}_{\mathcal{T}}(\mathbf{x})$ in Eq.~\eqref{eq:general_task_objective}. This separation allows the same generative update rule to be reused while only changing the task-specific evaluation functions. The design problem is written in the generic form
\begin{equation}
\mathbf{x}^{\star}\in
\arg\max_{\mathbf{x}}\, r(\mathbf{x})
\qquad
\textrm{s.t.}
\qquad
\phi_{\mathrm{task}}(\mathbf{x})=1.
\end{equation}
Minimization tasks are handled by maximizing a negated error score, while multi-objective tasks are handled by retaining feasible non-dominated samples or Pareto layers.

At iteration $i$, the current model defines a distribution $p_{i}(\mathbf{z})$ in the SDF space. A batch of candidate fields $\mathcal{S}_{i}=\{\mathbf{z}_{i}^{(k)}\}_{k=1}^{B}$ is sampled from $p_{i}$, converted into binary topologies $\mathbf{x}_{i}^{(k)}=\mathcal{B}_{V_{f}}(\mathbf{z}_{i}^{(k)})$, and evaluated by the user-defined score and feasibility criteria. The screened subset defines a filtered distribution
\begin{equation}
\tilde{p}_{i}(\mathbf{z})
=
\frac{p_{i}(\mathbf{z})\,\mathbf{1}\!\left[\mathbf{z}\in\Omega_{i}\right]}{Z_{i}},
\qquad
\Omega_{i}
=
\left\{
\mathbf{z}:\phi_{\mathrm{task}}\!\left(\mathcal{B}_{V_{f}}(\mathbf{z})\right)=1,\;
r\!\left(\mathcal{B}_{V_{f}}(\mathbf{z})\right)\ge\tau_{i}
\right\},
\end{equation}
where $\tau_{i}$ is the selection threshold, $\Omega_i$ denotes the accepted subset in field space, and $Z_i$ normalizes the truncated density so that $\tilde{p}_i$ remains a valid probability distribution. Here, $p_{i}$ is the generated distribution, whereas $\tilde{p}_{i}$ is the filtered distribution that retains only feasible and high-performing samples. For diffusion fine-tuning, a clean accepted field $\mathbf{z}_{0}\sim\tilde{p}_{i}$ is corrupted by the standard forward noising process to obtain $\mathbf{z}_{t}$ at diffusion time $t$. The model is then updated with the same denoising objective,
\begin{equation}
\theta_{i+1}
=
\arg\min_{\theta}\;
\mathbb{E}_{\mathbf{z}_{0}\sim\tilde{p}_{i},\,t,\,\boldsymbol{\epsilon}}
\left[
\left\|f_{\theta}(\mathbf{z}_{t},t)-\mathbf{z}_{0}\right\|_{2}^{2}
\right],
\end{equation}
which induces the updated distribution
\begin{equation}
p_{i+1}(\mathbf{z})=\mathcal{U}(p_{i}(\mathbf{z}),\tilde{p}_{i}(\mathbf{z})).
\end{equation}
The operator $\mathcal{U}$ denotes the implicit distribution update induced by fine-tuning the diffusion model on samples from $\tilde{p}_i$.
This yields the distributional chain illustrated in \textbf{Fig.~\ref{fig:overview}}:
\begin{equation}
\mathcal{D}_{x}
\longrightarrow
\mathcal{D}_{z}
\longrightarrow
p_{0}(\mathbf{z})
\longrightarrow
p_{i}(\mathbf{z})
\longrightarrow
\tilde{p}_{i}(\mathbf{z})
\longrightarrow
p_{i+1}(\mathbf{z})
\longrightarrow
\mathbf{x}^{\star}=\mathcal{B}_{V_{f}}(\mathbf{z}^{\star}).
\end{equation}
Representative steering behaviors of this same mechanism, including in-support optimization, multi-objective distribution migration, OOD search, and diverse inverse-design trajectories, are summarized in Fig.~\ref{fig:summary_framework}a. The single-objective steering procedure is listed in Supplementary Algorithm~\ref{alg:si_single_objective_steering}.

\subsection{Case-specific Objective Formulations}
\label{sec:methods_case_objectives}
Eq.~\eqref{eq:general_task_objective} gives the common optimization form used in the Results section. The same distribution-steering loop is used in all four case studies, but the task-specific objective $\mathcal{J}_{\mathcal{T}}$ and the feasibility constraint $\phi_{\mathcal{T}}$ change from case to case. The formulas below define these objective and constraint terms explicitly, with additional case-specific settings provided in Supplementary Section~\ref{sec:si_case_details}.

\runinhead{Thermal conductivity extremization.}
For the thermal case (Supplementary Section~\ref{sec:si_case1}), the design variable is the binary periodic topology $\mathbf{x}$ obtained from a generated SDF field by volume-fraction-controlled binarization. The task objective is the effective conductivity score,
\begin{equation}
\begin{aligned}
\max_{\mathbf{x}} \;
r_{\kappa}(\mathbf{x};\sigma_{\kappa})
&=
\sigma_{\kappa}\bar{\kappa}_{\mathrm{eff}}(\mathbf{x})
-
\lambda_{\kappa}\Delta\kappa_{\mathrm{eff}}(\mathbf{x}),\\
\textrm{s.t.}\qquad
\phi_{\mathrm{conn}}(\mathbf{x})&=1,\quad
w_{\min}(\mathbf{x})\geq w_0,\quad
V_f(\mathbf{x})=V_f^0,
\end{aligned}
\end{equation}
where $\sigma_{\kappa}=1$ gives conductivity maximization and $\sigma_{\kappa}=-1$ gives conductivity minimization. The mean response and anisotropy penalty are computed from the homogenized thermal response as
\begin{equation}
\begin{aligned}
\bar{\kappa}_{\mathrm{eff}}
&=
\frac{
\hat{q}_x(\bar{\boldsymbol{e}}^{x})
+
\hat{q}_y(\bar{\boldsymbol{e}}^{y})
}{2},
\qquad
\Delta\kappa_{\mathrm{eff}}
=
\left\lvert
\hat{q}_x(\bar{\boldsymbol{e}}^{x})
-
\hat{q}_y(\bar{\boldsymbol{e}}^{y})
\right\rvert.
\end{aligned}
\end{equation}
Here, $\hat{q}_x(\bar{\boldsymbol{e}}^{x})$ and $\hat{q}_y(\bar{\boldsymbol{e}}^{y})$ denote the normalized averaged heat-flux responses under the two orthogonal macroscopic temperature-gradient directions (Fig.~\ref{fig:thermal_conductivity}c). The superscripts $x$ and $y$ label the imposed loading directions, while the subscripts on $\hat{q}$ label the scalar response extracted from each loading. Equivalently, these two responses correspond to the diagonal entries $H_{\kappa,11}$ and $H_{\kappa,22}$ of the homogenized conductivity tensor, and $\lambda_{\kappa}$ is a mild anisotropy penalty weight. The subject-to terms in Eq.~\eqref{eq:general_task_objective} are instantiated as periodic connectivity, minimum feature size, and the prescribed volume fraction.

\runinhead{Multi-objective morphology control.}
For the morphology-control case (Supplementary Section~\ref{sec:si_case2}), the scalar objective $\mathcal{J}_{\mathcal{T}}$ in Eq.~\eqref{eq:general_task_objective} is replaced by a two-component objective vector since the goal is not to find one scalar optimum but to improve a Pareto set. Each feasible topology is assigned
\begin{equation}
\begin{aligned}
\max_{\mathbf{x}} \;
\mathbf{f}_{\mathrm{MO}}(\mathbf{x})
&=
\left(
f_{\mathrm{fractal}}(\mathbf{x}),
f_{\mathrm{size}}(\mathbf{x})
\right)^{\mathsf{T}}
=
\left(
D_{\Gamma}(\mathbf{x}),
w_{\min}(\mathbf{x})
\right)^{\mathsf{T}},\\
\textrm{s.t.}\qquad
\phi_{\mathrm{conn}}(\mathbf{x})&=1,\quad
V_f(\mathbf{x})=V_f^0.
\end{aligned}
\end{equation}
Here, $w_{\min}$ is the minimum feature size and $D_{\Gamma}$ is the boundary fractal dimension; their detailed computations are provided in Supplementary Section~\ref{sec:si_case2}. The subject-to terms enforce periodic connectivity and the prescribed volume fraction.

\runinhead{Property-targeted auxetic design.}
For the elastic property-targeting case (Supplementary Section~\ref{sec:si_case3}), each task specifies a target elasticity tensor $\mathbf{H}_{\mathrm{ES,target}}^{(\chi)}$ for a symmetry class $\chi$. In Eq.~\eqref{eq:general_task_objective}, the task objective is the tensor-matching score,
\begin{equation}
\begin{aligned}
\max_{\mathbf{x}} \;
r_{\mathrm{ES}}^{(\chi)}(\mathbf{x})
&=
-\operatorname{MAE}
\left(
\mathbf{H}_{\mathrm{ES}}(\mathbf{x}),
\mathbf{H}_{\mathrm{ES,target}}^{(\chi)}
\right),\\
\textrm{s.t.}\qquad
\phi_{\mathrm{conn}}(\mathbf{x})&=1,\quad
V_f(\mathbf{x})=V_f^0,
\end{aligned}
\end{equation}
Here, $\mathbf{H}_{\mathrm{ES}}(\mathbf{x})$ is the homogenized elasticity tensor computed from the generated topology, and $\operatorname{MAE}(\cdot,\cdot)$ denotes the mean absolute error over all tensor components. The subject-to terms enforce connectivity and the prescribed volume fraction. This formulation turns inverse property matching into the same generate--evaluate--select--fine-tune loop used by the other cases.

\runinhead{Frequency-domain vibration transmission design.}
For the vibration case (Supplementary Section~\ref{sec:si_case4}), the objective $\mathcal{J}_{\mathcal{T}}$ in Eq.~\eqref{eq:general_task_objective} is a function-targeted score defined on the transmission spectrum $T_{\mathbf{x}}(f)$ of each topology. Let $\Omega_{\mathrm{p},m}$ denote target pass-band intervals and $\Omega_{\mathrm{g},n}$ denote target gap-band intervals. The band-satisfaction score is
\begin{equation}
s_{\mathrm{FQ}}(\mathbf{x})
=
\sum_m \omega_{\mathrm{p},m}
\left\langle
\mathbf{1}\!\left[T_{\mathbf{x}}(f)\geq T_{\mathrm{th}}\right]
\right\rangle_{\Omega_{\mathrm{p},m}}
+
\sum_n \omega_{\mathrm{g},n}
\left\langle
\mathbf{1}\!\left[T_{\mathbf{x}}(f)<T_{\mathrm{th}}\right]
\right\rangle_{\Omega_{\mathrm{g},n}},
\end{equation}
where $\langle\cdot\rangle_{\Omega}$ denotes averaging over a frequency interval, $T_{\mathrm{th}}=-10$ dB is the threshold separating pass and gap behavior, and the weights $\omega_{\mathrm{p},m}$ and $\omega_{\mathrm{g},n}$ control the relative importance of the frequency regions. The optimization problem is therefore
\begin{equation}
\max_{\mathbf{x}} \; s_{\mathrm{FQ}}(\mathbf{x}),
\qquad
\textrm{s.t.}
\qquad
\phi_{\mathrm{v}}(\mathbf{x})=1,\quad
V_f(\mathbf{x})=V_f^0.
\end{equation}
The subject-to terms enforce the vertical-support indicator and the prescribed volume fraction. The vertical-support indicator $\phi_{\mathrm{v}}$ removes structures that are unsuitable for the transmission simulation, while the score itself only requires evaluated spectra and therefore remains compatible with black-box frequency-domain solvers.

\subsection{Interpretation of Distribution Steering}
\label{sec:methods_distribution_steering}
This section gives a formal interpretation of distribution steering in GenTO. The offline stage learns a reusable topology prior, while the online stage updates this prior through a KL-regularized task-driven distribution shift. Diffusion training provides a stable way to represent and update the topology distribution, including shifts beyond the original training support. For notation clarity, $\mathbf{z}$ denotes a vectorized SDF field in this theoretical discussion, and binary topologies are obtained through the binarization map $\mathcal{B}_{V_f}(\mathbf{z})$. 
\runinhead{Prior learning.}
Let $\mathcal{D}_{z}=\{\mathbf{z}^{(n)}\}_{n=1}^{N}$ denote the training set of SDF fields obtained from previously generated metamaterial structures, and let $p_0(\mathbf{z})$ denote the corresponding prior distribution. Writing
$p_0(\mathbf{z}) = \frac{e^{-\mathcal{E}_{\mathrm{prior}}(\mathbf{z})}}{Z_{\mathrm{prior}}}$ with intractable normalizing constant $Z_{\mathrm{prior}}$ motivates working with the score function instead:
\begin{equation}
    \nabla_{\mathbf{z}} \log p_0(\mathbf{z}) = -\nabla_{\mathbf{z}} \mathcal{E}_{\mathrm{prior}}(\mathbf{z}),
\end{equation}
where $\mathcal{E}_{\mathrm{prior}}(\mathbf{z})$ is an energy function and $Z_{\mathrm{prior}}$ is the normalizing constant. $\nabla_{\mathbf{z}} \log p_0(\mathbf{z})$ is the score function, namely the gradient of the log-density with respect to the SDF representation. Diffusion models learn this score function. In the forward process, the data distribution is gradually corrupted into a Gaussian prior over time $t \in [0, T]$ via a forward stochastic differential equation (SDE) $d\mathbf{z} = \mathbf{f}(\mathbf{z}, t) dt + g(t) d\mathbf{w}$, and generation corresponds to simulating the time-reversal of this system:
\begin{equation}
    d\mathbf{z} = [\mathbf{f}(\mathbf{z}, t) - g(t)^2 \nabla_{\mathbf{z}} \log p_t(\mathbf{z})] dt + g(t) d\bar{\mathbf{w}}.
\end{equation}
In the forward process, $\mathbf{f}(\mathbf{z}, t)$ is the deterministic drift, $g(t)$ controls the noise amplitude, and $d\mathbf{w}$ denotes Brownian noise; $d\bar{\mathbf{w}}$ denotes the corresponding reverse-time Brownian noise. The reverse-time dynamics recover the structural information gradually destroyed by forward noising.
In the reverse process, the score function $\nabla_{\mathbf{z}} \log p_t(\mathbf{z})$ serves as a restoring force, representing the stochastic trajectory down the energy landscape defined by the underlying data distribution toward states of lower energy (higher probability density). A neural network $\mathbf{s}_\theta(\mathbf{z}, t)$ is optimized to approximate the score function $\nabla_{\mathbf{z}} \log p_t(\mathbf{z})$:
\begin{equation}
    \mathbf{s}_\theta(\mathbf{z}, t) \approx \nabla_{\mathbf{z}} \log p_t(\mathbf{z}).
\end{equation}
The network $\mathbf{s}_\theta(\mathbf{z}, t)$ is vector-valued because it estimates a direction in SDF field space at each diffusion time. This prior-learning view underlies the transition from the categorized training families to the joint prior and the pretrained generator shown in Fig.~\ref{fig:summary_framework}a,c.

\runinhead{Task-Driven Generation via KL-Regularized Score Dynamics.}
With the binarized topology written as $\mathbf{x}=\mathcal{B}_{V_f}(\mathbf{z})$, let $r(\mathbf{x})$ denote its scalar design score (i.e., the objective function to be maximized). A naive approach is to steer the distribution by directly maximizing the expected score $\mathbb{E}[r(\mathbf{x})]$. However, score maximization alone does not control how far the updated distribution departs from the current prior. Distribution steering therefore requires a balance between physical exploitation and topology regularization. GenTO approaches this through an iterative optimization procedure, using KL divergence to regularize the updated distribution toward the current model. With $p_i(\mathbf{z})$ denoting the current model generation distribution at iteration $i$, the corresponding distribution-update objective is formulated as
\begin{equation}
    q_i^* =
    \arg\max_{q_i}\left\{
    \mathbb{E}_{\mathbf{z} \sim q_i}[r(\mathbf{x})]
    - \frac{1}{\beta} D_{KL}(q_i \| p_i) \right\},
\end{equation}
where $q_i$ denotes a candidate updated distribution at iteration $i$, $q_i^*$ denotes the corresponding optimal updated distribution, $p_i(\mathbf{z})$ denotes the current model generation distribution in SDF space, $r(\mathbf{x})$ denotes the task-specific scalar design score after binarization, $\mathbb{E}_{\mathbf{z}\sim q_i}[r(\mathbf{x})]$ is the mean task score under $q_i$, $\beta$ controls the trade-off between score improvement and regularization, and $D_{KL}(q_i \| p_i)$ measures the discrepancy between the updated distribution and the current model distribution. The analytical solution to this locally constrained problem yields the optimal target distribution:
\begin{equation}
    q_i^*(\mathbf{z}) = \frac{1}{Z_{i, \beta}} p_i(\mathbf{z}) \exp(\beta r(\mathbf{x})),
    \label{eq:target_score}
\end{equation}
where $Z_{i,\beta}$ is the normalizing constant that ensures $q_i^*$ integrates to one.
To achieve extreme performance, GenTO considers the limiting regime $\beta \to \infty$. In this regime, the smooth exponential weighting degenerates into a hard step function $\mathbf{1}[r(\mathbf{x}) \ge \tau_i]$, which corresponds to the heuristic thresholding operation used in our framework.

\runinhead{Time-Marginalized Score Matching and Singularity Resolution.}
Here we explain why GenTO fine-tunes the diffusion model instead of explicitly applying reward guidance during sampling.
Following Eq.~\eqref{eq:target_score}, the score function of the target distribution $q_i^*(\mathbf{z})$ can be expressed as:
\begin{equation}
    \nabla_{\mathbf{z}} \log q_i^*(\mathbf{z}) = \nabla_{\mathbf{z}} \log p_i(\mathbf{z}) + \beta \nabla_{\mathbf{z}} r(\mathbf{x}).
\end{equation}
In principle, this suggests steering the reverse process with an additional objective gradient. However, explicit sampling-time guidance would require evaluating or differentiating the physical objective on noisy intermediate fields $\mathbf{z}_{t}$, which is ill-defined for many black-box evaluators used here, such as thermal homogenization and vibration transmission analysis. The difficulty becomes more pronounced in the limiting threshold-selection regime: at $t=0$, the selected empirical distribution is concentrated on a finite set of high-performing samples, and the corresponding analytical drift becomes singular.
Rather than explicitly injecting this unstable guidance term into sampling, GenTO absorbs the distribution shift by fine-tuning the neural network $\mathbf{s}_{\theta_i}(\mathbf{z}, t)$ on the selected target distribution $q_i(\mathbf{z})$, where $\mathbf{s}_{\theta_0}(\mathbf{z}, t)$ represents the prior model.
Although the selected distribution is discrete at the clean-data level, diffusion training does not fit this singular distribution directly. The forward process smooths the target distribution into time-dependent Gaussian mixtures. The score learned by the diffusion model can then be expressed as:
\begin{equation}
    \nabla_{\mathbf{z}_{t}} \log q_{i, t}(\mathbf{z}_{t}) = \nabla_{\mathbf{z}_{t}} \log \int q_{i, 0}(\mathbf{z}_{0}) \mathcal{N}(\mathbf{z}_{t}; \sqrt{\bar{\alpha}_{t}}\mathbf{z}_{0}, (1-\bar{\alpha}_{t})\mathbf{I}) d\mathbf{z}_{0},
\end{equation}
where $q_{i,0}$ is the empirical target distribution at the clean-field level, $q_{i,t}$ is its noised marginal at diffusion time $t$, and $\mathcal{N}(\mathbf{z}_{t};\boldsymbol{\mu},\boldsymbol{\Sigma})$ denotes a Gaussian density with mean $\boldsymbol{\mu}$ and covariance $\boldsymbol{\Sigma}$. The target distribution is therefore transformed into a continuous Gaussian Mixture Model (GMM), that is, a weighted sum of Gaussian kernels centered at the selected samples. Consequently, for all $t > 0$, the previously singular target becomes a finite and well-behaved score field, allowing stable reverse-time sampling. In this way, the fine-tuned model can steer the distribution toward the user-defined target.

\runinhead{Out-of-Distribution (OOD) Extrapolation.}
Although the selected high-performance samples define a discrete empirical target at $t=0$, GenTO can still generate topologies whose responses lie outside the original training distribution, due to two complementary properties of diffusion models: support expansion induced by stochastic kernels, and cross-family interpolation in a smooth field representation.

\par\smallskip\noindent\textbf{1. Support expansion induced by stochastic diffusion kernels.}\enspace
Let $\mathcal{S}_{i,0}=\mathrm{supp}(q_{i,0})$ denote the support of the empirical target distribution at iteration $i$, that is, the set on which $q_{i,0}$ assigns non-zero probability mass. Although this target may be discrete at the clean-data level, diffusion sampling does not generate the final field through a deterministic projection onto $\mathcal{S}_{i,0}$. Instead, each reverse update is a stochastic Markov transition that can be written as
\begin{equation}
    \mathbf{z}_{k-1}=F_k(\mathbf{z}_k)+\sigma_k\boldsymbol{\zeta}_k, \qquad \boldsymbol{\zeta}_k\sim\mathcal{N}(0,\mathbf{I}),
\end{equation}
where $F_k$ is the neural-network-predicted denoising mean and $\sigma_k>0$. Equivalently, the transition kernel is
\begin{equation}
    q(\mathbf{z}_{k-1}\mid \mathbf{z}_k)=
    \mathcal{N}\!\left(\mathbf{z}_{k-1};F_k(\mathbf{z}_k),\sigma_k^2\mathbf{I}\right).
\end{equation}
Thus, even for a fixed $\mathbf{z}_k$, the next state is drawn from a non-degenerate Gaussian instead of a Dirac point. After marginalizing over $\mathbf{z}_k$, the distribution of $\mathbf{z}_{k-1}$ becomes a mixture of such Gaussian kernels. This stochastic reverse kernel prevents the generated SDF distribution from being confined to the finite support of the selected empirical samples. In this sense, diffusion introduces a built-in support-broadening mechanism, so the learned topology distribution can explore neighborhoods around the selected target set instead of only reproducing the selected samples.

\par\smallskip\noindent\textbf{2. Cross-family interpolation in SDF space.}\enspace
Now suppose the training distribution is composed of $K$ topology families:
\begin{equation}
    q_{i,0}(\mathbf{z}_{0})=\sum_{k=1}^{K}\pi_k q_{i,0}^{(k)}(\mathbf{z}_{0}), \qquad
    q_{i,t}(\mathbf{z}_{t})=\sum_{k=1}^{K}\pi_k q_{i,t}^{(k)}(\mathbf{z}_{t}),
\end{equation}
where each component $q_{i,0}^{(k)}$ is supported on one topology family, and $\pi_k$ is the weight of family $k$ in the mixture. Even if these families are disconnected at $t=0$, Gaussian smoothing makes their time-marginals overlap for $t>0$. Defining the posterior family weights
\begin{equation}
    \rho_k(\mathbf{z}_{t},t)=\frac{\pi_k q_{i,t}^{(k)}(\mathbf{z}_{t})}{\sum_{j=1}^{K}\pi_j q_{i,t}^{(j)}(\mathbf{z}_{t})},
\end{equation}
where $\rho_k(\mathbf{z}_{t},t)$ is the posterior probability that the noisy SDF sample $\mathbf{z}_{t}$ originates from family $k$. The posterior mean and the corresponding score can be written as
\begin{equation}
    \begin{aligned}
        \mathbb{E}[\mathbf{z}_{0}\mid \mathbf{z}_{t}] = \sum_{k=1}^{K}\rho_k(\mathbf{z}_{t},t)\mathbb{E}[\mathbf{z}_{0}\mid \mathbf{z}_{t},k], \\
    \nabla_{\mathbf{z}_{t}}\log q_{i,t}(\mathbf{z}_{t}) = \frac{\sqrt{\bar{\alpha}_{t}}\mathbb{E}[\mathbf{z}_{0}\mid \mathbf{z}_{t}]-\mathbf{z}_{t}}{1-\bar{\alpha}_{t}}.
    \end{aligned}
\end{equation}
Hence, in regions where several family tails overlap, the reverse vector field is not tied to a single observed family, but becomes a weighted interpolation of family-conditioned denoising directions. This mechanism fills the holes between disconnected topology families and provides a route to off-family generation. Because GenTO models SDF fields instead of binary images, this interpolation takes place in a smoother continuous representation in which averages remain meaningful. An example of this behavior is shown in Fig.~\ref{fig:summary_framework}b, which illustrates how linear interpolation in the SDF representation can generate new geometries from representative topologies. After thresholding the generated field back to a binary topology, the model can therefore produce geometries that are not confined to the finite set of categorized training families, but instead reflect a more general topology distribution learned through diffusion.

\subsection{Relation to Other Optimization Methods}
\label{sec:methods_other_optimizers}
GenTO is related to both gradient-based topology optimization \cite{bendsoe1988generating,sigmund200199} and population-based black-box optimization \cite{holland1975adaptation,goldberg1989genetic}, but differs from both in that the optimization variable is a generative distribution, not a single design or an explicit finite population.

\runinhead{Relation to gradient-based optimization.}
Conventional topology optimization and other gradient-based methods typically start from a single initial structure and improve it through local geometric updates driven by objective sensitivities. The optimization trajectory is therefore defined in design space, and the object being updated remains one design instance throughout the process. GenTO has an analogous iterative structure, but the role of the initial design is instead played by the pretrained prior distribution. Instead of moving one topology along a deterministic gradient path, GenTO repeatedly shifts probability mass from the current prior toward a task-relevant high-performance region. The result is thus an optimized geometry distribution from which the final design is selected, in addition to an optimized structure. In this sense, GenTO can be viewed as a distribution-level generalization of gradient-based topology optimization.

\runinhead{Relation to population-based black-box optimization.}
GenTO also bears a close relation to genetic algorithms and other population-based black-box methods. The pretrained prior plays a role analogous to an initial population, the task evaluation and filtering steps act as selection, and each fine-tuning round updates the generative model in a way that is analogous to population evolution. The main distinction lies in how exploration is achieved. In genetic algorithms, moving beyond the support of the current population typically relies on explicitly designed mutation operators. In GenTO, by contrast, exploration is built into the diffusion process itself: stochastic sampling naturally introduces variability during generation, allowing the model to explore structures beyond the currently dominant modes of the prior. Because this exploration occurs in a learned continuous field representation instead of through hand-crafted discrete mutations, the method can move beyond finite categorized families and progressively adapt toward broader topology distributions. Accordingly, GenTO may also be interpreted as a learned black-box optimizer in distribution space, where exploration is intrinsic to the stochastic generative dynamics.

\backmatter

\section*{Data Availability}
The data supporting the findings of this study are available via Google Drive at \url{https://drive.google.com/drive/folders/1iLriVSi7aBi89xtdFyk8-OWidYB0o3ys?usp=drive_link} and archived via Zenodo at \url{https://doi.org/10.5281/zenodo.20452173}. Source data are provided with this paper.

\section*{Code Availability}
The code developed in this study is available via GitHub at \url{https://github.com/hl4220/GenTO-Unified-generative-design-of-architected-metamaterials.git} and archived via Zenodo at \url{https://doi.org/10.5281/zenodo.20452336}.

\section*{Acknowledgments}
The authors have no acknowledgments to declare.

\section*{Author Contributions Statement}
H.L. and Y.M. conceived the study. H.L. developed the GenTO framework and performed the numerical experiments. Y.M. contributed to model development, data processing and analysis. M.L. and X.L. contributed to experimental work. J.B. contributed to simulations. X.L. and B.G. contributed to validation. Z.S.K., J.L., D.M. and M.H.A. contributed to interpretation and manuscript revision. L.W. and W.C. supervised the project. H.L., Y.M. and L.W. wrote the initial manuscript. All authors discussed the results and reviewed and edited the manuscript.

\section*{Competing Interests Statement}
The authors declare no competing interests.

\bibliography{references}

\clearpage
\appendix

\clearpage
\thispagestyle{empty}
\begin{center}
\vspace*{2.2cm}
{\Titlefont Steering topology distributions for unified\\
generative design of architected metamaterials\par}
\vspace{0.8cm}
{\large\bfseries Supplementary Information\par}
\end{center}

\clearpage
\section*{Contents}
\begingroup
\setlength{\parindent}{0pt}
\setlength{\parskip}{0.45em}

\textbf{\ref{sec:si_nomenclature}\quad Nomenclature}\hfill\pageref{sec:si_nomenclature}

\textbf{\ref{sec:si_method_details}\quad Details of the Proposed Method}\hfill\pageref{sec:si_method_details}

\hspace*{1.5em}\ref{sec:si_datasets}\quad Construction of the Pretraining Topology Datasets\hfill\pageref{sec:si_datasets}

\hspace*{1.5em}\ref{sec:si_model_architecture}\quad Diffusion Model and Generation Process\hfill\pageref{sec:si_model_architecture}

\hspace*{1.5em}\ref{sec:si_distance_field}\quad Signed-Distance Function Topology Representation\hfill\pageref{sec:si_distance_field}

\hspace*{1.5em}\ref{sec:si_learning_steering}\quad Learning and Steering the Topology Prior\hfill\pageref{sec:si_learning_steering}

\textbf{\ref{sec:si_case_details}\quad Details of Case Studies}\hfill\pageref{sec:si_case_details}

\hspace*{1.5em}\ref{sec:si_case1}\quad Thermal Conductivity Extremization\hfill\pageref{sec:si_case1}

\hspace*{1.5em}\ref{sec:si_case2}\quad Multi-Objective Morphology Control\hfill\pageref{sec:si_case2}

\hspace*{1.5em}\ref{sec:si_case3}\quad Property-Targeted Auxetic Design\hfill\pageref{sec:si_case3}

\hspace*{1.5em}\ref{sec:si_case4}\quad Frequency-Domain Vibration Transmission Design\hfill\pageref{sec:si_case4}

\textbf{\ref{sec:si_baselines}\quad Baseline Methods}\hfill\pageref{sec:si_baselines}

\hspace*{1.5em}\ref{sec:si_baseline_training}\quad Best in Training Sets\hfill\pageref{sec:si_baseline_training}

\hspace*{1.5em}\ref{sec:si_baseline_prior}\quad Best in Prior\hfill\pageref{sec:si_baseline_prior}

\hspace*{1.5em}\ref{sec:si_baseline_to}\quad Traditional Topology Optimization\hfill\pageref{sec:si_baseline_to}

\hspace*{1.5em}\ref{sec:si_baseline_ga}\quad Genetic Algorithm\hfill\pageref{sec:si_baseline_ga}

\textbf{\ref{sec:si_additional_results}\quad Additional Experimental Results}\hfill\pageref{sec:si_additional_results}

\hspace*{1.5em}\ref{sec:si_case1_additional}\quad Comparative Results for Thermal Conductivity Extremization\hfill\pageref{sec:si_case1_additional}

\hspace*{1.5em}\ref{sec:si_case2_additional}\quad Additional Results of Case 2\hfill\pageref{sec:si_case2_additional}

\hspace*{1.5em}\ref{sec:si_case3_additional}\quad Additional Results of Case 3\hfill\pageref{sec:si_case3_additional}

\hspace*{1.5em}\ref{sec:si_case3_experimental_setup}\quad Experimental Setup for Case 3\hfill\pageref{sec:si_case3_experimental_setup}

\hspace*{1.5em}\ref{sec:si_case4_additional}\quad Additional Results of Case 4\hfill\pageref{sec:si_case4_additional}

\textbf{\ref{sec:si_extended_discussions}\quad Extended Discussions}\hfill\pageref{sec:si_extended_discussions}

\hspace*{1.5em}\ref{sec:si_discussion_method}\quad Discussions on the Proposed Method\hfill\pageref{sec:si_discussion_method}

\hspace*{1.5em}\ref{sec:si_discussion_diffusion}\quad Discussions on the Diffusion Model\hfill\pageref{sec:si_discussion_diffusion}

\hspace*{1.5em}\ref{sec:si_discussion_cases}\quad Discussions on the Case Studies\hfill\pageref{sec:si_discussion_cases}
\endgroup

\clearpage

\setcounter{table}{0}
\renewcommand{\theHtable}{S\arabic{table}}
\setcounter{figure}{0}
\renewcommand{\theHfigure}{S\arabic{figure}}


\clearpage
\section{Nomenclature}
\label{sec:si_nomenclature}

\footnotesize
\begin{longtable}{@{}>{\raggedright\arraybackslash}p{0.20\textwidth}>{\raggedright\arraybackslash}p{0.25\textwidth}>{\raggedright\arraybackslash}p{0.48\textwidth}@{}}
\caption{Nomenclature used in the main text and Supplementary Information. The table lists the main variables used in the GenTO framework and in the four case studies; function definitions, operators, and algorithm-specific expressions are omitted for compactness.}
\label{tab:nomenclature}\\
\toprule
Category & Symbol & Meaning \\
\midrule
\endfirsthead
\toprule
Category & Symbol & Meaning \\
\midrule
\endhead
\midrule
\multicolumn{3}{r}{Continued on next page}\\
\endfoot
\bottomrule
\endlastfoot

\multicolumn{3}{@{}p{0.93\textwidth}}{\textit{General GenTO framework}}\\
Topology & $\mathbf{x}$, $x_{ij}$, $\mathbf{x}^{(n)}$, $\mathbf{x}^{\star}$ & Binary topology, pixel value, $n$-th training topology, and final selected topology. \\
SDF field & $\mathbf{z}$, $\mathbf{z}_{0}$, $\mathbf{z}_{t}$, $\mathbf{z}_{T}$, $\hat{\mathbf{z}}_{0}$ & Signed-distance-function field, clean field, noisy diffusion-time field, terminal noisy field, and denoised generated field. \\
SDF construction & $z_{\mathrm{m}}$, $z_{\mathrm{v}}$, $z_{\mathrm{raw}}$, $z_{\mathrm{norm}}$ & Material-side, void-side, raw, and normalized SDF variables. \\
Training sets & $\mathcal{D}_{x}$, $\mathcal{D}_{z}$ & Binary-topology and SDF-field training datasets. \\
Volume fraction & $V_f$, $V_f^0$ & Material volume fraction and prescribed target volume fraction. \\
Diffusion variables & $\theta$, $\theta_i$, $t$, $T$, $\beta_t$, $\alpha_t$, $\bar{\alpha}_t$, $\boldsymbol{\epsilon}$ & Model parameters, iteration-indexed model parameters, diffusion step variables, noise-schedule variables, and Gaussian noise. \\
Generative distributions & $p_0$, $p_i$, $\tilde{p}_i$, $p_{\mathcal{T}}^{\star}$, $p_{\mathrm{data}}$ & Initial prior, current generated distribution, filtered high-performing distribution, task-adapted distribution, and data distribution. \\
Distribution steering & $q_i$, $q_i^*$, $q_{i,0}$, $q_{i,t}$, $\Omega_i$, $Z_i$, $Z_{i,\beta}$ & Candidate and optimal updated distributions, clean and noised empirical target distributions, accepted subset, and normalizing constants. \\
Family-mixture variables & $\pi_k$, $\rho_k$ & Mixture weight and posterior family weight in the distribution-steering interpretation. \\
Steering variables & $r$, $\mathcal{J}_{\mathcal{T}}$, $\phi_{\mathcal{T}}$, $\phi_{\mathrm{task}}$, $B$, $N_{\mathrm{take}}$, $N_{\mathrm{iter}}$, $\tau_i$, $\beta$ & Design score, task-specific objective, task-specific and generic feasibility indicators, batch size, elite-set size, number of steering iterations, selection threshold, and KL trade-off parameter. \\
Topology families & GRF, CHPF, PR, TS & Gaussian Random Fields, Cahn-Hilliard Phase Fields, Porous Structures, and Truss-based Structures. \\

\multicolumn{3}{@{}l}{\textit{Case 1: Thermal conductivity extremization}}\\
Thermal objective & $\sigma_{\kappa}$, $\kappa$, $\bar{\kappa}_{\mathrm{eff}}$, $\Delta\kappa_{\mathrm{eff}}$, $\lambda_{\kappa}$ & Objective-direction sign, effective conductivity, mean conductivity, anisotropy difference, and anisotropy penalty weight. \\
Homogenized response & $\mathbf{H}_{\kappa}$, $H_{\kappa,ij}$ & Homogenized conductivity tensor and its components. \\
Thermal fields & $\Theta$, $\boldsymbol{e}$, $\boldsymbol{q}$ & Temperature, temperature-gradient field, and heat-flux field. \\
Material parameters & $k$, $k_{\mathrm{s}}$, $k_{\mathrm{v}}$, $k_0$ & Local, solid-phase, void-like, and reference conductivity. \\
Constraints & $\phi_{\mathrm{conn}}$, $w_{\min}$, $w_0$, $\delta_{\mathrm{v}}$ & Connectivity indicator, minimum feature size, prescribed lower bound, and local void-distance variable. \\

\multicolumn{3}{@{}l}{\textit{Case 2: Multi-objective morphology control}}\\
Morphology variables & $w_{\min}$, $D_{\Gamma}$ & Minimum feature size and boundary fractal dimension. \\
Box-counting variables & $s$, $N_{\Gamma}$, $C$, $R^2$ & Box size, counted boundary boxes, intercept, and fitting coefficient. \\
Feasibility & $\phi_{\mathrm{conn}}$, $V_f^0$ & Periodic-connectivity indicator and prescribed volume fraction. \\

\multicolumn{3}{@{}l}{\textit{Case 3: Property-targeted auxetic design}}\\
Symmetry and error & $\chi$, $\varepsilon_{\mathrm{ES}}^{(\chi)}$, $r_{\mathrm{ES}}^{(\chi)}$ & Target symmetry class, tensor-matching error, and corresponding steering score. \\
Elastic tensors & $\mathbf{H}_{\mathrm{ES}}$, $H_{\mathrm{ES},ij}$, $\mathbf{H}_{\mathrm{ES,target}}^{(\chi)}$ & Homogenized elasticity tensor, its Voigt components, and prescribed target tensor. \\
Elastic fields & $\boldsymbol{\sigma}$, $\boldsymbol{\varepsilon}$, $\overline{\boldsymbol{\varepsilon}}$, $\widetilde{\boldsymbol{\varepsilon}}$ & Local stress, local strain, macroscopic strain, and periodic strain fluctuation. \\
Material parameters & $\mathbb{C}$, $K_{\mathrm{bulk}}$, $\mu_{\mathrm{s}}$ & Local elastic tensor, solid-phase bulk modulus, and solid-phase shear modulus. \\
Geometry constraints & $\delta_{\mathrm{m}}$, $t_{\mathrm{skel}}$, $w_{\mathrm{skel}}$ & Skeleton distance, local branch thickness, and skeleton-based minimum feature size. \\
Reported properties & $E_1$, $E_2$, $\nu_{12}$, $G_{12}$ & Effective Young's moduli, Poisson's ratio, and shear modulus. \\

\multicolumn{3}{@{}l}{\textit{Case 4: Frequency-domain vibration transmission design}}\\
Frequency objective & $s_{\mathrm{FQ}}$, $\phi_{\mathrm{v}}$ & Weighted band-satisfaction score and vertical-support indicator. \\
Target bands & $\Omega_{\mathrm{p}}$, $\Omega_{\mathrm{g}}$, $T_{\mathbf{x}}$, $T_{\mathrm{th}}$ & Pass-band region, gap-band region, transmission spectrum, and classification threshold. \\
Frequency variables & $f$, $\omega$, $\omega_n$, $N_m$, $\omega_{\mathrm{p},m}$, $\omega_{\mathrm{g},n}$ & Frequency, circular frequency, retained modal frequency, number of retained modes, and pass/gap interval weights. \\
Material parameters & $E_{\mathrm{s}}$, $\nu_{\mathrm{s}}$, $\rho_{\mathrm{s}}$, $\eta_{\mathrm{damp}}$ & Solid Young's modulus, Poisson's ratio, density, and damping factor. \\
Dynamic response & $\mathbf{K}$, $\mathbf{M}$, $\mathbf{u}$, $\boldsymbol{\psi}_n$, $u_{\mathrm{out}}$ & Global stiffness matrix, mass matrix, displacement vector, modal shape, and output displacement response. \\

\multicolumn{3}{@{}l}{\textit{Comparison and baseline methods}}\\
Topology optimization & $\boldsymbol{\rho}$, $\rho_e$, $\rho_{\min}$, $p_{\mathrm{SIMP}}$, $\tau_{\mathrm{TO}}$ & Density field, element density, minimum density, SIMP exponent, and TO step size. \\
Genetic algorithm & $\mathbf{g}^{(\ell)}_m$, $N_p$, $\ell$, $\alpha$, $\alpha_{\min}$, $\alpha_{\max}$, $\sigma_{\mathrm{mut}}$ & Morphology field of individual $m$, population size, generation index, crossover mixing ratio and bounds, and mutation amplitude. \\
\end{longtable}
\normalsize

\clearpage
\section{Details of the Proposed Method}
\label{sec:si_method_details}

\begin{figure}[H]
\centering
\includegraphics[width=\textwidth]{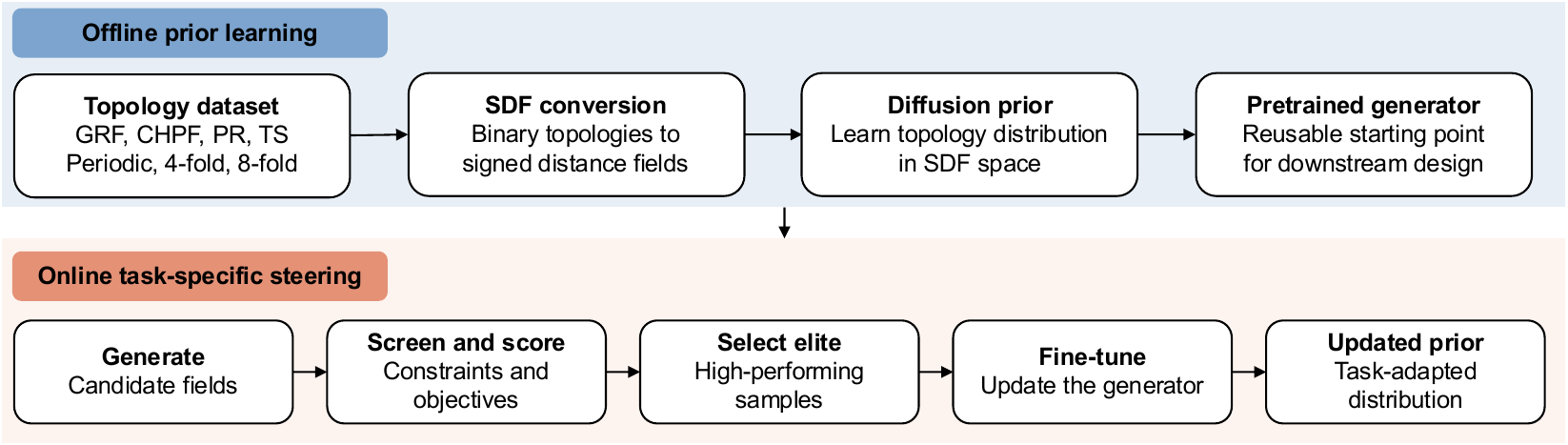}
\caption{\textbf{Overview of the GenTO framework.} In the offline stage, binary topologies from multiple topology families and symmetry types are converted into SDF fields and used to pretrain a diffusion model as a reusable topology prior. In the online stage, the pretrained prior is sampled, the generated fields are binarized at the prescribed volume fraction, and the resulting topologies are evaluated using task-specific constraints and performance measures. High-performing samples are retained to fine-tune the model, which progressively shifts the generated distribution toward the target task.}
\label{fig:SI_F1}
\end{figure}

GenTO starts with an offline prior-learning stage and then moves to an online task-specific steering stage (Supplementary Fig.~\ref{fig:SI_F1}). Instead of retraining a task-specific conditional generator for each new problem, GenTO first learns a reusable topology prior from a broad topology dataset and then iteratively steers this prior toward the high-performing region defined by the downstream task. The method thereby optimizes a topology distribution instead of directly searching for a single topology.

\runinhead{Offline prior learning.}
The offline stage constructs a training set that covers diverse topology families and symmetry types. In this work, the binary unit cells are collected from Gaussian random fields, Cahn-Hilliard phase fields, porous structures, and truss-based structures, and are organized into periodic, 4-fold, and 8-fold symmetry groups. Each binary topology is then converted into an SDF field. The continuous representation preserves the geometric information near the material boundary and provides a smoother learning space than a binary image. A diffusion model is pretrained on these SDF fields to learn a broad topology prior for each symmetry type.

\runinhead{Online distribution steering.}
For a given design task, the pretrained model is used as the initial generator. Candidate SDF fields are sampled from the current model, converted into binary topologies at the target volume fraction, and then evaluated by the task-specific constraints and performance measures. Feasible and high-performing samples are selected as elite samples and used to fine-tune the model for the next iteration. Repeating this generate--evaluate--select--fine-tune loop gradually shifts the generated distribution toward the target region while keeping the benefits of the learned prior. The same overall workflow is used in all case studies in this work; only the constraints, evaluation criteria, and selection rules are changed for different tasks. The following subsections provide detailed descriptions of the pretraining datasets, model architecture, prior-learning procedure, and online steering algorithm.

\subsection{Construction of the Pretraining Topology Datasets}
\label{sec:si_datasets}

The pretraining data are constructed from four topology families, namely Gaussian Random Fields (GRF), Cahn-Hilliard Phase Fields (CHPF), Porous Structures (PR), and Truss-based Structures (TS). For each family, three symmetry types are prepared: periodic, 4-fold symmetric, and 8-fold symmetric. The combination gives a total of 12 binary topology groups, as illustrated in Supplementary Fig.~\ref{fig:SI_F2}. All generated unit cells are stored on a $256\times256$ grid and saved as binary images, where the material phase is represented by 1 and the void phase is represented by 0. The aim of this dataset construction step is to provide a broad topology prior with different geometric motifs, feature scales, and symmetry patterns, not to represent only one morphology class.

\begin{figure}[t]
\centering
\includegraphics[width=\textwidth]{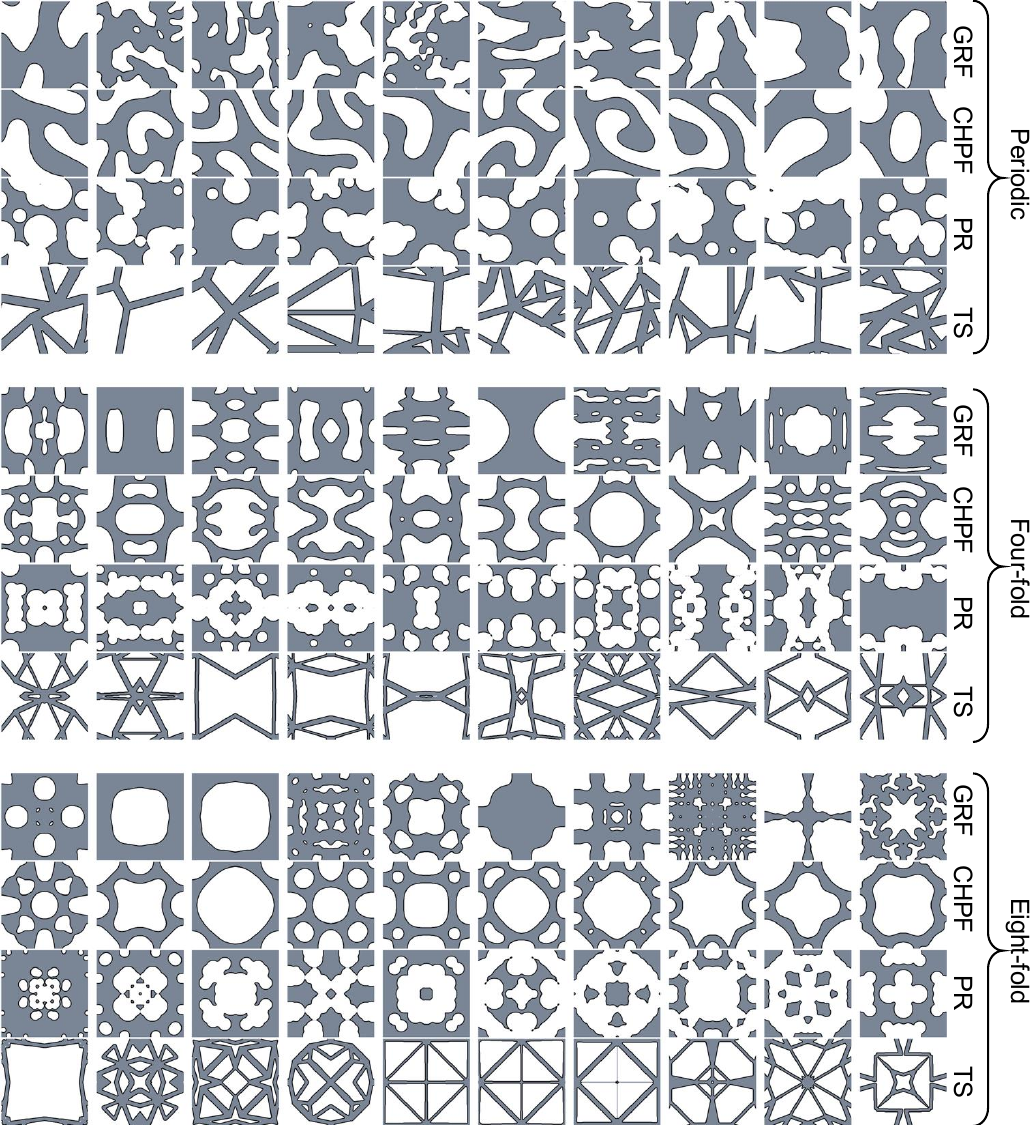}
\caption{\textbf{Examples of the binary topology datasets used for prior learning.} The 12 rows correspond to the 12 combinations of four topology families and three symmetry types. From top to bottom, the datasets are organized as periodic, 4-fold symmetric, and 8-fold symmetric groups. Within each symmetry group, the four topology families are Gaussian Random Fields, Cahn-Hilliard Phase Fields, Porous Structures, and Truss-based Structures. Each row shows 10 representative samples.}
\label{fig:SI_F2}
\end{figure}

\runinhead{Organization of the dataset.}
The 12 topology groups are formed by combining the four topology families with the three symmetry classes. The periodic group is used to learn a general periodic prior, while the two symmetric groups provide higher-symmetry topology priors for cases where stronger geometric regularity is desired.

\runinhead{Gaussian Random Field.}
The GRF family is generated in the frequency domain. For each sample, a random field is defined on a $256\times256$ grid. A Gaussian-type spectrum is first constructed as
\begin{equation}
S(\mathbf{k}) = \sigma_{\mathrm{GRF}}^{2} l_{c}^{2}\exp\!\left(-\frac{\|\mathbf{k}\|^{2} l_{c}^{2}}{2}\right),
\end{equation}
where $\sigma_{\mathrm{GRF}}=1.0$ is the field amplitude and $l_{c}$ is the correlation length. In the implementation, $l_{c}$ is randomly sampled from $[0.3,1.0]$. The physical lengths $L_{x}$ and $L_{y}$ are also random. In the periodic implementation, they are repeatedly sampled from $[1,11]$ when candidate fields are generated. In the 4-fold and 8-fold symmetric implementations, they are sampled from $[1,21]$ before the inner generation loop. Random Fourier coefficients are then multiplied by $\sqrt{S(\mathbf{k})}$, and an inverse FFT is used to obtain the spatial field. The field is thresholded at zero to produce a binary topology.

For the periodic GRF dataset, the Fourier field is generated directly and therefore satisfies periodicity naturally. For the 4-fold symmetric dataset, a random complex field is first mirrored between the four quadrants before being transformed back to the spatial domain. For the 8-fold symmetric dataset, the spatial field is further symmetrized by averaging it with its diagonal, anti-diagonal, horizontal, and vertical reflections. After thresholding, the resulting binary cell is screened for geometric admissibility. The binary cell is tiled into a $3\times3$ array, and the largest connected component is required to span the whole tiled domain in both directions. Additional checks are used to reject pathological cases with too many repeated disconnected components or overly dominant secondary components. A target upper bound on the material fraction is also imposed through a random threshold parameter sampled from $[0.5,0.9]$. If the complement of a candidate satisfies the screening conditions whereas the original candidate does not, the complement is retained instead.

\runinhead{Cahn-Hilliard Phase-Field.}
The CHPF family is generated by simulating phase separation with a spectral Cahn-Hilliard solver~\cite{cahn1958free}. The field is evolved on a $256\times256$ grid using the semi-implicit update
\begin{equation}
\hat{c}^{\,n+1}
=
\frac{\hat{c}^{\,n}-\Delta t\,K^{2}\widehat{\left((c^{n})^{3}-c^{n}\right)}}{1+\Delta t\,\epsilon_{\mathrm{CH}}^{2}K^{4}},
\end{equation}
where $c^{n}$ is the real-space phase field at time step $n$, $\hat{c}^{\,n}$ is its Fourier coefficient, $\widehat{(\cdot)}$ denotes the Fourier transform, $K^{2}=k_{x}^{2}+k_{y}^{2}$ is the squared Fourier wavenumber magnitude, and $K^{4}=(K^{2})^{2}$. The nonlinear term is evaluated explicitly at step $n$, while the fourth-order diffusion term is treated implicitly through the denominator. In the implementation, $\epsilon_{\mathrm{CH}}=3.0$, $\Delta t=8.0$, and the total number of time steps is 2000. The initial condition is a small random perturbation with amplitude $0.01$. For the periodic dataset, the characteristic domain scale is controlled by a random factor $DC1\in[0.6,1.3]$, with $L_{x}=256\times DC1$ and $L_{y}=256\times DC1$. For the 4-fold and 8-fold symmetric datasets, the same random factor is used but the physical domain lengths are doubled, namely $L_{x}=512\times DC1$ and $L_{y}=512\times DC1$.

After time evolution, the real-space field is tiled into a $3\times3$ array, smoothed with a Gaussian filter of width $\sigma_{\mathrm{CH}}=10$, and then cropped back to the central unit cell. The cropped field is thresholded at zero. In the periodic case, the thresholded cell is screened by the same type of connectivity test used in the GRF dataset. In the 4-fold symmetric case, the initial condition is first defined on one quarter of the domain and then mirrored to form a 4-fold symmetric full field. After evolution and thresholding, only the upper-left quadrant is examined. The largest connected component of either the solid phase or the void phase is selected if it spans all four boundaries of the quadrant, and this quadrant is then mirrored back to form the final $256\times256$ unit cell. In the 8-fold symmetric case, one additional step is introduced: the quarter-domain initial condition is first averaged with its transpose to enforce diagonal symmetry, and the accepted quadrant is again symmetrized with respect to the diagonal before the 4-fold reconstruction. In this way, the final sample has both 4-fold and diagonal symmetry, which gives the 8-fold symmetric dataset.

\runinhead{Porous Structures.}
The porous family is generated by carving circular holes from an initially solid domain. For the periodic porous dataset, the unit cell size is $256\times256$. The number of holes is randomly sampled from 4 to 20, and each hole radius is sampled from 10 to 50 pixels. Hole centers are sampled uniformly over the cell. Periodic boundary conditions are enforced directly in the distance calculation by using wrapped distances in both coordinate directions. As a result, pores can pass through one boundary and continue from the opposite boundary without geometric discontinuity.

For the 4-fold symmetric porous dataset, the domain is again $256\times256$, but the holes are generated only in one quadrant and then copied to the remaining three quadrants by mirror operations. In this case, the number of holes is sampled from 5 to 15 and the radius range is 10 to 30 pixels. For the 8-fold symmetric porous dataset, holes are generated only in a one-eighth wedge between the horizontal axis and the diagonal. The number of holes is also sampled from 5 to 15, while the radius range is reduced to 7 to 15 pixels. Each sampled hole is copied to eight symmetry-related positions, including both rotational and diagonal mirrors. Compared with the GRF and CHPF datasets, the porous dataset provides more explicit hole-dominated morphologies and enriches the diversity of void-centric topological patterns in the prior.

\runinhead{Truss-based Structures.}
The truss family is generated by first sampling a sparse set of nodes and then converting a random graph into a binary image. The basic graph is constructed from a Delaunay triangulation of the sampled nodes. Candidate edges are then activated or deactivated randomly, assigned random thicknesses, rasterized onto an image grid, and finally screened for connectivity.

For the periodic truss dataset, the base cell size is $256\times256$. The number of interior nodes is randomly sampled from 2 to 10, with a minimum nodal spacing of $L/4=64$ pixels. To enforce periodicity, the sampled points are copied into a $3\times3$ tiled arrangement before triangulation. After triangulation, edges that correspond to the same periodic connection are identified and assigned the same thickness. The nominal edge width is 15 pixels, and the actual width is sampled from 8 to 25 pixels. Each edge is then removed with probability 0.5. After activation, isolated edges are deleted by removing edges attached to nodes of degree one. The remaining graph is rasterized, and the final image is accepted only if it is connected and touches all four boundaries of the unit cell.

For the 4-fold symmetric truss dataset, a smaller base domain of size $128\times128$ is used first. The number of interior nodes is sampled from 1 to 4, the minimum nodal spacing is $128/4=32$ pixels, and the nominal edge width is $\mathrm{round}(128/18)=7$ pixels. The sampled graph is rasterized on the $128\times128$ base domain, and the final topology is obtained by assembling four mirrored copies into a $256\times256$ 4-fold symmetric unit cell. The edge width is randomly sampled from 4 to 12 pixels, and each edge is again removed with probability 0.5 before isolated-edge cleanup.

For the 8-fold symmetric truss dataset, the generation starts from a triangular wedge in which the sampled coordinates satisfy $y\le x$. These points are mirrored across the diagonal to obtain one quadrant with diagonal symmetry. Periodic copies of this quadrant are then used for Delaunay triangulation. After the triangulation, diagonally mirrored edge pairs are identified and constrained to share the same activation state and thickness. Isolated-edge removal is also carried out in a symmetry-preserving manner, so that mirrored edges are deleted together. The quadrant image is then symmetrized again by averaging it with its transpose, which corrects small asymmetries caused by rasterization, and is finally assembled into a $256\times256$ unit cell through 4-fold mirroring. The family provides lattice-like and frame-like morphologies that are complementary to the smoother patterns generated by GRF, CHPF, and porous sampling.

\runinhead{Scope and extensibility of the dataset.}
The four topology families used here were not chosen to claim that they exhaust all possible classes of architected topologies. Rather, they were selected because they summarize four structure forms that are currently common in the literature and in practical design studies: random-field-like morphologies, phase-separation-like bicontinuous morphologies, hole-dominated porous morphologies, and lattice- or frame-like morphologies. These four families already cover a wide range of geometric characteristics, including smooth and sharp boundaries, connected and disconnected void patterns, distributed and localized features, and both continuum-like and skeleton-like structures. For the purpose of learning a reusable topology prior in this work, this level of diversity is sufficient.

An important point is that the pretrained prior is not expected to merely memorize these 12 topology groups. The prior is learned from their joint distribution in the representation space, and the subsequent steering process further shifts this learned distribution toward task-relevant regions. As a result, the generated topologies are not restricted to exact replicas of the training families, and can exhibit new combinations of geometric traits that are not explicitly present in any one dataset family. These observations motivate combining several topology families in the offline stage instead of training on a single morphology class only.

The framework is also open to extension. If a user is interested in other topology classes, such as application-specific cellular layouts, fabrication-constrained motifs, or structures collected from a separate simulation pipeline, these additional samples can be incorporated into the offline dataset and used to enlarge the prior. If the target application is strongly specialized, the user may also construct and train a topology prior only for that problem setting. Accordingly, the dataset used in this work should be viewed as one practical and sufficiently rich choice for demonstrating the GenTO framework, not as a fixed or exclusive definition of the topology prior.

\subsection{Diffusion Model and Generation Process}
\label{sec:si_model_architecture}

The generative model used in this work is a diffusion U-Net. Although earlier versions of the workflow included a conditional naming convention, the active model used here is effectively unconditional. The network takes only a noisy topology field and the diffusion time step as input, and does not use task labels or property conditions during prior pretraining. The task adaptation is handled later by the steering loop described in Supplementary Section~\ref{sec:si_learning_steering}.

\begin{figure}[H]
\centering
\includegraphics[width=\textwidth]{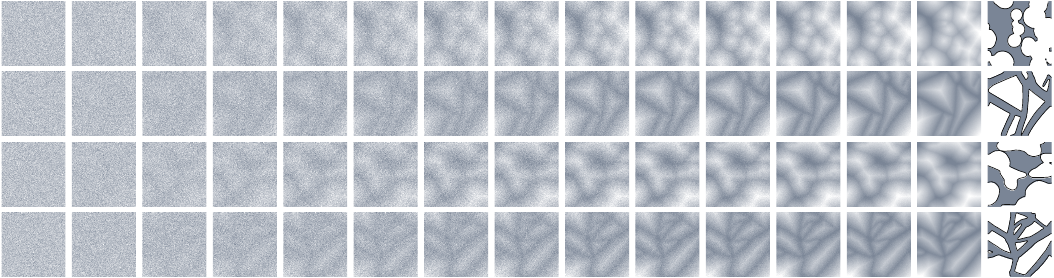}
\caption{\textbf{Progressive diffusion generation of topology fields.} Each row shows one DDPM sampling trajectory from Gaussian noise to a generated topology. Intermediate frames are sampled non-uniformly, with more frames near the later denoising stages where structural features become clear. The final column is binarized at the prescribed volume fraction.}
\label{fig:SI_F3}
\end{figure}

\runinhead{Input and output format.}
The network operates on single-channel continuous topology fields of size $256\times256$. The input tensor has shape $B\times1\times256\times256$, where $B$ is the batch size, and the output has the same spatial resolution and channel number. During training, the model is used to predict the clean field $\mathbf{z}_{0}$ from a noisy field $\mathbf{z}_{t}$. During generation, the same model is used repeatedly in the reverse diffusion process to map Gaussian noise back to a topology field.

\runinhead{Time embedding branch.}
The diffusion step is encoded by a small multilayer perceptron. The scalar time step $t$ is first converted into a one-dimensional floating-point input, then passed through two fully connected layers with one GELU activation in between, and finally mapped to a 256-dimensional time embedding. The embedding is injected into each residual block through a learned linear projection, allowing the network to modulate its behavior according to the current noise level while keeping the spatial backbone unchanged.

\runinhead{Encoder-decoder backbone.}
The main spatial backbone is a three-level U-Net. The input field first passes through an initial $3\times3$ convolution with circular padding that maps the single input channel to 64 feature channels. The use of circular padding is important here because the topologies are periodic unit cells, and wrap-around padding avoids introducing artificial boundary effects. The encoder then contains three downsampling stages. The channel width increases from 64 to 128, then to 256, and finally to 512, while the spatial resolution decreases from $256\times256$ to $128\times128$, then to $64\times64$, and finally to $32\times32$. Each downsampling stage contains one residual block followed by a stride-2 convolution with kernel size 4 and padding 1. At the bottleneck, two residual blocks with 512 channels are used to further process the latent representation before decoding.

The decoder mirrors the encoder through three upsampling stages. It gradually upsamples the latent field back to the original resolution, reduces the channel width from 512 to 256, then to 128, and finally to 64, and uses skip connections from the encoder at the matching resolutions. Each upsampling stage first uses a transposed convolution to double the spatial resolution, then concatenates the corresponding encoder feature map, and finally applies a residual block for feature refinement. After the last upsampling stage, an output head consisting of a $3\times3$ convolution from 64 to 32 channels, one GELU activation, and a final $3\times3$ convolution from 32 channels back to a single channel produces the denoised topology field.

\runinhead{Residual blocks and normalization.}
Each residual block contains two $3\times3$ convolutions, Group Normalization, GELU activation, and one shortcut branch. If the input and output channel numbers are different, a $1\times1$ convolution is used in the shortcut; otherwise an identity shortcut is used. In each block, the time embedding is projected to the current channel width and added to the intermediate feature map after the first normalization step, so that the block is explicitly aware of the diffusion time. GroupNorm uses 8 groups for all channel widths. The block design keeps the network simple, stable, and lightweight enough for repeated fine-tuning during the online steering stage.

\runinhead{Model simplicity.}
The current model is intentionally simple. It does not use attention layers, and task conditions are not injected into the network during prior pretraining. The choice is consistent with the main idea of GenTO: the offline model is trained as a reusable topology prior, while task adaptation is performed later by distribution steering instead of by conditioning the generator directly on each task.

\subsection{Signed-Distance Function Topology Representation}
\label{sec:si_distance_field}

Before diffusion pretraining, each binary topology is converted into a continuous SDF representation. Supplementary Fig.~\ref{fig:SI_F4} is obtained by applying this SDF processing to the binary topology examples shown in Supplementary Fig.~\ref{fig:SI_F2}. Although the main text gives this representation in a compact form, the actual processing contains several steps that are worth stating explicitly in what follows.

\begin{figure}[t]
\centering
\includegraphics[width=\textwidth]{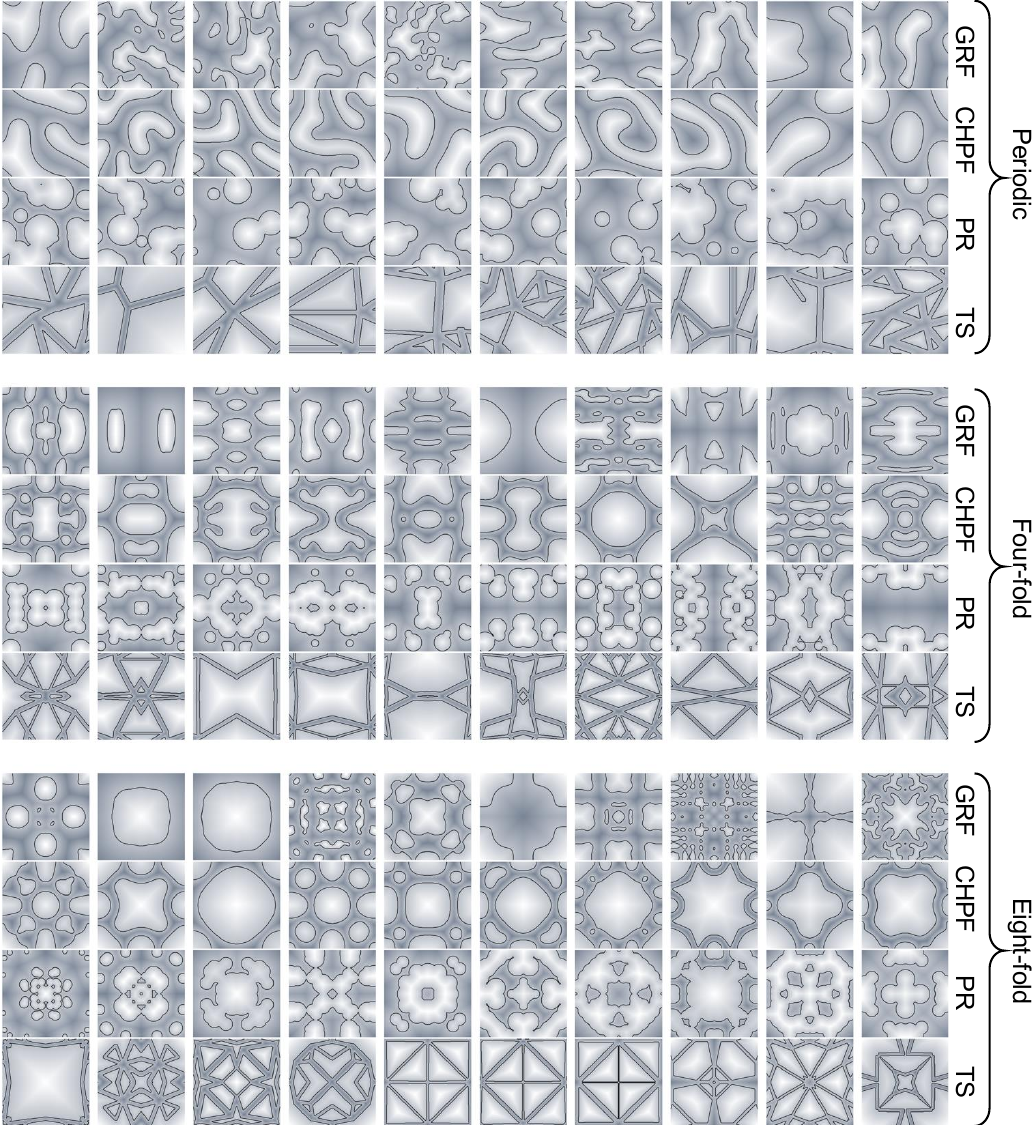}
\caption{\textbf{SDF continuous representations.} 
The figure is obtained by applying the SDF processing to the binary samples in Supplementary Fig.~\ref{fig:SI_F2}. In the SDF convention used here, positive values lie in the material phase and negative values lie in the void phase. Darker regions indicate points deeper inside the material phase after the visual remapping used for display.}
\label{fig:SI_F4}
\end{figure}

\runinhead{SDF construction.}
Let $\mathbf{x}\in\{0,1\}^{H\times W}$ denote a binary topology. For each pixel, two unsigned distance maps are first evaluated: $z_{\mathrm{m}}$ measures the distance to the nearest material--void interface inside the material phase and is zero in the void phase, while $z_{\mathrm{v}}$ measures the corresponding distance inside the void phase and is zero in the material phase. In the Python implementation, these unsigned maps are computed with a Euclidean distance transform (EDT) routine. The raw SDF field is written as
\begin{equation}
z_{\mathrm{raw}}
=
z_{\mathrm{m}}-z_{\mathrm{v}}.
\end{equation}
In this way, $z_{\mathrm{raw}}>0$ in the material phase and $z_{\mathrm{raw}}<0$ in the void phase, while the zero level set coincides exactly with the material--void interface.

\runinhead{Normalization and smoothing.}
After the raw SDF field is computed, the field is smoothed and normalized. A Gaussian filter with width $\sigma_{\mathrm{SDF}}=1$ is applied under periodic boundary treatment:
\begin{equation}
\bar{z}_{\mathrm{raw}}
=
\mathcal{G}_{\sigma_{\mathrm{SDF}}}^{\mathrm{per}}(z_{\mathrm{raw}}).
\end{equation}
The periodic treatment reflects the unit-cell setting of the topology images. It prevents artificial edge effects during smoothing by treating opposite boundaries as connected. The positive and negative parts are then normalized separately:
\begin{equation}
\begin{aligned}
m_{+}
&=
\max\left\{\bar{z}_{\mathrm{raw}}(i,j):\bar{z}_{\mathrm{raw}}(i,j)>0\right\},\\
m_{-}
&=
\max\left\{-\bar{z}_{\mathrm{raw}}(i,j):\bar{z}_{\mathrm{raw}}(i,j)<0\right\}.
\end{aligned}
\end{equation}
\begin{equation}
z_{\mathrm{norm}}(i,j)
=
\begin{cases}
\dfrac{\bar{z}_{\mathrm{raw}}(i,j)}{m_{+}}, & \bar{z}_{\mathrm{raw}}(i,j)>0,\\[8pt]
\dfrac{\bar{z}_{\mathrm{raw}}(i,j)}{m_{-}}, & \bar{z}_{\mathrm{raw}}(i,j)<0,\\[8pt]
0, & \bar{z}_{\mathrm{raw}}(i,j)=0.
\end{cases}
\end{equation}
Finally, the continuous field used for learning is written as
\begin{equation}
\mathbf{z}=z_{\mathrm{norm}}.
\end{equation}
The normalization preserves the sign convention used in the main text: positive values lie in the material phase and negative values lie in the void phase.

\runinhead{Why the SDF representation is used.}
The main reason for using this SDF representation is to replace the discontinuous binary image with a continuous geometry field that carries distance-to-boundary information. In a binary topology image, a small interface displacement causes a jump between 0 and 1 at many pixels. In contrast, the SDF field varies smoothly away from the interface, which makes the representation more stable under noise perturbation and denoising. The smoother representation is useful for both prior learning and later sampling.

The SDF field also carries explicit geometric meaning. Its magnitude measures the local distance to the interface, i.e. thin members, narrow void channels, and thick regions are encoded in a physically interpretable way. Since the zero contour corresponds exactly to the boundary, the generated field can be converted back to a topology without losing the main geometric information. In addition, smooth interpolation between topology patterns is easier in the SDF space than in the binary image space, which is one reason why the learned prior can move between different morphological families instead of staying locked to one discrete pattern class.

\runinhead{Relation to later binarization.}
Another practical advantage is that the SDF field works well with the volume-fraction-controlled binarization used in the online stage. Once a field $\hat{\mathbf{z}}_{0}$ is generated, the final binary topology is obtained by thresholding it at a sample-wise quantile. The continuity and ordering of the SDF field make this thresholding operation stable and allow the prescribed volume fraction to be matched exactly. The property is important in the case studies, where the volume fraction is fixed before physics evaluation.

\subsection{Learning and Steering the Topology Prior}
\label{sec:si_learning_steering}

After the binary topologies are converted into continuous SDF fields, GenTO learns a reusable prior distribution and then steers this prior toward task-specific high-performing regions. The workflow consists of three parts: diffusion-based prior learning, volume-fraction-controlled conversion from continuous fields to binary topologies, and iterative task-driven updating of the current generative distribution.

\runinhead{Forward diffusion and training objective.}
The diffusion process uses 100 time steps. The variance schedule is defined as
\begin{equation}
\beta_{t}
=
\left[
\mathrm{linspace}\!\left(\sqrt{10^{-4}},\sqrt{0.5},100\right)
\right]^{2},
\qquad t=1,\dots,100.
\end{equation}
The corresponding
\begin{equation}
\alpha_{t}=1-\beta_{t},
\qquad
\bar{\alpha}_{t}=\prod_{s=1}^{t}\alpha_{s}
\end{equation}
are stored in the diffusion pipeline and used in both training and sampling. For a clean field $\mathbf{z}_{0}$, the noisy field at step $t$ is formed as
\begin{equation}
\mathbf{z}_{t}
=
\sqrt{\bar{\alpha}_{t}}\,\mathbf{z}_{0}
+
\sqrt{1-\bar{\alpha}_{t}}\,\boldsymbol{\epsilon},
\qquad
\boldsymbol{\epsilon}\sim\mathcal{N}(\mathbf{0},\mathbf{I}).
\end{equation}
Two standard parameterizations can be used in diffusion training, namely $\boldsymbol{\epsilon}$-prediction and $\mathbf{z}_{0}$-prediction. In this work, the $\mathbf{z}_{0}$-prediction mode is used, that is,
\begin{equation}
\mathcal{L}_{\mathrm{diff}}
=
\frac{1}{256\times256}
\left\|
f_{\theta}(\mathbf{z}_{t},t)-\mathbf{z}_{0}
\right\|_{2}^{2}.
\end{equation}
The same objective is used both for the offline pretraining stage and for the later task-specific fine-tuning stage.

\runinhead{Symmetry-specific pretrained priors.}
In practice, separate pretrained priors are used for different symmetry classes. The prior is not trained as one single universal model across all symmetries. Instead, each symmetry type has its own reusable prior, and the downstream task chooses the one that matches its design space. The choice is consistent with the dataset construction in Supplementary Section~\ref{sec:si_datasets}, where the topologies are organized into periodic, 4-fold symmetric, and 8-fold symmetric groups.

\runinhead{Sampling from the current prior.}
Given the current model parameters, candidate fields are generated by starting from Gaussian noise and applying the reverse diffusion process. Two sampling modes are available. One is a standard stochastic reverse process, in which fresh Gaussian noise is injected at intermediate steps according to the variance term of the reverse transition. In the current workflow, this standard stochastic reverse process is used for sample collection during the steering stage, and an additional scalar parameter controls the magnitude of the stochastic term in the reverse step and thus the exploration strength during sampling. The other is a DDIM-style deterministic sampler, which uses 10 reduced sampling steps and is mainly used for generating samples from saved checkpoints after optimization.

\runinhead{Volume-fraction-controlled binarization.}
The generated outputs are continuous fields, but the physical evaluation functions operate on binary topologies. For each generated sample, all pixel values are flattened and a sample-wise quantile threshold is computed:
\begin{equation}
\tau
=
q_{1-V_{f}}(\hat{\mathbf{z}}_{0}),
\end{equation}
where $V_{f}$ is the prescribed material volume fraction. The binary topology is then obtained by
\begin{equation}
\mathbf{x}
=
\mathbf{1}\!\left[\hat{\mathbf{z}}_{0}>\tau\right].
\end{equation}
The operation ensures that every evaluated sample matches the target volume fraction exactly. The implementation detail separates the geometry generation problem from the volume-fraction constraint and makes the steering loop easier to stabilize.

\runinhead{Single-objective steering loop.}
The single-objective case studies follow a common loop. At iteration $i$, the current model first generates candidate continuous fields. These are binarized at the prescribed volume fraction, filtered by the task-specific feasibility conditions, and then scored by the task-specific objective function. The process is repeated until enough valid samples have been collected. The target number of valid samples in each iteration is set to twice the elite-set size:
\begin{equation}
N_{\mathrm{valid,target}} = 2N_{\mathrm{take}}.
\end{equation}
The oversampling step helps maintain selection pressure while avoiding the case where the valid set is too small.

After the valid samples of the current iteration are obtained, they are merged with the elite pool retained from the previous iteration. The combined candidate set is then sorted by score, and the top $N_{\mathrm{take}}$ samples are kept as the new elite set. If $\mathcal{E}_{i-1}$ denotes the previous elite pool and $\mathcal{V}_{i}$ denotes the current valid set, then the update can be written as
\begin{equation}
\mathcal{C}_{i}
=
\mathcal{E}_{i-1}\cup\mathcal{V}_{i},
\qquad
\mathcal{E}_{i}
=
\operatorname{Top}_{N_{\mathrm{take}}}\!\left(\mathcal{C}_{i}\right).
\end{equation}
The best sample and the mean elite score are recorded after each iteration. The model is then fine-tuned on the elite continuous fields using the same diffusion loss as above. In the case studies, this fine-tuning is carried out by ADAM with task-specific learning rates, batch sizes, and epoch numbers.

Algorithm~\ref{alg:si_single_objective_steering} summarizes the steering procedure used in the single-objective setting.

\begin{algorithm}[H]
\caption{Single-objective distribution steering}
\label{alg:si_single_objective_steering}
\begin{algorithmic}[1]
\Require prior $f_{\theta_0}$, volume fraction $V_f$, feasibility test, score, $N_{\mathrm{take}}$, batch size $B$, iterations $N_{\mathrm{iter}}$
\State $\mathcal{E}_{0}\leftarrow \varnothing$
\For{$i=1$ to $N_{\mathrm{iter}}$}
    \State $\mathcal{V}_{i}\leftarrow \varnothing$
    \While{fewer than $2N_{\mathrm{take}}$ valid samples have been collected}
        \State sample $\{\hat{\mathbf{z}}_{0}^{(b)}\}_{b=1}^{B}$ from $f_{\theta_{i-1}}$
        \State binarize the batch at $V_f$ to obtain candidate topologies $\{\mathbf{x}^{(b)}\}$
        \State keep feasible candidates and evaluate their scores $\{s^{(b)}\}$
        \State add the valid field--score pairs to $\mathcal{V}_{i}$
    \EndWhile
    \State $\mathcal{C}_{i}\leftarrow \mathcal{E}_{i-1}\cup \mathcal{V}_{i}$
    \State $\mathcal{E}_{i}\leftarrow \operatorname{Top}_{N_{\mathrm{take}}}(\mathcal{C}_{i})$
    \State $f_{\theta_i}\leftarrow$ fine-tune $f_{\theta_{i-1}}$ on the elite fields in $\mathcal{E}_{i}$
\EndFor
\State \Return $f_{\theta_{N_{\mathrm{iter}}}}$ and $\mathcal{E}_{N_{\mathrm{iter}}}$
\end{algorithmic}
\end{algorithm}

\noindent
Here, $f_{\theta_0}$ is the pretrained topology prior, $f_{\theta_{i-1}}$ is the current prior at steering iteration $i$, $V_f$ is the prescribed material volume fraction, $N_{\mathrm{take}}$ is the elite-set size, $B$ is the generation batch size, and $N_{\mathrm{iter}}$ is the total number of steering iterations. $\hat{\mathbf{z}}_{0}^{(b)}$ denotes the $b$-th generated continuous field, $\mathbf{x}^{(b)}$ is its binarized topology, $s^{(b)}$ is the corresponding objective score, $\mathcal{V}_{i}$ is the valid field--score set collected at iteration $i$, $\mathcal{E}_{i}$ is the elite set retained after iteration $i$, and $\mathcal{C}_{i}$ is the merged candidate set formed from the current valid samples and the previous elite pool.

\runinhead{Multi-objective steering and Pareto layers.}
For multi-objective tasks, the sample-collection stage is similar, but the scores are stored as multi-component objective values. Instead of sorting the candidates by one scalar value, the combined candidate set is filtered by Pareto dominance. Non-dominated layers are repeatedly extracted until the desired elite-set size is reached. If the total number of samples in the current Pareto layer exceeds the remaining quota, a random subset of that layer is retained. The design allows the elite set to preserve several competing design directions instead of collapsing too early onto one scalarized objective. The first Pareto front is also saved separately for visualization in the multi-objective case study. Algorithm~\ref{alg:si_multi_objective_steering} summarizes the multi-objective steering procedure.

\begin{algorithm}[H]
\caption{Multi-objective distribution steering with Pareto layers}
\label{alg:si_multi_objective_steering}
\begin{algorithmic}[1]
\Require prior $f_{\theta_0}$, volume fraction $V_f$, optional task-specific transform, feasibility test, objective vector, $N_{\mathrm{take}}$, batch size $B$, iterations $N_{\mathrm{iter}}$
\State $\mathcal{E}_{0}\leftarrow \varnothing$
\For{$i=1$ to $N_{\mathrm{iter}}$}
    \State $\mathcal{V}_{i}\leftarrow \varnothing$
    \While{fewer than $2N_{\mathrm{take}}$ valid samples have been collected}
        \State sample $\{\hat{\mathbf{z}}_{0}^{(b)}\}_{b=1}^{B}$ from $f_{\theta_{i-1}}$
        \State apply the optional transform if needed and binarize the batch at $V_f$ to obtain $\{\mathbf{x}^{(b)}\}$
        \State keep feasible candidates and evaluate their objective vectors $\{\mathbf{s}^{(b)}\}$
        \State add the valid field--score pairs to $\mathcal{V}_{i}$
    \EndWhile
    \State $\mathcal{C}_{i}\leftarrow \mathcal{E}_{i-1}\cup \mathcal{V}_{i}$
    \State initialize $\mathcal{E}_{i}\leftarrow \varnothing$
    \While{the elite set is not full and candidates remain}
        \State extract the current Pareto layer $\mathcal{L}_{\mathrm{P}}$ from $\mathcal{C}_{i}$
        \If{the whole layer can be accepted}
            \State $\mathcal{E}_{i}\leftarrow \mathcal{E}_{i}\cup\mathcal{L}_{\mathrm{P}}$
        \Else
            \State randomly select the required subset $\mathcal{L}_{\mathrm{sub}}\subset\mathcal{L}_{\mathrm{P}}$
            \State $\mathcal{E}_{i}\leftarrow \mathcal{E}_{i}\cup\mathcal{L}_{\mathrm{sub}}$
        \EndIf
        \State remove $\mathcal{L}_{\mathrm{P}}$ from $\mathcal{C}_{i}$
    \EndWhile
    \State $f_{\theta_i}\leftarrow$ fine-tune $f_{\theta_{i-1}}$ on the elite fields in $\mathcal{E}_{i}$
\EndFor
\State \Return $f_{\theta_{N_{\mathrm{iter}}}}$, $\mathcal{E}_{N_{\mathrm{iter}}}$, and the first Pareto front
\end{algorithmic}
\end{algorithm}

\noindent
Here, $f_{\theta_0}$, $f_{\theta_{i-1}}$, $V_f$, $N_{\mathrm{take}}$, $B$, $\mathcal{V}_{i}$, $\mathcal{E}_{i}$, and $\mathcal{C}_{i}$ have the same meanings as in Algorithm~\ref{alg:si_single_objective_steering}. $\hat{\mathbf{z}}_{0}^{(b)}$ denotes the $b$-th generated continuous field, $\mathbf{x}^{(b)}$ is the topology used for evaluation after optional task-specific transformation and binarization, and $\mathbf{s}^{(b)}$ is the corresponding multi-objective score vector. $\mathcal{L}_{\mathrm{P}}$ denotes the current Pareto layer extracted from the candidate set.

\runinhead{Task-specific transforms and evaluation.}
The steering framework is general because the same outer loop is reused across tasks, while the task-specific feasibility constraints and objective definitions are replaced. In the periodic multi-objective morphology setting used in this work, the generated fields are binarized and evaluated directly under the periodic-connectivity constraint. If a downstream task requires a specific symmetry projection, that projection can be applied before evaluation while the model is still fine-tuned on the raw generated continuous fields. The distinction is important: the model is updated in the learned representation space, not on the already binarized topologies used in the physical evaluation.

\runinhead{Why this loop steers a prior instead of retraining a task-specific generator.}
The role of the elite set is to define a filtered version of the current generated distribution. By repeatedly generating, screening, selecting, and fine-tuning, the model distribution is gradually shifted toward the region preferred by the downstream task. Since the optimization acts on a pretrained generator instead of starting from scratch, the model retains access to the broad topology knowledge learned offline. The method is better understood as steering a prior distribution than as directly fitting a new inverse model for each task.

\clearpage
\section{Details of Case Studies}
\label{sec:si_case_details}

The four case studies are designed to test GenTO under progressively different types of downstream design objectives while keeping the same overall offline--online framework. All cases start from a task-matched pretrained topology prior learned from the joint topology dataset described in Supplementary Section~\ref{sec:si_datasets}, and all cases use the same basic steering loop of sampling, feasibility screening, score evaluation, elite selection, and iterative fine-tuning. What changes from case to case are the physical objective, the feasibility constraints, the symmetry class, and the numerical evaluator used to assign design scores.

Together, the four cases cover four representative design regimes. Case 1 addresses single-objective extremization of a homogenized thermal property and includes both maximization and minimization tasks. Case 2 considers multi-objective morphology control, in which the optimization target is a Pareto trade-off between feature size and boundary complexity. Case 3 addresses property-targeted inverse design, where the goal is to match a prescribed elasticity tensor associated with auxetic effective behavior. Case 4 moves to function-targeted design in the frequency domain, where the target is a transmission spectrum with prescribed pass-band and stop-band regions. These four tasks span scalar optimization, vector-valued optimization, inverse constitutive design, and non-differentiable function-targeted optimization within one common framework.

The organization is important for the logic of the paper. The case studies are designed to demonstrate strong performance on four separate benchmarks and to show that the same learned topology-prior framework can be repeatedly repurposed as the task definition changes. They probe the scope of prior reusability across changing objectives, constraints, and physics solvers. The following subsections provide the detailed formulations, evaluators, and design rationales for each case.

\subsection{Thermal Conductivity Extremization}
\label{sec:si_case1}

\runinhead{Overview.}
Case 1 considers topology design for periodic heat conduction. The task is defined on a two-dimensional periodic unit cell, and each candidate topology is evaluated as a heterogeneous conducting medium with a solid phase and a void-like phase. This case is used as the first benchmark because it is physically intuitive, computationally efficient to evaluate, and already rich enough to expose two distinct optimization regimes: conductivity maximization and conductivity minimization. The same task-matched pretrained topology prior, steering loop, and geometric constraints can be reused for both tasks, making this case a clean first demonstration of the generality of GenTO.

\runinhead{Problem definition.}
Let $\mathbf{x}\in\{0,1\}^{256\times256}$ denote the binary topology of a periodic unit cell. The material volume fraction is fixed at
\begin{equation}
V_f(\mathbf{x}) = V_f^0 = 0.4.
\end{equation}
The design variable is the arrangement of the material phase, not the amount of material used. Two complementary design problems are considered:
\begin{equation}
\max_{\mathbf{x}} \; J_{\mathrm{HE}}(\mathbf{x}),
\qquad
\min_{\mathbf{x}} \; J_{\mathrm{HE}}(\mathbf{x}),
\end{equation}
\begin{equation}
\textrm{s.t.}\qquad \phi_{\mathrm{conn}}(\mathbf{x}) = 1,
\qquad
w_{\min}(\mathbf{x}) \geq w_0,
\qquad
V_f(\mathbf{x}) = V_f^0,
\end{equation}
where $\phi_{\mathrm{conn}}$ is the periodic connectivity indicator, $w_{\min}$ is the minimum feature size, and $w_0=5$ pixels is the prescribed lower bound.
The reported conductivity quantity is denoted by $J_{\mathrm{HE}}\equiv\kappa$, while the steering score follows the signed form used in Methods:
\begin{equation}
\begin{aligned}
J_{\mathrm{HE}}(\mathbf{x})
&\equiv
\kappa(\mathbf{x})
=
\bar{\kappa}_{\mathrm{eff}}(\mathbf{x}),\\
r_{\kappa}(\mathbf{x};\sigma_{\kappa})
&=
\sigma_{\kappa}\bar{\kappa}_{\mathrm{eff}}(\mathbf{x})
-
\lambda_{\kappa}\Delta\kappa_{\mathrm{eff}}(\mathbf{x}),
\end{aligned}
\end{equation}
where $\sigma_{\kappa}=1$ gives conductivity maximization and $\sigma_{\kappa}=-1$ gives conductivity minimization, $\bar{\kappa}_{\mathrm{eff}} = [\hat{q}_x(\bar{\boldsymbol{e}}^{x})+\hat{q}_y(\bar{\boldsymbol{e}}^{y})]/2$, $\Delta\kappa_{\mathrm{eff}} = \lvert \hat{q}_x(\bar{\boldsymbol{e}}^{x})-\hat{q}_y(\bar{\boldsymbol{e}}^{y})\rvert$, and $\lambda_{\kappa}=0.1$. Here, $\bar{\boldsymbol{e}}^{x}$ and $\bar{\boldsymbol{e}}^{y}$ are the two imposed macroscopic temperature-gradient directions. The superscripts $x$ and $y$ label loading directions rather than vector components, and $\hat{q}_x$ and $\hat{q}_y$ are scalar normalized averaged heat-flux responses. They are equivalent to the homogenized conductivity responses $H_{\kappa,11}$ and $H_{\kappa,22}$ along the two orthogonal loading directions. The dominant term in the score is the signed average effective conductivity, while the second term introduces a mild penalty on anisotropy.

The score is used for two reasons. First, averaging the two orthogonal responses gives a more balanced measure of thermal transport than using only one loading direction and provides a cleaner benchmark for periodic metamaterial design. Second, the anisotropy penalty suppresses highly directional solutions whose performance is dominated by one privileged transport path. As a result, the optimization is encouraged to search for topologies with strong overall thermal response instead of trivially directional layouts. In the maximization task, the signed score drives the distribution toward highly conductive topologies. In the minimization task, it drives the distribution toward low-conductivity topologies while retaining the same anisotropy penalty, which makes the search landscape much more irregular and provides a stronger test of the steering strategy.

\runinhead{FFT-based evaluation of the objective.}
The objective is evaluated by first-order periodic homogenization of steady-state heat conduction. The evaluation follows the approach described in Ref.~\cite{li2025enhanced}. The local problem is governed by
\begin{equation}
\nabla\cdot\boldsymbol{q}(\mathbf{r}) = 0,
\qquad
\boldsymbol{q}(\mathbf{r}) = k(\mathbf{r})\boldsymbol{e}(\mathbf{r}),
\qquad
\boldsymbol{e}(\mathbf{r}) = -\nabla \Theta(\mathbf{r}),
\end{equation}
where $\boldsymbol{q}$ is the local heat flux, $\boldsymbol{e}$ is the local temperature-gradient field, and $k(\mathbf{r})$ is the local scalar conductivity. In this case, both phases are isotropic. The material phase is assigned conductivity $k_{\mathrm{s}}=1.0$, whereas the void-like phase is assigned conductivity $k_{\mathrm{v}}=10^{-3}$.
The strong contrast between these two conductivities makes the task sensitive to topology while still avoiding the singularity of a perfectly insulating void phase.

To obtain the effective conductivity tensor, two independent macroscopic temperature-gradient loadings are imposed, namely,
\begin{equation}
\bar{\boldsymbol{e}}^{x} = [1,0]^{\mathsf{T}},
\qquad
\bar{\boldsymbol{e}}^{y} = [0,1]^{\mathsf{T}}.
\end{equation}
The superscripts $x$ and $y$ label the two imposed loading directions. In the equations below, $\bar{\boldsymbol{e}}^{\alpha}$ with $\alpha\in\{x,y\}$ denotes either prescribed macroscopic loading. For each loading, the local field is solved by an FFT-based Lippmann--Schwinger scheme following the same thermal homogenization framework used in Ref.~\cite{li2025enhanced}. A homogeneous reference medium with conductivity $k_0= (k_{\mathrm{s}}+k_{\mathrm{v}})/2$ is introduced first. The local heat flux can then be decomposed as
\begin{equation}
\boldsymbol{q}(\mathbf{r})
=
k_0\boldsymbol{e}(\mathbf{r})
\,+\,
\boldsymbol{\tau}_{\kappa}(\mathbf{r}),
\end{equation}
where $\boldsymbol{\tau}_{\kappa}$ is the polarization heat flux. Since the local conductivity is scalar in each phase, the polarization field is
\begin{equation}
\boldsymbol{\tau}_{\kappa}(\mathbf{r})
=
\left(k(\mathbf{r})-k_0\right)\boldsymbol{e}(\mathbf{r}).
\end{equation}
Applying the Fourier transform to the equilibrium equation and the constitutive decomposition gives, for every non-zero Fourier mode $\boldsymbol{\xi}\neq \mathbf{0}$,
\begin{equation}
\hat{\boldsymbol{e}}(\boldsymbol{\xi})
=
-\hat{\boldsymbol{\Gamma}}^{0}_{\kappa}(\boldsymbol{\xi})\hat{\boldsymbol{\tau}}_{\kappa}(\boldsymbol{\xi}),
\end{equation}
where $\hat{\boldsymbol{\Gamma}}^{0}_{\kappa}$ is the thermal Green operator of the reference medium. In real space, this is equivalent to the periodic Lippmann--Schwinger equation
\begin{equation}
\boldsymbol{e}(\mathbf{r})
\,+\,
\int_{\Omega}
\boldsymbol{\Gamma}^{0}_{\kappa}(\mathbf{r}-\mathbf{r}')
\cdot
\left[\left(k(\mathbf{r}')-k_0\right)\boldsymbol{e}(\mathbf{r}')\right]
\mathrm{d}\mathbf{r}'
-\bar{\boldsymbol{e}}^{\alpha}
=
\mathbf{0}.
\end{equation}
The zero Fourier mode is fixed by the prescribed macroscopic loading,
\begin{equation}
\hat{\boldsymbol{e}}(\mathbf{0})=\bar{\boldsymbol{e}}^{\alpha}.
\end{equation}

For an isotropic reference medium, the continuous Green operator is
\begin{equation}
\hat{\Gamma}^{0}_{ij}(\boldsymbol{\xi})
=
\frac{\xi_i\xi_j}{k_0(\xi_1^2+\xi_2^2)},
\qquad
\boldsymbol{\xi}\neq \mathbf{0}.
\end{equation}
In the discrete solver used here, a staggered-grid central-difference form is adopted, so the continuous wave vector is replaced by its discrete counterpart $\tilde{\boldsymbol{\xi}}=(\tilde{\xi}_1,\tilde{\xi}_2)$, which gives
\begin{equation}
\hat{\Gamma}^{0}_{ij}(\tilde{\boldsymbol{\xi}})
=
\frac{\tilde{\xi}_i\tilde{\xi}_j}{k_0(\tilde{\xi}_1^2+\tilde{\xi}_2^2)},
\qquad
\tilde{\boldsymbol{\xi}}\neq \mathbf{0},
\end{equation}
with
\begin{equation}
\hat{\Gamma}^{0}_{ij}(\mathbf{0})=0.
\end{equation}

The local field is then updated by fixed-point iteration. At iteration $m$,
\begin{equation}
\boldsymbol{\tau}_{\kappa}^{(m)}(\mathbf{r})
=
\left(k(\mathbf{r})-k_0\right)\boldsymbol{e}^{(m)}(\mathbf{r}),
\end{equation}
\begin{equation}
\hat{\boldsymbol{e}}^{(m+1)}(\boldsymbol{\xi})
=
-\hat{\boldsymbol{\Gamma}}^{0}_{\kappa}(\boldsymbol{\xi})\hat{\boldsymbol{\tau}}_{\kappa}^{(m)}(\boldsymbol{\xi}),
\qquad
\boldsymbol{\xi}\neq\mathbf{0},
\end{equation}
and the zero mode is reset to $\bar{\boldsymbol{e}}^{\alpha}$. The updated local heat flux is
\begin{equation}
\boldsymbol{q}^{(m+1)}(\mathbf{r})
=
k(\mathbf{r})\boldsymbol{e}^{(m+1)}(\mathbf{r}).
\end{equation}
In the present implementation, the unit cell is discretized on a $256\times256$ regular grid and the fixed-point iteration is performed for 500 steps for each macroscopic loading direction.

The FFT-based formulation is particularly suitable here because the design domain is periodic and voxel-based. Under these conditions, Fourier differentiation and convolution can be carried out efficiently on regular grids without mesh generation or repeated assembly of large sparse systems. This makes the cost of repeated physics evaluation much lower than in a comparable finite-element implementation, which is especially important in GenTO because the solver must be called many times during iterative sampling and steering. The conductivity benchmark is physically meaningful and computationally compatible with distribution-level optimization.

An additional practical advantage is that the FFT-based homogenization procedure can be implemented efficiently on GPUs and extended naturally to batch computation~\cite{li2025batch}. In this setting, multiple candidate unit cells are evaluated simultaneously, while the Fourier transforms, Green-operator applications, and pointwise constitutive updates are all carried out in parallel on the GPU. Since GenTO repeatedly evaluates many sampled topologies at every steering iteration, this batch-processing strategy provides a substantial speed advantage and makes FFT-based homogenization particularly suitable for distribution-level optimization.

After solving the two loading cases, the local heat flux and temperature-gradient fields are volume-averaged as $\hat{\boldsymbol{q}} = \lvert\Omega\rvert^{-1}\int_{\Omega}\boldsymbol{q}(\mathbf{r})\,\mathrm{d}\mathbf{r}$ and $\bar{\boldsymbol{e}} = \lvert\Omega\rvert^{-1}\int_{\Omega}\boldsymbol{e}(\mathbf{r})\,\mathrm{d}\mathbf{r}$. The homogenized conductivity tensor is then defined by
\begin{equation}
\hat{\boldsymbol{q}}
=
\mathbf{H}_{\kappa}\bar{\boldsymbol{e}}.
\end{equation}
With the two orthogonal macroscopic loadings, the tensor is assembled as
\begin{equation}
\mathbf{H}_{\kappa}
=
\begin{bmatrix}
H_{\kappa,11} & H_{\kappa,12}\\
H_{\kappa,21} & H_{\kappa,22}
\end{bmatrix}.
\end{equation}
The reported quantity $J_{\mathrm{HE}}$ and the steering score $r_{\kappa}$ are then evaluated from $H_{\kappa,11}$ and $H_{\kappa,22}$ through the definitions above. In the optimization loop, both the maximization and minimization tasks rank candidates by descending $r_{\kappa}$, with the task direction set by $\sigma_{\kappa}$.

\runinhead{Evaluation of the constraints.}
Two geometric constraints are imposed before a candidate is accepted for physics-based scoring. The first is periodic connectivity. This constraint is introduced because the benchmark is intended to compare connected architected materials, not trivial disconnected layouts. Without such a restriction, the minimization task could collapse toward fragmented patterns that suppress transport simply by destroying the material network, which would make the design problem less informative. The connectivity requirement therefore keeps the search within the class of periodic topologies that retain a meaningful transport skeleton.

\runinhead{Periodic connectivity.}
Let $\mathbf{x}=\{x_{ij}\}_{i,j=1}^{N}$ denote the binary unit cell, where $x_{ij}=1$ and $x_{ij}=0$ correspond to the material and void phases, respectively. The material phase is
\begin{equation}
\Omega_{\mathrm{m}}=\{(i,j):x_{ij}=1\}.
\end{equation}
The connectivity test follows the periodic boundary-stitching logic used in the optimization code. It labels material components in the unit cell, merges components that touch opposite periodic boundaries at matching positions, and then checks whether all material components participate in horizontal or vertical periodic paths, with at least one component spanning both directions.

The connected components of $\Omega_{\mathrm{m}}$ are identified on the image grid using the four-neighbor relation
\begin{equation}
(i,j)\sim_4(k,\ell)
\quad \Longleftrightarrow \quad
\lvert i-k\rvert+\lvert j-\ell\rvert=1.
\end{equation}
Let these preliminary material components be denoted by
\begin{equation}
\mathcal{C}_{\mathrm{m}}=\{C_1,C_2,\dots,C_{N_c}\}.
\end{equation}
Since the unit cell is periodic, two material components that touch opposite boundaries at the same periodic position should be treated as a single periodic component. This is enforced by introducing the boundary-matching relation
$C_a \equiv_p C_b$
whenever at least one of the following conditions is satisfied:
\begin{equation}
\exists\, i\in\{1,\dots,N\}
\quad s.t. \quad
(i,1)\in C_a,\ (i,N)\in C_b,
\end{equation}
or
\begin{equation}
\exists\, j\in\{1,\dots,N\}
\quad s.t. \quad
(1,j)\in C_a,\ (N,j)\in C_b.
\end{equation}
The transitive closure of this relation defines the periodic component classes. In practice, this amounts to repeatedly merging component labels connected through opposite boundaries until no further merge is needed. Each preliminary component $C_a$ is then mapped to its periodic root class $\Pi(C_a)$.

Four boundary-supported root sets are then defined:
\begin{equation}
\mathcal{R}_{L}=\{\Pi(C_a): C_a \cap \partial\Omega_{L}\neq\varnothing\},
\qquad
\mathcal{R}_{R}=\{\Pi(C_a): C_a \cap \partial\Omega_{R}\neq\varnothing\},
\end{equation}
\begin{equation}
\mathcal{R}_{T}=\{\Pi(C_a): C_a \cap \partial\Omega_{T}\neq\varnothing\},
\qquad
\mathcal{R}_{B}=\{\Pi(C_a): C_a \cap \partial\Omega_{B}\neq\varnothing\},
\end{equation}
where $\partial\Omega_{L}$, $\partial\Omega_{R}$, $\partial\Omega_{T}$, and $\partial\Omega_{B}$ denote the left, right, top, and bottom boundaries of the unit cell. From these sets, the horizontally spanning, vertically spanning, and fully spanning root classes are written as
\begin{equation}
\mathcal{R}_{H}=\mathcal{R}_{L}\cap\mathcal{R}_{R},
\qquad
\mathcal{R}_{V}=\mathcal{R}_{T}\cap\mathcal{R}_{B},
\qquad
\mathcal{R}_{HV}=\mathcal{R}_{H}\cap\mathcal{R}_{V}.
\end{equation}
In addition, the set of all periodic material root classes is
\begin{equation}
\mathcal{R}_{\mathrm{all}}=\{\Pi(C_a): C_a\in\mathcal{C}_{\mathrm{m}}\}.
\end{equation}
The periodic-connectivity test can be written as the following algorithm:

\textit{Step 1.} Label all material pixels using the four-neighbor relation and obtain $\mathcal{C}_{\mathrm{m}}$.

\textit{Step 2.} For every row $i$, if $x_{i1}=x_{iN}=1$, merge the labels attached to $(i,1)$ and $(i,N)$. For every column $j$, if $x_{1j}=x_{Nj}=1$, merge the labels attached to $(1,j)$ and $(N,j)$.

\textit{Step 3.} Build the root sets $\mathcal{R}_{L}$, $\mathcal{R}_{R}$, $\mathcal{R}_{T}$, and $\mathcal{R}_{B}$ from the merged labels.

\textit{Step 4.} Compute $\mathcal{R}_{H}$, $\mathcal{R}_{V}$, and $\mathcal{R}_{HV}$, and define the periodic-connectivity indicator
\begin{equation}
\phi_{\mathrm{conn}}(\mathbf{x})=
\begin{cases}
1, & \mathcal{R}_{HV}\neq\varnothing \text{ and } \mathcal{R}_{\mathrm{all}}\subseteq (\mathcal{R}_{H}\cup\mathcal{R}_{V}),\\
0, & \text{otherwise.}
\end{cases}
\end{equation}

The indicator removes isolated islands and boundary-supported fragments that do not participate in a periodic transport path. The first condition requires at least one periodic component to span both horizontal and vertical directions. The second condition requires every material component to belong to either a horizontally or vertically supported periodic path.

The second constraint is the minimum feature size. This constraint is introduced to suppress arbitrarily thin bridges, reduce grid-scale artifacts, and maintain a more realistic level of manufacturability. It also stabilizes the optimization by preventing the solver from repeatedly selecting structures that achieve extreme conductivity values through vanishingly small local features.

\runinhead{Minimum feature size.}
The minimum feature size is estimated from the narrowest material bridge separating neighboring void regions. To remove boundary bias, the binary unit cell is first periodically extended to a $3\times3$ tiled image $\mathbf{x}^{\mathrm{tile}}\in\{0,1\}^{3N\times 3N}$ obtained by repeating $\mathbf{x}$ in both coordinate directions. All subsequent computations are carried out on this periodic extension. Let $\Omega_{\mathrm{v}}^{\mathrm{tile}}$ denote the void phase on the tiled image,
\begin{equation}
\Omega_{\mathrm{v}}^{\mathrm{tile}}=\{(i,j):x^{\mathrm{tile}}_{ij}=0\},
\end{equation}
and let the connected void components be
\begin{equation}
\mathcal{C}_{\mathrm{v}}=\{V_1,V_2,\dots,V_{N_v}\}.
\end{equation}
If $N_v<2$, the minimum feature size is regarded as undefined and the sample is rejected.

Next, for every material pixel $\mathbf{r}\in\Omega_{\mathrm{m}}^{\mathrm{tile}}$, the Euclidean distance to the nearest void pixel is evaluated:
\begin{equation}
\Omega_{\mathrm{m}}^{\mathrm{tile}}=\{(i,j):x^{\mathrm{tile}}_{ij}=1\},
\end{equation}
\begin{equation}
\delta_{\mathrm{v}}(\mathbf{r})=\min_{\mathbf{s}\in\Omega_{\mathrm{v}}^{\mathrm{tile}}}\|\mathbf{r}-\mathbf{s}\|_2,
\qquad \mathbf{r}\in\Omega_{\mathrm{m}}^{\mathrm{tile}}.
\end{equation}
At the same time, each material pixel is assigned the label of its nearest void component. Let this nearest-void map be denoted by
\begin{equation}
\eta_{\mathrm{v}}(\mathbf{r})\in\{1,\dots,N_v\},
\qquad \mathbf{r}\in\Omega_{\mathrm{m}}^{\mathrm{tile}}.
\end{equation}
More precisely,
\begin{equation}
\eta_{\mathrm{v}}(\mathbf{r})=\alpha
\quad \Longleftrightarrow \quad
\exists\,\mathbf{s}^{\ast}\in V_{\alpha}
\quad \text{such that} \quad
\|\mathbf{r}-\mathbf{s}^{\ast}\|_2=\delta_{\mathrm{v}}(\mathbf{r}).
\end{equation}
The idea is that if two neighboring material pixels are closest to different void components, then they lie across a material bridge separating those voids. For every horizontally or vertically adjacent material-pixel pair $(\mathbf{r}_p,\mathbf{r}_n)$ satisfying $\eta_{\mathrm{v}}(\mathbf{r}_p)\neq \eta_{\mathrm{v}}(\mathbf{r}_n)$, a local bridge width is estimated by
\begin{equation}
w(\mathbf{r}_p,\mathbf{r}_n)=\delta_{\mathrm{v}}(\mathbf{r}_p)+\delta_{\mathrm{v}}(\mathbf{r}_n).
\end{equation}
The minimum feature size is then defined as
\begin{equation}
w_{\min}(\mathbf{x})
=
\min_{(\mathbf{r}_p,\mathbf{r}_n)\in\mathcal{N}_{\mathrm{diff}}}
w(\mathbf{r}_p,\mathbf{r}_n),
\end{equation}
where $\mathcal{N}_{\mathrm{diff}}$ is the set of all horizontal and vertical neighboring material-pixel pairs whose nearest void labels are different. Intuitively, this quantity measures the narrowest material neck between adjacent void domains under periodic continuation. A candidate is retained only if $w_{\min}(\mathbf{x}) > 5$.
The feature-size test can be summarized as follows:

\textit{Step 1.} Periodically tile the unit cell to form $\mathbf{x}^{\mathrm{tile}}$ and label all connected void components $\mathcal{C}_{\mathrm{v}}$.

\textit{Step 2.} For each material pixel in the tiled image, compute its distance $\delta_{\mathrm{v}}(\mathbf{r})$ to the nearest void and record the corresponding nearest-void label $\eta_{\mathrm{v}}(\mathbf{r})$.

\textit{Step 3.} Visit every horizontal and vertical neighboring material-pixel pair. If the two pixels have different nearest-void labels, compute the local bridge width $w(\mathbf{r}_p,\mathbf{r}_n)=\delta_{\mathrm{v}}(\mathbf{r}_p)+\delta_{\mathrm{v}}(\mathbf{r}_n)$.

\textit{Step 4.} Take the smallest such value over all eligible neighboring pairs and define it as $w_{\min}(\mathbf{x})$.

\textit{Step 5.} Accept the sample only if $w_{\min}(\mathbf{x})>5$.

The construction is simple, but it has a clear geometric meaning. The distance field measures how far the material lies from the surrounding void phase, while the nearest-void labels distinguish which two void domains are being separated by a given ligament. Their combination detects the narrowest material neck in a way that is naturally compatible with periodic continuation. The resulting constraint suppresses vanishingly thin ligaments, removes many grid-scale artifacts, and gives a simple but effective proxy for manufacturability and numerical robustness.

\runinhead{Role of this case study.}
Case~1 is a useful benchmark for several reasons. From the physical side, thermal conductivity is one of the most classical homogenized properties in topology optimization, so the task is easy to interpret and compare across methods. From the numerical side, FFT-based homogenization makes repeated evaluation sufficiently fast for iterative distribution steering. From the optimization side, the pair of maximization and minimization tasks is especially informative: conductivity maximization is comparatively well behaved, whereas conductivity minimization under connectivity and feature-size constraints is much more non-convex and admits many competing topological strategies. This asymmetry makes the case a strong first test of GenTO.

It is also a useful benchmark at the methodological level. With the same task-matched pretrained prior, volume-fraction control, and steering framework used for both tasks, the only substantive difference lies in the objective direction. This makes it possible to isolate the effect of the downstream physics objective while keeping the rest of the pipeline unchanged. As discussed in the main text, this case already reveals one of the main strengths of GenTO: distribution steering remains effective in the relatively smooth maximization landscape and in the more irregular minimization landscape where single-trajectory optimization methods are more prone to poor local solutions.

\subsection{Multi-Objective Morphology Control}
\label{sec:si_case2}

\runinhead{Overview.}
Case 2 considers a morphology-level multi-objective design problem. Instead of targeting a homogenized physical property, this case directly controls two geometric descriptors of the topology itself, namely the minimum feature size and the boundary fractal dimension. The candidate design is again a two-dimensional periodic unit cell, but the optimization target is no longer a single scalar score. Rather, the aim is to improve the Pareto region formed by these two competing objectives. The case examines whether GenTO can steer a reusable topology prior toward a broader and better-performing trade-off set in morphology space while preserving a diverse population of candidate structures.

In this case, the periodic pretrained prior and the steering process follow the same general framework introduced in Supplementary Sections~\ref{sec:si_method_details} and \ref{sec:si_learning_steering}. The main difference is that the objective is vector-valued, and the retained elite set is selected by Pareto dominance instead of scalar ranking. The optimization is performed in the periodic setting, so that the resulting Pareto front reflects morphology evolution under the same periodic-connectivity condition used for evaluation.

\runinhead{Problem definition.}
Let $\mathbf{x}\in\{0,1\}^{256\times256}$ denote the binary topology of a periodic unit cell. The material volume fraction is fixed at
\begin{equation}
V_f(\mathbf{x})=V_f^0=0.5.
\end{equation}
The design problem is written as
\begin{equation}
\max_{\mathbf{x}} \;
\mathbf{f}_{\mathrm{MO}}(\mathbf{x})
=
\Bigl(
f_{\mathrm{fractal}}(\mathbf{x}),
f_{\mathrm{size}}(\mathbf{x})
\Bigr)^{\mathsf T},
\end{equation}
\begin{equation}
\textrm{s.t.}
\qquad
\phi_{\mathrm{conn}}(\mathbf{x})=1,
\qquad
V_f(\mathbf{x})=V_f^0,
\end{equation}
where $f_{\mathrm{fractal}}$ is the boundary-fractal-dimension objective, $f_{\mathrm{size}}$ is the minimum feature-size objective, and $\phi_{\mathrm{conn}}$ is the periodic-connectivity indicator.

These two objectives are chosen because they capture two practically important but conflicting aspects of topology morphology. Larger feature sizes generally improve geometric robustness and manufacturability, whereas larger boundary fractal dimensions indicate richer geometric complexity and more intricate interfacial patterns. In architected materials, these two tendencies are often in direct competition: coarse and regular morphologies tend to have smoother boundaries, while highly complex boundaries are usually associated with finer structural details. The resulting trade-off makes this case a natural benchmark for multi-objective design.

It is also important that the minimum feature size is treated here as an optimization objective, not as a feasibility constraint. In Case 1, a lower bound on feature size was introduced to suppress vanishingly thin ligaments in a thermal-transport setting. Here, however, feature size itself is one of the two target descriptors to be improved. Imposing an additional hard lower bound would artificially truncate the morphology trade-off and remove a meaningful part of the Pareto landscape. For this reason, the only explicit geometric feasibility requirement retained in this case is periodic connectivity.

\runinhead{Evaluation of the objectives.}
The feature-size objective, $f_{\mathrm{size}}(\mathbf{x})$, is defined using the same minimum-feature measure introduced in Case 1. The unit cell is periodically tiled to form a $3\times3$ extended image, the void domains are labeled, and for each material pixel the distance to the nearest void and the identity of the nearest void component are evaluated. The minimum feature size is then written as
\begin{equation}
f_{\mathrm{size}}(\mathbf{x})
=
w_{\min}(\mathbf{x})
=
\min_{(\mathbf{r}_p,\mathbf{r}_n)\in\mathcal{N}_{\mathrm{diff}}}
\left[\delta_{\mathrm{v}}(\mathbf{r}_p)+\delta_{\mathrm{v}}(\mathbf{r}_n)\right],
\end{equation}
where $\mathcal{N}_{\mathrm{diff}}$ is the set of neighboring material-pixel pairs associated with different nearest void components. As explained in Case 1, this quantity measures the narrowest material bridge separating adjacent void domains under periodic continuation. In the present case it is not used to reject samples, but directly serves as one objective to be maximized.

The fractal objective, $f_{\mathrm{fractal}}(\mathbf{x})$, measures the geometric complexity of the material boundary through a box-counting estimate of fractal dimension~\cite{liebovitch1989fast}. Let $\Gamma(\mathbf{x}) \subset \Omega$ denote the inner boundary of the material phase extracted from the binary topology. For a box size $s$, the unit cell is partitioned into square boxes of side length $s$, and the number of boxes intersecting the boundary is counted:
\begin{equation}
N_{\Gamma}(s)
=
\#\Bigl\{
Q_m(s): Q_m(s)\cap\Gamma(\mathbf{x})\neq\varnothing
\Bigr\},
\end{equation}
where $\{Q_m(s)\}$ denotes the full set of grid-aligned boxes at scale $s$.

If the boundary exhibits scale-dependent complexity, the box count approximately follows the relation
\begin{equation}
N_{\Gamma}(s)\propto s^{-D_{\Gamma}},
\end{equation}
which leads to the linear form
\begin{equation}
\log N_{\Gamma}(s)
=
-D_{\Gamma}\log s + C.
\end{equation}
The boundary fractal dimension is estimated from the negative slope of the log--log regression:
\begin{equation}
f_{\mathrm{fractal}}(\mathbf{x}) = D_{\Gamma}.
\end{equation}

In the present implementation, the box sizes are sampled progressively from small to moderate scales, starting from $s=2$ pixels and increasing approximately geometrically until the box size reaches one quarter of the unit-cell width. For each scale, only boxes containing at least one boundary pixel contribute to $N_{\Gamma}(s)$. A linear least-squares fit is then performed in the $(\log s,\log N_{\Gamma}(s))$ plane. In addition to the slope, the coefficient of determination $R^2$ is also computed as a diagnostic of the scaling quality, although only the fitted fractal dimension is used in the optimization.

The descriptor is chosen because it captures a type of geometric richness that is not reflected by volume fraction or by a single characteristic length scale. Two topologies may have the same material fraction and similar minimum feature sizes, yet differ greatly in interfacial complexity. The boundary fractal dimension provides a compact quantitative measure of this complexity and complements the feature-size objective in a meaningful way.

\runinhead{Evaluation of the constraint.}
The feasibility constraint in this case is the same periodic boundary-stitching criterion introduced in Case 1. In compact form, only samples with $\phi_{\mathrm{conn}}(\mathbf{x})=1$ are retained for objective evaluation and Pareto selection. The detailed construction of this indicator is exactly the same as that described in Case 1 and is not repeated here.

This choice of constraint is deliberate. Since the purpose of this case is to study morphology trade-offs within the class of meaningful periodic material networks, disconnected samples are excluded. At the same time, no additional minimum-feature constraint is imposed, because doing so would interfere directly with one of the two optimization objectives. The resulting formulation keeps the feasible set broad enough to expose a non-trivial Pareto structure while still preventing degenerate disconnected layouts.

\runinhead{Role of this case study.}
Case~2 plays a distinct role in the validation of GenTO. Unlike Case 1, where the objective is scalar and the optimization seeks one dominant performance direction, the present problem requires the method to improve a whole set of mutually competing designs. The quality of optimization is reflected not by a single best topology, but by the outward movement, coverage, and continuity of the Pareto front in morphology space.

The case is especially suitable for testing the core idea of distribution steering. Since GenTO operates on a sample population instead of on a single optimization trajectory, it can naturally retain and refine multiple trade-off modes in parallel. From this perspective, the method is well aligned with the structure of multi-objective design: the target is an improved distribution of non-dominated topologies, not a single optimum.

This case is also a useful benchmark because the two objectives are easy to interpret visually. As the Pareto region expands, the corresponding topology population evolves continuously from coarse, large-feature designs to highly intricate, boundary-rich designs. This provides a direct way to assess both optimization quality and whether the learned prior remains capable of supporting diverse morphology evolution under steering. For these reasons, this case serves as a representative benchmark for multi-objective topology generation in GenTO.

\subsection{Property-Targeted Auxetic Design}
\label{sec:si_case3}

\runinhead{Overview.}
Case 3 considers a property-targeted inverse-design problem for auxetic metamaterials. Unlike Cases 1 and 2, where the objective is defined either by a homogenized scalar property or by morphology descriptors, the present case aims to match a prescribed effective elasticity tensor as closely as possible. The target is given directly in property space, and the optimization seeks topologies whose homogenized elastic response approaches this target under periodic boundary conditions.

This case is particularly important because it tests whether the pretrained topology prior can be steered from extreme values of a scalar metric to a narrowly specified region in a high-dimensional property space. Three tasks are considered, corresponding to periodic, four-fold, and eight-fold symmetry groups. All three targets are chosen to represent auxetic effective behavior, and together they form a sequence of increasing structural restriction under progressively stronger symmetry conditions.

\runinhead{Problem definition.}
Let $\mathbf{x}\in\{0,1\}^{256\times256}$ denote the binary topology of a periodic unit cell. The material volume fraction is fixed at
\begin{equation}
V_f(\mathbf{x})=V_f^0=0.5.
\end{equation}
For each symmetry class $\chi\in\{\mathrm{P},4,8\}$, the design problem is written as
\begin{equation}
\min_{\mathbf{x}} \;
\varepsilon_{\mathrm{ES}}^{(\chi)}(\mathbf{x}),
\end{equation}
\begin{equation}
\textrm{s.t.}
\qquad
\phi_{\mathrm{conn}}(\mathbf{x})=1,
\qquad
w_{\mathrm{skel}}(\mathbf{x})\geq w_0,
\qquad
V_f(\mathbf{x})=V_f^0,
\end{equation}
where $\phi_{\mathrm{conn}}$ is the periodic-connectivity indicator, $w_{\mathrm{skel}}$ is a skeleton-based minimum feature size, and $w_0=5$ pixels.

The objective is the mean absolute mismatch between the homogenized elasticity tensor of the candidate topology and the prescribed target tensor:
\begin{equation}
\varepsilon_{\mathrm{ES}}^{(\chi)}(\mathbf{x})
=
\frac{1}{9}
\sum_{i=1}^{3}\sum_{j=1}^{3}
\lvert
H_{\mathrm{ES},ij}(\mathbf{x})-H_{\mathrm{ES,target},ij}^{(\chi)}
\rvert.
\end{equation}
In the actual steering loop, this mismatch is converted into a maximization score through
\begin{equation}
r_{\mathrm{ES}}^{(\chi)}(\mathbf{x})=-\varepsilon_{\mathrm{ES}}^{(\chi)}(\mathbf{x}),
\end{equation}
so that samples with smaller tensor error receive larger scores.

The target properties are first specified as engineering constants. For the periodic target, $E_1=0.4$, $E_2=0.2$, $\nu_{12}=-0.8$, $G_{12}=0.06$, $\eta_1=0.2$, and $\eta_2=0.04$. For the four-fold target, $E_1=0.8$, $E_2=0.2$, $\nu_{12}=-0.7$, and $G_{12}=0.08$. For the eight-fold target, $E_1=E_2=0.7$, $\nu_{12}=\nu_{21}=-0.36$, and $G_{12}=0.25$. These engineering constants are converted to stiffness matrices through the corresponding plane-stress compliance matrix, with $\mathbf{S}^{(\chi)}=(\mathbf{H}_{\mathrm{ES,target}}^{(\chi)})^{-1}$. The resulting target stiffness matrices used in the optimization are
\begin{equation}
\mathbf{H}_{\mathrm{ES,target}}^{(\mathrm{P})}
=
\begin{bmatrix}
0.5919 & -0.2362 & -0.0149\\
-0.2362 & 0.2943 & 0.0036\\
-0.0149 & 0.0036 & 0.0604
\end{bmatrix},
\end{equation}
\begin{equation}
\mathbf{H}_{\mathrm{ES,target}}^{(4)}
=
\begin{bmatrix}
0.9117 & -0.1595 & 0\\
-0.1595 & 0.2279 & 0\\
0 & 0 & 0.0800
\end{bmatrix},
\end{equation}
\begin{equation}
\mathbf{H}_{\mathrm{ES,target}}^{(8)}
=
\begin{bmatrix}
0.8042 & -0.2895 & 0\\
-0.2895 & 0.8042 & 0\\
0 & 0 & 0.2500
\end{bmatrix}.
\end{equation}
These targets all correspond to auxetic effective behavior while spanning different symmetry classes. The case therefore does not merely ask whether GenTO can optimize one prescribed response, but whether the same distribution-steering framework can repeatedly adapt to different inverse targets under different structural priors.

\runinhead{Evaluation of the objective.}
The effective elasticity tensor is evaluated by first-order periodic homogenization for linear elasticity. Let $\boldsymbol{\sigma}(\mathbf{r})$ and $\boldsymbol{\varepsilon}(\mathbf{r})$ denote the local stress and strain fields. The local equilibrium and constitutive equations are
\begin{equation}
\nabla\cdot \boldsymbol{\sigma}(\mathbf{r})=\mathbf{0},
\qquad
\boldsymbol{\sigma}(\mathbf{r})
=
\mathbb{C}(\mathbf{r}):\boldsymbol{\varepsilon}(\mathbf{r}),
\end{equation}
with
\begin{equation}
\boldsymbol{\varepsilon}(\mathbf{r})
=
\overline{\boldsymbol{\varepsilon}}
\,+\,
\widetilde{\boldsymbol{\varepsilon}}(\mathbf{r}),
\end{equation}
where $\overline{\boldsymbol{\varepsilon}}$ is the prescribed macroscopic strain and $\widetilde{\boldsymbol{\varepsilon}}$ is the periodic fluctuation field. In the present case, the solid phase is treated as an isotropic elastic material with bulk modulus $K_{\mathrm{bulk}}=8.33$ and shear modulus $\mu_{\mathrm{s}}=3.86$, whereas the void-like phase is assigned vanishing stiffness. The local constitutive tensor is
\begin{equation}
\mathbb{C}(\mathbf{r})
=
K_{\mathrm{bulk}}(\mathbf{r})\,\mathbf{I}\otimes\mathbf{I}
\, + \,
2\mu_{\mathrm{s}}(\mathbf{r})
\left(
\mathbb{I}^{s}-\frac{1}{3}\mathbf{I}\otimes\mathbf{I}
\right),
\end{equation}
where $\mathbf{I}$ is the second-order identity tensor and $\mathbb{I}^{s}$ is the symmetric fourth-order identity.

The same FFT-based homogenization philosophy used in Case 1 is adopted here, but now for linear elasticity instead of scalar heat conduction~\cite{moulinec1998numerical}. More specifically, the elasticity solver follows a Galerkin-FFT formulation in which the periodic strain correction is represented on the regular grid, projected through a Fourier-space Green operator, and solved with a MINRES-style iterative solver~\cite{zeman2017finite}. The unit cell is subjected to three independent macroscopic strain loadings corresponding to axial loading in the first direction, axial loading in the second direction, and in-plane shear. Denoting these three loadings by $\overline{\boldsymbol{\varepsilon}}^{(1)}$, $\overline{\boldsymbol{\varepsilon}}^{(2)}$, and $\overline{\boldsymbol{\varepsilon}}^{(3)}$, the associated local stress fields are $\boldsymbol{\sigma}^{(1)}$, $\boldsymbol{\sigma}^{(2)}$, and $\boldsymbol{\sigma}^{(3)}$.

The spectral equilibrium problem is built from the Fourier-space Green operator for linear elasticity~\cite{zeman2017finite},
\begin{equation}
\widehat{\Gamma}^{0}_{ijkl}(\boldsymbol{\xi})
\propto
-\frac{\xi_i\xi_j\xi_k\xi_l}{(\boldsymbol{\xi}\cdot\boldsymbol{\xi})^2}
\,+\,
\frac{
\delta_{jk}\xi_i\xi_l
+\delta_{jl}\xi_i\xi_k
+\delta_{ik}\xi_j\xi_l
+\delta_{il}\xi_j\xi_k
}{
2(\boldsymbol{\xi}\cdot\boldsymbol{\xi})
},
\qquad
\boldsymbol{\xi}\neq \mathbf{0},
\end{equation}
which maps the stress field to a compatible strain correction under periodic boundary conditions. Starting from the prescribed macroscopic strain, the local field is updated by the MINRES-style iteration until the strain correction becomes sufficiently small. This Galerkin-FFT formulation is used here for the same reason as in Case 1: on a periodic voxel grid, it is substantially more efficient than repeated finite-element homogenization and is better suited to the repeated many-sample evaluations required by distribution steering.

After solving the three loading cases, the homogenized elasticity tensor is obtained by averaging the local stresses over the unit cell. In two-dimensional Voigt notation,
\begin{equation}
\overline{\boldsymbol{\sigma}}
=
\mathbf{H}_{\mathrm{ES}}\,
\overline{\boldsymbol{\varepsilon}},
\qquad
\mathbf{H}_{\mathrm{ES}}\in\mathbb{R}^{3\times3}.
\end{equation}
The three averaged stress responses are assembled into
\begin{equation}
\mathbf{H}_{\mathrm{ES}}(\mathbf{x})
=
\begin{bmatrix}
H_{\mathrm{ES},11} & H_{\mathrm{ES},12} & H_{\mathrm{ES},13}\\
H_{\mathrm{ES},21} & H_{\mathrm{ES},22} & H_{\mathrm{ES},23}\\
H_{\mathrm{ES},31} & H_{\mathrm{ES},32} & H_{\mathrm{ES},33}
\end{bmatrix},
\end{equation}
and the tensor-matching error $\varepsilon_{\mathrm{ES}}^{(\chi)}(\mathbf{x})$ is then computed from the definition above. Every candidate topology is thereby assigned a score directly in property space instead of through an indirect scalar descriptor.

The objective expresses the inverse-design task in the most direct form. Instead of optimizing one scalar elastic modulus and hoping that the remaining components evolve favorably, the full target tensor is prescribed explicitly. The tensor-level target is especially important for auxetic design, where the desired behavior depends on the coupled elastic response, not on one isolated entry of the constitutive matrix.

\runinhead{Evaluation of the constraints.}
The first feasibility requirement is periodic connectivity, which is defined exactly as in Cases 1 and 2 through the combined criterion
\begin{equation}
\phi_{\mathrm{conn}}(\mathbf{x})=
\phi_{\mathrm{stitch}}(\mathbf{x})\,\phi_{\mathrm{tile}}(\mathbf{x}).
\end{equation}
Only samples that satisfy this periodic-network condition are retained. The reason is the same as before: disconnected fragments do not define meaningful load-bearing metamaterials and would make the property-matching problem physically uninformative.

The second feasibility requirement is a minimum feature size, but in this case the thickness is evaluated differently from Cases 1 and 2. Rather than measuring the narrowest bridge between neighboring void regions, the present case estimates the local material thickness from the skeleton of the structure. This choice is more suitable for auxetic topologies, which often contain slender re-entrant ligaments and hinged load paths whose mechanical integrity depends directly on the thickness of the solid branches.

To remove boundary bias, the topology is first periodically tiled to form a $3\times3$ extended image $\mathbf{x}^{\mathrm{tile}}$. A mild morphological cleaning is then applied to suppress pixel-scale artifacts before thickness evaluation. Let $\Omega_{\mathrm{m}}^{\mathrm{tile}}$ denote the material phase on this tiled image. The Euclidean distance field inside the material is
\begin{equation}
\delta_{\mathrm{m}}(\mathbf{r})
=
\operatorname{dist}\!\left(
\mathbf{r},
\partial \Omega_{\mathrm{m}}^{\mathrm{tile}}
\right),
\qquad
\mathbf{r}\in\Omega_{\mathrm{m}}^{\mathrm{tile}}.
\end{equation}
The material skeleton, denoted by $\mathcal{S}_{\mathrm{m}}$, is then extracted from the tiled image. Restricting attention to the central unit cell, each skeleton point $\mathbf{r}\in\mathcal{S}_{\mathrm{m}}$ carries a local radius equal to $\delta_{\mathrm{m}}(\mathbf{r})$. The corresponding local thickness is therefore
\begin{equation}
t_{\mathrm{skel}}(\mathbf{r})=2\,\delta_{\mathrm{m}}(\mathbf{r}),
\qquad
\mathbf{r}\in\mathcal{S}_{\mathrm{m}}.
\end{equation}
The skeleton-based minimum feature size is defined as
\begin{equation}
w_{\mathrm{skel}}(\mathbf{x})
=
\min_{\mathbf{r}\in\mathcal{S}_{\mathrm{m}}^{\mathrm{center}}}
t_{\mathrm{skel}}(\mathbf{r}),
\end{equation}
where $\mathcal{S}_{\mathrm{m}}^{\mathrm{center}}$ denotes the skeleton points lying in the central copy of the tiled image. The candidate is accepted only if
$w_{\mathrm{skel}}(\mathbf{x})>5$.

The skeleton-based constraint has a clear mechanical meaning. Along the medial axis of the material network, the distance field records the radius of the largest inscribed circle supported by the surrounding boundary. Doubling this radius gives an effective local branch thickness. Taking the minimum over the skeleton detects the thinnest load-bearing ligament in the structure. The constraint is a simple and effective way to exclude mechanically fragile or pixel-scale auxetic patterns before homogenization.

\runinhead{Role of this case study.}
Case~3 plays a different role from the previous two benchmarks. Cases 1 and 2 mainly examine whether the topology prior can be steered toward extreme scalar objectives or expanded Pareto regions. By contrast, the present case asks whether GenTO can reach a prescribed target in property space. The challenge includes inverse design under a highly nonlinear and many-to-one mapping between topology and effective response, beyond optimization alone.

The case is especially demanding. Many different topologies can lead to broadly similar elastic trends, but matching a full target tensor requires coordinated control of multiple constitutive components at once. The three symmetry-specific targets further increase the difficulty by forcing the framework to adapt under different structural restrictions. In this sense, the case provides a direct test of whether a reusable prior can be steered beyond its initial support toward narrowly specified, task-dependent property regions.

Finally, this case is a particularly informative benchmark because the selected targets correspond to auxetic behavior, a canonical but non-trivial class of architected response. Auxetic metamaterials are both mechanically distinctive and practically important, and they are widely used as representative inverse-design targets in the literature. Demonstrating accurate tensor-level matching in this setting provides strong evidence that GenTO can support property-targeted metamaterial design beyond objective extremization.

\subsection{Frequency-Domain Vibration Transmission Design}
\label{sec:si_case4}

\runinhead{Overview.}
Case 4 considers function-targeted topology design in the frequency domain. Instead of targeting a homogenized effective property, the goal is to generate periodic structures whose vibration-transmission response matches a prescribed pass-band and stop-band pattern over a finite frequency interval. The task is qualitatively different from the previous cases: the design target is now a frequency-dependent response curve, not a static scalar or tensor quantity.

Two target functions are considered. The first is a single-pass-band design, in which the structure is required to transmit vibration efficiently in the middle part of the frequency range while suppressing transmission at lower and higher frequencies. The second is a dual-pass-band design, in which efficient transmission is required in two separated frequency windows with a stop band in between. In both settings, the same 4-fold pretrained prior and the same iterative steering framework are used, but the candidate structures are ranked by how well their transmission spectra satisfy the prescribed band requirements.

\runinhead{Problem definition.}
Let $\mathbf{x}\in\{0,1\}^{256\times256}$ denote the binary topology of the unit cell. The material volume fraction is fixed at
\begin{equation}
V_f(\mathbf{x})=V_f^0=0.5.
\end{equation}
The optimization problem is written as
\begin{equation}
\max_{\mathbf{x}}\;
s_{\mathrm{FQ}}(\mathbf{x})
\qquad
\textrm{s.t.}
\qquad
\phi_{\mathrm{v}}(\mathbf{x})=1,
\qquad
V_f(\mathbf{x})=V_f^0,
\end{equation}
where $s_{\mathrm{FQ}}$ is a band-satisfaction score and $\phi_{\mathrm{v}}$ is a vertical-support indicator.

Two target band configurations are used. For the single-pass-band task,
\begin{equation}
\Omega_{\mathrm{p}}^{(1)}=[2,4]\ \mathrm{kHz},
\qquad
\Omega_{\mathrm{g}}^{(1)}=[1,2]\cup[4,5]\ \mathrm{kHz}.
\end{equation}
For the dual-pass-band task,
\begin{equation}
\Omega_{\mathrm{p}}^{(2)}=[1,2]\cup[4,5]\ \mathrm{kHz},
\qquad
\Omega_{\mathrm{g}}^{(2)}=[2,4]\ \mathrm{kHz}.
\end{equation}
In both cases, the target is expressed by a transmission threshold of $-10$ dB. Frequencies above this threshold are regarded as belonging to a transmitted pass response, whereas frequencies below it are regarded as belonging to a suppressed stop response.

To reflect the relative widths of the frequency intervals, the score is constructed as a weighted band-satisfaction measure. For a general target configuration with pass regions $\{\Omega_{\mathrm{p},m}\}$ and gap regions $\{\Omega_{\mathrm{g},n}\}$, the score is written as
\begin{equation}
s_{\mathrm{FQ}}(\mathbf{x})
=
\sum_{n} \omega_{\mathrm{g},n}\,
\mathbb{P}_{f\in\Omega_{\mathrm{g},n}}
\left[T_{\mathbf{x}}(f)<T_{\mathrm{th}}\right]
\,+\,
\sum_{m} \omega_{\mathrm{p},m}\,
\mathbb{P}_{f\in\Omega_{\mathrm{p},m}}
\left[T_{\mathbf{x}}(f)\geq T_{\mathrm{th}}\right],
\end{equation}
where $T_{\mathbf{x}}(f)$ is the transmission response in dB, $T_{\mathrm{th}}=-10$ dB is the classification threshold, and $\mathbb{P}_{f\in\Omega}[\cdot]$ denotes the fraction of sampled frequencies in the interval $\Omega$ satisfying the specified inequality.

For the single-pass-band task, the weights are
\begin{equation}
(\omega_{\mathrm{g},1},\omega_{\mathrm{p},1},\omega_{\mathrm{g},2})
=(0.25,0.50,0.25),
\end{equation}
so that the full score sums to one over the two gap intervals and the central pass interval. For the dual-pass-band task, the weighting is reversed:
\begin{equation}
(\omega_{\mathrm{p},1},\omega_{\mathrm{g},1},\omega_{\mathrm{p},2})
=(0.25,0.50,0.25).
\end{equation}
The score is chosen because it measures target satisfaction directly at the level of the response function. It does not require gradient information, and it naturally accommodates discontinuous or threshold-based function-valued goals that are difficult to express through standard scalar surrogates.

\runinhead{Evaluation of the objective.}
The transmission response is evaluated by a finite-element frequency-domain model. The binary topology is interpreted as a two-dimensional elastic medium discretized on a regular quadrilateral mesh, with one bilinear quadrilateral element per pixel. Solid pixels are assigned linear elastic material properties, while void pixels are treated as zero-stiffness regions. Denoting the Young's modulus, Poisson's ratio, and density of the solid phase by $E_{\mathrm{s}}$, $\nu_{\mathrm{s}}$, and $\rho_{\mathrm{s}}$, the plane-stress constitutive matrix is
\begin{equation}
\mathbf{D}
=
\frac{E_{\mathrm{s}}}{1-\nu_{\mathrm{s}}^2}
\begin{bmatrix}
1 & \nu_{\mathrm{s}} & 0\\
\nu_{\mathrm{s}} & 1 & 0\\
0 & 0 & (1-\nu_{\mathrm{s}})/2
\end{bmatrix}.
\end{equation}
Using the standard strain-displacement matrix $\mathbf{B}$ of a bilinear quadrilateral element, the elemental stiffness matrix is
\begin{equation}
\mathbf{K}_e
=
\int_{\Omega_e}
\mathbf{B}^{\mathsf T}\mathbf{D}\mathbf{B}\,
\mathrm{d}\Omega,
\end{equation}
which is evaluated numerically by $2\times2$ Gauss quadrature. The consistent elemental mass matrix is written as
\begin{equation}
\mathbf{M}_e
=
\rho h^2
\mathbf{M}_e^{(0)},
\end{equation}
where $h=5\times10^{-4}$ m is the element size and $\mathbf{M}_e^{(0)}$ is the standard non-dimensional quadrilateral mass template. After assembling all elemental contributions, the global stiffness and mass matrices satisfy
\begin{equation}
\left(
\mathbf{K}
-\omega^2\mathbf{M}
\right)\mathbf{u}(\omega)
=
\mathbf{f}(\omega),
\end{equation}
where $\mathbf{K}$ and $\mathbf{M}$ are the assembled global stiffness and mass matrices and $\omega=2\pi f$ is the circular frequency.

The material parameters used in this case are Young's modulus $E_{\mathrm{s}}=50$ MPa, Poisson's ratio $\nu_{\mathrm{s}}=0.3$, and density $\rho_{\mathrm{s}}=1050$ kg/m$^3$. A damping factor $\eta_{\mathrm{damp}}=0.1$ is introduced in the modal response evaluation. The frequency range is sampled uniformly from $1$ kHz to $5$ kHz using 200 points, and the reduced dynamic response is represented by the lowest 50 modes. These settings define the transmission benchmark used throughout this case.

The excitation is applied from the lower boundary, and the transmitted response is measured from the upper boundary. The left and right boundaries are coupled periodically in the transverse direction, so the model behaves as a laterally repeated transmission channel. Let $\mathbf{T}_{\mathrm{red}}$ denote the reduction operator that enforces these boundary conditions and removes constrained degrees of freedom. The reduced matrices are
\begin{equation}
\mathbf{K}_r=\mathbf{T}_{\mathrm{red}}^{\mathsf T}\mathbf{K}\mathbf{T}_{\mathrm{red}},
\qquad
\mathbf{M}_r=\mathbf{T}_{\mathrm{red}}^{\mathsf T}\mathbf{M}\mathbf{T}_{\mathrm{red}}.
\end{equation}
The reduced free-vibration problem is then solved as
\begin{equation}
\mathbf{K}_r\boldsymbol{\psi}_n
=
\omega_n^2\mathbf{M}_r\boldsymbol{\psi}_n,
\qquad
n=1,\dots,N_m,
\end{equation}
where $N_m$ is the number of retained modes. The corresponding mode matrix is denoted by
\begin{equation}
\boldsymbol{\Psi}=
\begin{bmatrix}
\boldsymbol{\psi}_1 & \cdots & \boldsymbol{\psi}_{N_m}
\end{bmatrix}.
\end{equation}

To evaluate the forced response, the prescribed bottom-boundary motion is first converted into equivalent reduced load vectors associated with stiffness and inertia, denoted here by $\mathbf{g}_K$ and $\mathbf{g}_M$. The output response is then reconstructed by modal superposition. If $\boldsymbol{\psi}_{\mathrm{top}}$ collects the averaged modal amplitudes at the output boundary, the complex response at frequency $\omega$ is written as
\begin{equation}
u_{\mathrm{out}}(\omega)
=
\boldsymbol{\psi}_{\mathrm{top}}^{\mathsf T}
\left[
\frac{
-(1+\mathrm{i}\eta_{\mathrm{damp}})\mathbf{g}_K+\omega^2\mathbf{g}_M
}{
(1+\mathrm{i}\eta_{\mathrm{damp}})\boldsymbol{\omega}_n^2-\omega^2
}
\right]
+u_{\mathrm{res}},
\end{equation}
where the division is understood componentwise over the retained modes, $\boldsymbol{\omega}_n^2=[\omega_1^2,\dots,\omega_{N_m}^2]^{\mathsf T}$, and $u_{\mathrm{res}}$ is a static residual correction used to recover the non-modal contribution omitted by the truncated modal basis. The representation provides an efficient approximation of the full harmonic response over the entire frequency sweep.

The transmission amplitude is finally reported in decibels as
\begin{equation}
T_{\mathbf{x}}(f)
=
20\log_{10}\!\left(
\lvert u_{\mathrm{out}}(f)\rvert
\right),
\end{equation}
where $u_{\mathrm{out}}(f)$ denotes the averaged vertical displacement response measured at the output boundary under harmonic excitation. The resulting curve $T_{\mathbf{x}}(f)$ is then inserted into the score definition above. In practical terms, the objective evaluation follows four steps:

\textit{Step 1.} Convert the binary topology into a finite-element mesh and assemble $\mathbf{K}$ and $\mathbf{M}$ from the solid and void elements.

\textit{Step 2.} Enforce the periodic side coupling and the prescribed boundary motion to obtain the reduced matrices $\mathbf{K}_r$ and $\mathbf{M}_r$.

\textit{Step 3.} Solve the reduced eigenvalue problem, reconstruct $u_{\mathrm{out}}(\omega)$ by modal superposition with damping and residual correction, and compute the transmission spectrum $T_{\mathbf{x}}(f)$ over the sampled frequency range.

\textit{Step 4.} Evaluate the fractions of frequencies satisfying the pass-band and stop-band inequalities and combine them with the prescribed interval weights to obtain $s_{\mathrm{FQ}}(\mathbf{x})$.

Each candidate topology is thereby evaluated against the target transmission function over the full frequency range instead of at a single resonant frequency.

The objective is more difficult than the previous cases for two reasons. First, the mapping from topology to spectrum is highly nonlinear and depends on the coupled interaction of stiffness, inertia, and mode shapes. Second, the score itself is threshold-based and piecewise defined through pass-band and stop-band satisfaction. The problem is naturally suited to the black-box distribution-steering framework used in GenTO, since no analytical gradient of the score with respect to the topology is required.

\runinhead{Evaluation of the constraint.}
The feasibility constraint in this case is different from those used in the previous three benchmarks. Rather than requiring full two-directional periodic connectivity, the present task only requires a vertically supported transmission path. Since the excitation is applied from the bottom and the transmitted response is measured at the top, the most relevant structural requirement is the existence of a continuous load path between these two boundaries.

Let $\mathcal{C}_{\mathrm{m}}=\{C_1,\dots,C_{N_c}\}$ denote the connected material components identified using four-neighbor connectivity. As in the previous cases, left and right boundary contacts at matching vertical positions are treated as periodically connected, so that the structure remains laterally periodic. After this periodic merging step, let $\Pi(C_a)$ denote the resulting root class of component $C_a$.

The top-supported and bottom-supported root sets are then defined as
\begin{equation}
\mathcal{R}_{T}=\{\Pi(C_a): C_a\cap\partial\Omega_T\neq\varnothing\},
\qquad
\mathcal{R}_{B}=\{\Pi(C_a): C_a\cap\partial\Omega_B\neq\varnothing\}.
\end{equation}
The set of vertically spanning components is
\begin{equation}
\mathcal{R}_{TB}=\mathcal{R}_{T}\cap\mathcal{R}_{B},
\end{equation}
and the set of all periodic material root classes is
\begin{equation}
\mathcal{R}_{\mathrm{all}}=\{\Pi(C_a): C_a\in\mathcal{C}_{\mathrm{m}}\}.
\end{equation}
The vertical-support indicator is defined by
\begin{equation}
\phi_{\mathrm{v}}(\mathbf{x})=
\begin{cases}
1, & \mathcal{R}_{TB}\neq\varnothing
\text{ and }
\mathcal{R}_{\mathrm{all}}\subseteq (\mathcal{R}_{T}\cup\mathcal{R}_{B}),\\
0, & \text{otherwise.}
\end{cases}
\end{equation}

The condition has two roles. The requirement $\mathcal{R}_{TB}\neq\varnothing$ guarantees that at least one material network connects the input and output boundaries, so that vibration can in principle be transmitted through the structure. The additional condition $\mathcal{R}_{\mathrm{all}}\subseteq (\mathcal{R}_{T}\cup\mathcal{R}_{B})$ removes unsupported floating fragments that do not participate in the main transmission path. Together, these conditions define a physically meaningful class of candidate unit cells for the present benchmark.

\runinhead{Role of this case study.}
Case~4 plays a distinct role in the overall validation of GenTO because it moves beyond static property design into function-targeted optimization over an entire response spectrum. The optimization target is no longer a scalar extremum, a morphology descriptor, or a constitutive tensor, but a band-structured frequency response. As a result, this case tests whether the same reusable prior can be steered under a genuinely different class of downstream objective.

It is also an especially demanding benchmark from the optimization point of view. Different topologies can produce similar transmission levels at isolated frequencies while still differing strongly in their global spectral patterns. A successful method must shape the response over a whole interval, not simply improve one local quantity. The benchmark tests whether distribution steering can operate effectively in a non-convex, non-differentiable function-targeted design setting.

Finally, this case highlights one of the broadest claims of GenTO: the framework does not depend on a particular physics solver or on differentiable objective functions. Once a candidate topology can be sampled, screened, and assigned a score, the same steering mechanism can still be applied. The vibration-transmission benchmark tests the extent to which GenTO can act as a general black-box design engine across different physical domains.

\clearpage
\section{Baseline Methods}
\label{sec:si_baselines}

All comparison methods are formulated against the same case-specific design problems defined in Supplementary Section~\ref{sec:si_case_details}. Let $\mathcal{F}_c$ denote the feasible set of Case $c$, after accounting for the prescribed volume fraction, symmetry class, and all task-specific geometric or physical constraints. Let $J_c(\mathbf{x})$ denote the corresponding scalar score. Then the common comparison problem may be written as
\begin{equation}
\mathbf{x}^{\star}_{\mathcal{A},c}
\in
\arg\max_{\mathbf{x}\in \mathcal{F}_c\cap \mathcal{A}_c}
J_c(\mathbf{x}),
\end{equation}
where $\mathcal{A}_c$ denotes the admissible search set associated with a particular baseline method. For minimization tasks, the original error is rewritten in score form by negation. For multi-objective tasks, the same notation is understood in the Pareto sense, so that the comparison concerns the non-dominated set, not one scalar optimum.

Under this notation, the baselines differ in two essential aspects. First, they do not search over the same design space. Some methods are restricted to previously generated data, some are allowed to sample from a fixed pretrained generator, some move a continuous density field by sensitivity updates, and some evolve a population of morphology fields. Second, they do not update candidates in the same way. Retrieval baselines do not update at all, prior baselines only rank samples from a fixed distribution, classical topology optimization follows one continuous optimization trajectory, and genetic algorithms evolve a population through selection, crossover, mutation, and immigration. This section introduces these baselines in a unified form so that the comparisons in the following experimental section can be interpreted more clearly.

\subsection{Best in Training Sets}
\label{sec:si_baseline_training}
The first baseline searches only within the available pretraining dataset. Let
\begin{equation}
\mathcal{D}_{\mathrm{train},c}
=
\left\{
\mathbf{x}^{(n)}
\right\}_{n=1}^{N_c}
\end{equation}
denote the subset of the topology dataset compatible with the symmetry requirement of Case $c$. The corresponding reference solution is
\begin{equation}
\mathbf{x}^{\star}_{\mathrm{train},c}
\in
\arg\max_{\mathbf{x}\in \mathcal{F}_c\cap \mathcal{D}_{\mathrm{train},c}}
J_c(\mathbf{x}).
\end{equation}
For multi-objective tasks, the comparison is made through the non-dominated subset
\begin{equation}
\mathcal{P}_{\mathrm{train},c}
=
\left\{
\mathbf{x}\in\mathcal{F}_c\cap \mathcal{D}_{\mathrm{train},c}:
\mathbf{x}\ \text{is non-dominated}
\right\}.
\end{equation}

The baseline does not perform optimization in the usual sense. It only identifies the best feasible structure already contained in the known dataset. Its role is therefore to establish the best performance already available when no new topology is generated and no search is performed beyond the pretraining samples. The comparison with GenTO reveals whether the final results merely retrieve favorable known structures or genuinely improve beyond the best examples already present in the training set.

\subsection{Best in Prior}
\label{sec:si_baseline_prior}
The second baseline searches within the support of the pretrained generator without any task-specific adaptation. Let $p_0(\mathbf{z})$ denote the pretrained prior in SDF space, and let $\mathcal{B}_{V_f}$ denote the fixed-volume binarization operator introduced in Supplementary Section~\ref{sec:si_model_architecture}. Sampling from the pretrained model gives
\begin{equation}
\mathbf{z}^{(m)}\sim p_0(\mathbf{z}),
\qquad
\mathbf{x}^{(m)}=\mathcal{B}_{V_f}\!\left(\mathbf{z}^{(m)}\right),
\end{equation}
and the admissible search set is
\begin{equation}
\mathcal{S}_{0,c}
=
\left\{
\mathbf{x}^{(m)}
\right\}_{m=1}^{M_c}.
\end{equation}
The baseline is therefore written as
\begin{equation}
\mathbf{x}^{\star}_{\mathrm{prior},c}
\in
\arg\max_{\mathbf{x}\in \mathcal{F}_c\cap \mathcal{S}_{0,c}}
J_c(\mathbf{x}),
\end{equation}
or, for multi-objective tasks, the non-dominated subset of $\mathcal{F}_c\cap\mathcal{S}_{0,c}$.

The baseline is stronger than the previous one because it is not limited to a finite topology library. It can generate new candidate structures by sampling from the learned prior. However, the prior itself remains fixed throughout the process. No steering, fine-tuning, or task-dependent distribution update is performed. The comparison with GenTO therefore isolates the value of iterative prior adaptation: it separates the contribution of the pretrained generator itself from the contribution of steering the generator toward a task-specific high-performing region.

\subsection{Traditional Topology Optimization}
\label{sec:si_baseline_to}
The third baseline is a classical density-based topology optimization method, following established topology-optimization and homogenization-based design formulations~\cite{bendsoe1988generating,allaire2002shape,sigmund200199}. Instead of searching directly in binary topology space, it introduces a continuous density field
\begin{equation}
\boldsymbol{\rho}\in[0,1]^{N\times N},
\end{equation}
whose entries represent local material densities. The corresponding physical design is obtained through a projection operator $\Pi$, so that the optimization is written in the general form
\begin{equation}
\boldsymbol{\rho}^{\star}_c
\in
\arg\max_{\boldsymbol{\rho}}
J_c\!\left(\Pi(\boldsymbol{\rho})\right)
\qquad
\textrm{s.t.}
\qquad
\Pi(\boldsymbol{\rho})\in\mathcal{F}_c.
\end{equation}

The physical response is evaluated by finite-element analysis, and the sensitivities are computed with respect to the density field. In the density-based formulation used here, the elemental stiffness is interpolated by a penalized density law of the form
\begin{equation}
\mathbf{K}_e(\rho_e)
=
\rho_e^{\,p_{\mathrm{SIMP}}}\mathbf{K}_e^{(0)},
\end{equation}
where $\mathbf{K}_e^{(0)}$ is the full-solid elemental stiffness matrix and $p_{\mathrm{SIMP}}$ is the penalization exponent. The objective and its sensitivities are then assembled from the finite-element solution, followed by sensitivity filtering to suppress checkerboard patterns and mesh-scale oscillations. Denoting the filtered sensitivity by $\widetilde{\nabla}J_c$, the generic update step may be written as
\begin{equation}
\boldsymbol{\rho}^{(k+1)}
=
\mathcal{U}_{\mathrm{TO}}
\left(
\boldsymbol{\rho}^{(k)},
\widetilde{\nabla}J_c(\boldsymbol{\rho}^{(k)})
\right),
\end{equation}
subject to the admissible density bounds and the prescribed volume fraction.

For maximization-type tasks, the implementation follows the optimality-criteria philosophy, in which each density variable is rescaled according to the local sensitivity while a Lagrange multiplier enforces the volume constraint. In schematic form,
\begin{equation}
\rho_e^{(k+1)}
=
\Pi_{[\rho_{\min},1]}
\left[
\rho_e^{(k)}
\left(
\frac{-\partial J_c/\partial \rho_e}{\lambda}
\right)^{\gamma}
\right],
\end{equation}
where $\lambda$ is chosen so that the volume constraint is satisfied, $\gamma$ is a damping exponent, and $\Pi_{[\rho_{\min},1]}$ denotes clipping to the admissible density interval.

The topology-optimization baseline represents the classical trajectory-based paradigm. It does not rely on any learned topology prior and instead updates one continuous design field through physics-based sensitivity information. The comparison with GenTO therefore highlights the difference between classical task-specific optimization and prior-guided distribution steering.

From a broader methodological perspective, this baseline is representative of gradient- or sensitivity-based design methods. Its main strength is that it exploits local derivative information efficiently when the governing physics is differentiable and the constraint structure is well defined. For such problems, topology optimization can often produce strong task-specific solutions with relatively direct use of the underlying physics. However, it usually follows a single optimization trajectory from a chosen initial design, and its behavior may therefore depend on the initialization, regularization settings, and the local non-convexity of the design landscape. It is also tightly tied to a particular physical formulation and sensitivity derivation. In this sense, topology optimization provides a useful contrast to GenTO: the former improves one design field by gradient-driven local updates, whereas the latter adapts a learned topology distribution and searches by repeated sampling, selection, and fine-tuning.

\subsection{Genetic Algorithm}
\label{sec:si_baseline_ga}
The fourth baseline is a population-based genetic algorithm operating in a continuous morphology-field representation~\cite{holland1975adaptation,goldberg1989genetic}. Rather than evolving binary topologies directly, the algorithm maintains a population of grayscale morphology fields at generation $\ell$,
\begin{equation}
\mathcal{P}^{(\ell)}
=
\left\{
\mathbf{g}^{(\ell)}_m
\right\}_{m=1}^{N_p},
\qquad
\mathbf{g}^{(\ell)}_m\in\mathbb{R}^{N\times N},
\end{equation}
which are converted into binary topologies by fixed-volume thresholding,
\begin{equation}
\mathbf{x}^{(\ell)}_m
=
\mathcal{B}_{V_f}\!\left(\mathbf{g}^{(\ell)}_m\right).
\end{equation}
Each generation is evaluated by the same feasibility filters and objective functions used for GenTO, so that only the search mechanism differs.

The initial population is seeded from the pretrained generator instead of sampled from an unconstrained random binary distribution. Accordingly,
\begin{equation}
\mathbf{g}^{(0)}_m \sim p_0(\mathbf{g}),
\end{equation}
after which candidates are screened and high-scoring valid seeds are favored or retained. Each subsequent generation consists of three parts: elites preserved from the previous generation, offspring produced by crossover and mutation, and a set of new immigrants sampled again from the fixed pretrained generator. The population update is therefore written as
\begin{equation}
\mathcal{P}^{(\ell+1)}
=
\mathcal{E}^{(\ell)}
\cup
\mathcal{C}^{(\ell)}
\cup
\mathcal{I}^{(\ell)}.
\end{equation}

Parent selection is rank- or Pareto-rank-based and gives preference to high-scoring feasible individuals, following the elitist selection philosophy used in NSGA-II-style multi-objective genetic algorithms~\cite{deb2002fast}. If $w_m^{(\ell)}$ denotes the corresponding sampling weight, then
\begin{equation}
\mathbb{P}\!\left(
\mathbf{g}^{(\ell)}_m\ \text{is selected}
\right)
\propto
w_m^{(\ell)}.
\end{equation}
Given two selected parents, crossover is performed directly in the continuous field space by convex mixing,
\begin{equation}
\mathbf{g}_{\mathrm{child}}
=
\alpha\,\mathbf{g}_{a}^{(\ell)}
\,+\,
(1-\alpha)\mathbf{g}_{b}^{(\ell)},
\qquad
\alpha\sim\mathcal{U}[\alpha_{\min},\alpha_{\max}],
\end{equation}
which preserves the smooth morphology-field representation induced by the generator.

The crossover process is illustrated in Supplementary Fig.~\ref{fig:SI_F5}. Two parent topologies are first represented in the corresponding SDF form, after which a child field is obtained by continuous interpolation between the two parent fields. Different values of $\alpha$ lead to different geometric combinations of the parent patterns. After interpolation, the blended field is converted back to a binary topology by the same fixed-volume thresholding rule used throughout the optimization. This representation is more suitable than direct pixel-wise crossover on binary images, because it preserves coherent boundaries and yields smoother morphological transitions between parent and child structures.

\begin{figure}[t]
\centering
\includegraphics[width=0.95\textwidth]{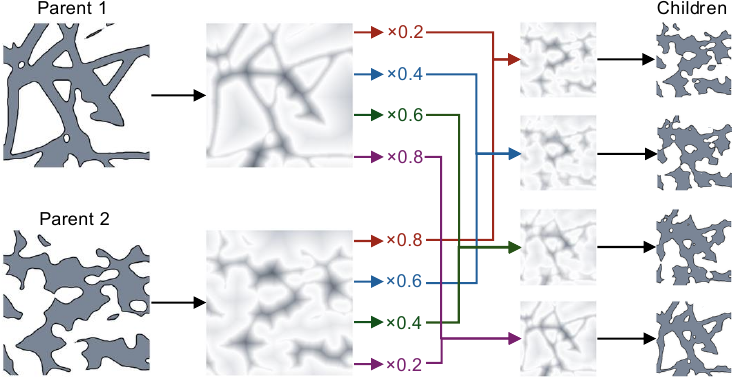}
\caption{\textbf{Schematic illustration of the crossover operation.} Two parent topologies are first expressed in the SDF representation, and the child field is then obtained by continuous interpolation between the two parent fields with different mixing ratios. The interpolated field is finally mapped back to a binary topology by fixed-volume thresholding, producing offspring structures that inherit geometric characteristics from both parents while remaining in a smooth morphology space.}
\label{fig:SI_F5}
\end{figure}

Mutation is introduced by adding a periodic Gaussian random field perturbation,
\begin{equation}
\mathbf{g}_{\mathrm{child}}
\leftarrow
\mathbf{g}_{\mathrm{child}}
\,+\,
\sigma_{\mathrm{mut}}
\boldsymbol{\zeta}_{\mathrm{GRF}},
\end{equation}
where $\boldsymbol{\zeta}_{\mathrm{GRF}}$ is a zero-mean correlated random field with prescribed correlation length. This mutation operator perturbs morphology smoothly instead of by isolated pixel flips, which makes it better aligned with the continuous representation of the pretrained generator. In addition, a fraction of fresh candidates is injected from the fixed prior,
\begin{equation}
\mathcal{I}^{(\ell)}\sim p_0(\mathbf{g}),
\end{equation}
to maintain exploration and reduce premature population collapse.

This genetic-algorithm baseline is more informative than a purely random evolutionary search because it is already generator-aware: it explores the morphology space induced by the pretrained prior. However, unlike GenTO, the generator itself never changes. The population evolves while the prior remains fixed. The comparison with GenTO therefore isolates a central methodological difference: GA explores around a fixed learned prior, whereas GenTO updates the learned prior itself through iterative fine-tuning.

More generally, this baseline is representative of black-box population-based optimization. It does not require gradient information and can therefore be applied to rugged, discontinuous, or weakly differentiable objectives for which gradient-based updates are difficult to derive or unreliable to use. This flexibility makes genetic algorithms broadly applicable across very different design problems. Their main limitation is that they often require a large number of function evaluations to achieve strong performance, and their efficiency depends strongly on the population size, variation operators, and the balance between exploration and exploitation. In the present generator-guided form, the initialization and immigration come from the pretrained prior, while crossover and mutation operate in the same continuous morphology-field space. Nevertheless, this prior remains static throughout the search. The difference from GenTO is therefore fundamental: GA uses the prior as a fixed search environment, whereas GenTO progressively updates the prior itself so that the entire sampling distribution moves toward the task-specific high-performing region.

\clearpage
\section{Additional Experimental Results}
\label{sec:si_additional_results}


The GenTO runs in the four case studies follow the same steering loop: candidate topologies are sampled from the current generator, valid samples are screened and scored, an elite subset is retained, and the generator is fine-tuned on this subset before the next iteration. Unless otherwise stated, each iteration collects $2N_{\mathrm{take}}$ valid candidates and keeps $N_{\mathrm{take}}$ elite samples for fine-tuning. The main run parameters used for the four cases are summarized in Supplementary Table~\ref{tab:SI_P5_T0}. For fair comparison, the population-based baselines are evaluated under the same evaluation budget as GenTO; the definition and case-specific values of these budgets are provided in Supplementary Section~\ref{sec:si_discussion_efficiency} and Supplementary Tables~\ref{tab:SI_eval_budget_definition} and \ref{tab:SI_eval_budget}.

\begin{table}[htbp!]
\centering
\caption{\textbf{GenTO steering parameters used in the four case studies.} The table lists the target volume fraction, the number of steering iterations, the number of elite samples retained after each screening step, and the generation/training batch sizes. ``ES'' denotes elasticity target.}
\label{tab:SI_P5_T0}
\scriptsize
\setlength{\tabcolsep}{1.6pt}
\begin{tabular*}{\textwidth}{@{\extracolsep{\fill}}lllcccccc@{}}
\toprule
Case & Task & Prior/sym. & $V_f$ & Iter. & Elite & Batch & Ep. & LR \\
\midrule
Case~1 & Thermal ext. & Periodic & 0.4 & 100 & 64 & 64/64 & 100 & $10^{-4}$ \\
Case~2 & MO morph. & Periodic & 0.5 & 50 & 2048 & 64/64 & 10 & $5\times10^{-6}$ \\
Case~3 & Periodic ES & Periodic & 0.5 & 100 & 256 & 64/64 & 100 & $5\times10^{-5}$ \\
Case~3 & 4-fold ES & 4-fold & 0.5 & 100 & 256 & 64/64 & 100 & $5\times10^{-5}$ \\
Case~3 & 8-fold ES & 8-fold & 0.5 & 100 & 256 & 64/64 & 50 & $10^{-5}$ \\
Case~4 & Single pass & 4-fold & 0.5 & 100 & 256 & 64/64 & 100 & $10^{-5}$ \\
Case~4 & Dual pass & 4-fold & 0.5 & 100 & 256 & 64/64 & 100 & $10^{-5}$ \\
\bottomrule
\end{tabular*}
\end{table}

\subsection{Comparative Results for Thermal Conductivity Extremization}
\label{sec:si_case1_additional}

\runinhead{Comparison setup.}
Case~1 is used here as the main benchmark to compare the five methods introduced in Supplementary Section~\ref{sec:si_baselines}, namely Best in training sets, Best in Prior, traditional topology optimization (TO), genetic algorithm (GA), and GenTO. All methods are evaluated using the same Case~1 objective and the same two geometric constraints introduced in Supplementary Section~\ref{sec:si_case1}. All candidate structures are first normalized to the same target material fraction $V_f=0.4$, and then evaluated by the same FFT-based conductivity solver. The comparison mainly reflects the difference in search strategy, not differences in material usage or physics evaluation.

Two tasks are considered here, namely conductivity maximization and conductivity minimization. For the training-set and fixed-prior baselines, the comparison is based on the best retrieved topology and, when applicable, the mean score of the top 16 valid candidates. The same top-16 statistic is also used for GA and GenTO, because both methods return a set of valid samples. For TO, only the best score is reported.

\begin{figure}[htbp!]
\centering
\includegraphics{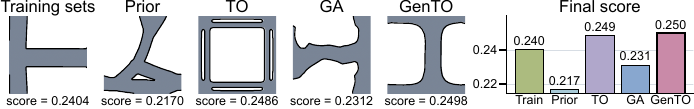}
\caption{\textbf{Comparison of conductivity maximization in Case~1}. Each panel shows the best topology returned by the corresponding method under the same volume fraction and constraint settings.}
\label{fig:SI_F6}
\end{figure}

\begin{figure}[htbp!]
\centering
\includegraphics{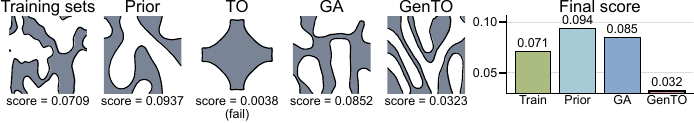}
\caption{\textbf{Comparison of conductivity minimization in Case~1.} The corresponding TO result is shown for reference, but its best returned structure does not satisfy the final feasibility check.}
\label{fig:SI_F7}
\end{figure}

\begin{table}[htbp!]
\centering
\caption{\textbf{Quantitative comparison of the five methods for Case~1.} The \emph{mean top-16} statistic is not applicable to TO because TO follows a single deterministic optimization trajectory instead of generating a population of candidates.}
\label{tab:SI_P5_T1}
\small
\begin{tabular}{lcccc}
\hline
Method & Max. best & Max. mean & Min. best & Min. mean \\
\hline
Best in training sets & 0.2404 & 0.2321 & 0.0709 & 0.0907 \\
Best in Prior & 0.2170 & 0.1906 & 0.0937 & 0.1375 \\
TO & 0.2493 & / & fail & / \\
GA & 0.2346 & 0.2313 & 0.0568 & 0.0913 \\
GenTO & 0.2498 & 0.2497 & 0.0323 & 0.0333 \\
\hline
\end{tabular}
\end{table}

\runinhead{Results for conductivity maximization.}
The conductivity-maximization results are summarized in Supplementary Fig.~\ref{fig:SI_F6} and Supplementary Table~\ref{tab:SI_P5_T1}. The best topology already contained in the pretraining datasets reaches a score of 0.2404, indicating that the training data include several topology patterns that are favorable for heat transport. In contrast, direct sampling from the fixed pretrained prior gives a lower best score of 0.2170, showing that the reusable prior alone is not enough to focus the search on the best conductive region.

Among the iterative baselines, GA improves over the fixed-prior reference and reaches 0.2346, but it is still below the best result already contained in the datasets. TO and GenTO give the two highest valid scores, namely 0.2493 and 0.2498. The difference between these two best scores is small. However, GenTO also gives the highest mean score of the top 16 valid candidates, namely 0.2497. Thus, the improvement of GenTO is not due to one extreme sample only; nearby sampled designs are also consistently strong.

\begin{figure}[htbp!]
\centering
\includegraphics{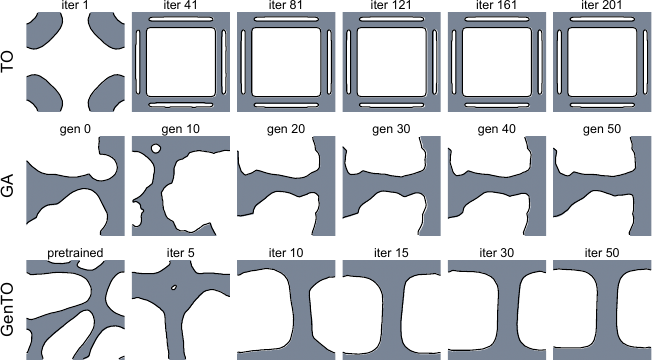}
\caption{\textbf{Historical evolution of the best topology for conductivity maximization.} Each row corresponds to one iterative method, and the panels show selected stages along the optimization history.}
\label{fig:SI_F8}
\end{figure}

\begin{figure}[htbp!]
\centering
\includegraphics[width=11.64cm]{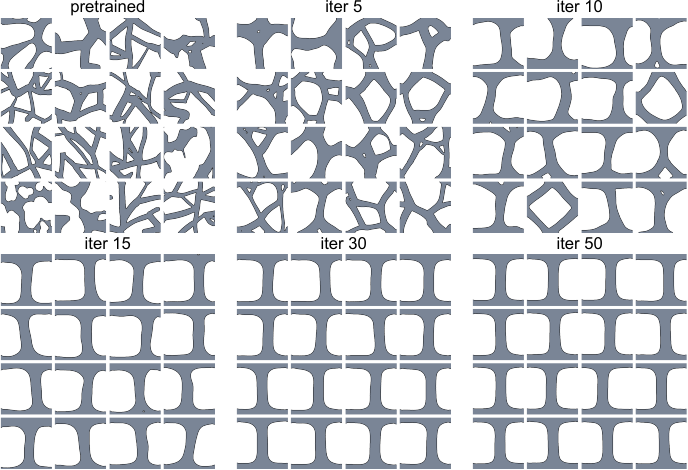}
\caption{\textbf{Evolution of the sampled $4\times4$ topology grid from GenTO for conductivity maximization.} The panels correspond to selected checkpoints of the steered prior.}
\label{fig:SI_F9}
\end{figure}

The structural histories in Supplementary Fig.~\ref{fig:SI_F8} further show how the three iterative methods move toward the maximization target. TO follows one deterministic path and gradually builds a strong transport network. GA also improves the designs step by step, but its evolution is more scattered because crossover and mutation act on a population instead of on one continuous path. GenTO shows a different behavior. Already at the early checkpoints, especially around iterations 5, 10, and 15, the best samples shift quickly toward layouts with stronger and more direct conductive channels. The later checkpoints mainly refine these favorable patterns.

The same trend can be seen more clearly in the distribution-level visualization in Supplementary Fig.~\ref{fig:SI_F9}. At the pretrained stage, the sampled $4\times4$ grid still covers a broad set of generic periodic morphologies. After steering, the generated samples become more concentrated around topologies with continuous material backbones and direct transport paths. Thus, the maximization task is improved at the level of one best design and across the sampled topology distribution, which also moves toward a high-conductivity region.

\runinhead{Results for conductivity minimization.}
The minimization task is more difficult than the maximization task since low conductivity must be achieved without violating periodic connectivity or the minimum-feature-size requirement. This gives a narrower feasible design space and makes the search more sensitive to invalid disconnected layouts. The final results are shown in Supplementary Fig.~\ref{fig:SI_F7} and Supplementary Table~\ref{tab:SI_P5_T1}. Here, the best topology found directly in the datasets gives a score of 0.0709, whereas direct sampling from the fixed prior only reaches 0.0937. The gap is even larger than in the maximization case. This again shows that the fixed pretrained prior is not naturally aligned with the target task.

GA improves the result to 0.0568 and clearly outperforms both static references. The TO topology fails the final feasibility check and is therefore not counted as a valid solution. Among the valid methods, GenTO gives the lowest best score, namely 0.0323, and also the lowest top-16 mean score, namely 0.0333. The small difference between these two values suggests that the GenTO distribution has become concentrated around feasible low-conductivity topologies, instead of producing only one accidental outlier.

\begin{figure}[htbp!]
\centering
\includegraphics{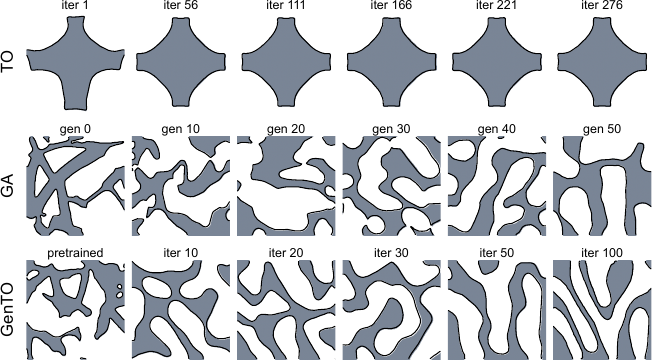}
\caption{\textbf{Historical evolution of the best topology for conductivity minimization.} Each row corresponds to one iterative method, and the panels show selected stages along the optimization history.}
\label{fig:SI_F10}
\end{figure}

\begin{figure}[htbp!]
\centering
\includegraphics[width=11.64cm]{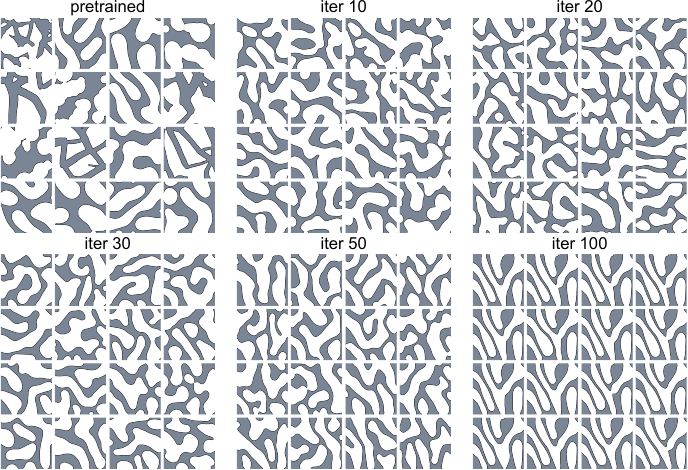}
\caption{\textbf{Evolution of the sampled $4\times4$ topology grid for conductivity minimization.} The panels correspond to selected checkpoints of the steered prior.}
\label{fig:SI_F11}
\end{figure}

The history plots in Supplementary Fig.~\ref{fig:SI_F10} show that minimization needs a different structural evolution from maximization. TO tends to move toward layouts that strongly block direct transport, but this also increases the risk of producing invalid or nearly disconnected structures. GA produces more diverse blocking patterns and gives better valid scores than the static baselines, but its evolution remains relatively broad. GenTO changes more gradually than in the maximization task, but the checkpoint sequence still shows a clear direction. The best samples evolve from broadly connected periodic morphologies into sparse, tortuous, and transport-blocking layouts that still satisfy the same connectivity and feature-size rules.

The $4\times4$ sample grids in Supplementary Fig.~\ref{fig:SI_F11} show the same trend at the distribution level. The pretrained prior still produces many samples with relatively direct material pathways, which explains its weak performance on the minimization task. After steering, these direct pathways are gradually suppressed, and the sampled population becomes dominated by more insulating but still valid structures. This is the expected behavior of distribution steering: the method does not search only for one extreme design, but moves the whole topology distribution toward a feasible low-conductivity region.

\runinhead{Discussion of the benchmark results.}
Overall, the five-way comparison separates several possible sources of improvement. First, GenTO exceeds the best result already present in the training datasets in both tasks, so the method is not simply retrieving good training samples. Second, GenTO also improves clearly over the Best in Prior baseline, so the gain does not come only from having a pretrained generator. Third, the comparison with GA shows that the benefit is not solely due to population search in a fixed morphology space. Keeping the prior fixed still limits how far the search can move toward task-specific topology regions. Finally, the comparison with TO shows a complementary strength. In maximization, GenTO reaches a best score very close to the strongest valid TO solution. In minimization, it gives the strongest valid result among all methods considered here.

The combination of the best-sample histories and the $4\times4$ sample-grid histories shows that GenTO improves the sampling distribution as well as one design. The small difference between the best score and the top-16 mean score of GenTO in both tasks gives the same message in quantitative form. The benchmark supports the main claim of the paper: the advantage of GenTO comes from steering a reusable topology prior toward a task-specific high-performing region, not from direct retrieval, fixed-prior sampling, or a conventional optimizer working in a static design space.

\subsection{Additional Results of Case 2}
\label{sec:si_case2_additional}

\runinhead{Comparison setup.}
Case~2 is used here as a multi-objective benchmark in morphology space. In contrast to Case~1, the comparison is centered on the quality of the final Pareto set instead of on one scalar optimum. Four methods are compared, namely Best in training sets, Best in Prior, genetic algorithm (GA), and GenTO. All methods follow the same periodic setting used for the main-text figure of Case~2. All candidate structures are normalized to the same target material fraction $V_f=0.5$, and then evaluated by the same periodic-connectivity constraint and the same two objectives introduced in Supplementary Section~\ref{sec:si_case2}, namely the boundary fractal dimension and the minimum feature size. The comparison again mainly reflects the difference in search strategy, not differences in material usage or evaluation rules.

The setup of this case is also different from that of Case~1 in a more basic way. In Case~1, all valid structures can be ranked by one conductivity-oriented score, so the comparison is naturally written in terms of one best design and one optimization history. In Case~2, this is no longer possible. The minimum feature size, which acted as a geometric constraint in Case~1, now becomes one of the two objectives. The second objective favors complex boundaries with larger fractal dimension. These two tendencies compete with each other. Large feature size usually favors thicker and smoother morphologies, while large fractal dimension usually favors finer and more irregular boundaries. For this reason, the goal is to find a broad non-dominated set that covers different trade-offs between the two objectives, not one single best topology.

This point also changes how the benchmark should be read. For a single-objective problem, improvement can often be judged by whether one curve goes higher or whether one final score is larger. For a multi-objective problem, improvement means that the final front moves outward, that a wider trade-off region becomes available, or that the representative set covers a richer morphology range. A later checkpoint or generation is not expected to improve every structure in the same way. Instead, the optimization process redistributes the sampled designs along the Pareto front and gradually improves the frontier as a whole.

In this case, the comparison is based on the final Pareto front and on representative Pareto samples. For the training-set and fixed-prior baselines, the reported front is the non-dominated set extracted from the evaluated candidates. For GA and GenTO, the comparison is based on the final Pareto set returned by each iterative process. Traditional topology optimization is not included here, because the present benchmark is centered on methods that naturally return a Pareto set in morphology space. In the topology visualizations below, the representative samples are arranged along the final front from the coarse-feature side to the high-fractal side, so that the change of morphology across the trade-off can be read directly from left to right.

\begin{figure}[htbp!]
\centering
\IfFileExists{Figures/SI_F12.pdf}{%
\includegraphics[width=7.38cm]{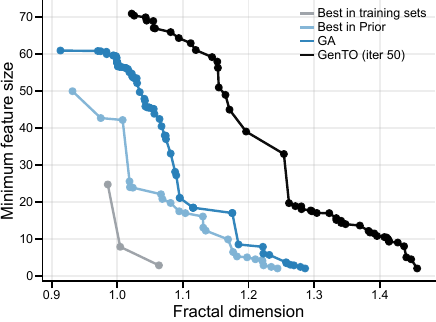}%
}{%
\fbox{\rule{0pt}{2.2in}\rule{0.70\textwidth}{0pt}}%
}
\caption{\textbf{Final Pareto-front comparison of the four methods for Case~2 in the periodic setting.} Each curve shows the extracted final Pareto front in the space of fractal dimension and minimum feature size.}
\label{fig:SI_F12}
\end{figure}

\begin{table}[htbp!]
\centering
\caption{\textbf{Quantitative comparison of the four methods for Case~2.} \emph{Best fractal} and \emph{best feature} denote the outermost values reached by the final extracted Pareto front. The normalized hypervolume measures the area dominated by the Pareto front in the min--max normalized two-objective space; larger values indicate broader and more outward Pareto coverage.}
\label{tab:SI_P5_T2}
\small
\setlength{\tabcolsep}{2.2pt}
\begin{tabular}{@{}lcccc@{}}
\hline
Method & Best fractal & Best feature & Pareto samples & Norm. hypervolume \\
\hline
Best in training sets & 1.0634 & 24.7386 & 3 & 0.0480 \\
Best in Prior & 1.2440 & 50.0000 & 24 & 0.1773 \\
GA & 1.2861 & 61.0000 & 71 & 0.2851 \\
GenTO & 1.4567 & 71.0000 & 48 & 0.5723 \\
\hline
\end{tabular}
\end{table}

\runinhead{Final Pareto-front comparison.}
The final Pareto fronts are shown in Supplementary Fig.~\ref{fig:SI_F12}, and the corresponding summary is given in Supplementary Table~\ref{tab:SI_P5_T2}. Here, normalized hypervolume is used as a scalar measure of Pareto-front coverage: after normalizing both objectives, it measures the area dominated by the extracted front relative to a common reference point. The front extracted from the training-set baseline is very limited. In the evaluated set, it only reaches 1.0634 in fractal dimension and 24.7386 in minimum feature size, with a normalized hypervolume of 0.0480. This indicates that direct retrieval from the pretraining data gives only a narrow trade-off region for this morphology-control problem.

Direct sampling from the fixed pretrained prior already improves the result clearly. The best fractal dimension increases to 1.2440, the best minimum feature size increases to 50.0000, and the normalized hypervolume increases to 0.1773. This indicates that the reusable prior itself already contains useful morphology knowledge beyond the few strong samples explicitly present in the training-set baseline. However, the front returned by the fixed prior is still clearly inside the final GenTO front.

With the larger GA population, GA returns a stronger fixed-prior population-search baseline. Its final front reaches 1.2861 in fractal dimension and 61.0000 in minimum feature size, improving over the fixed-prior reference at both extremes, with a normalized hypervolume of 0.2851. However, the final front is still less balanced than the GenTO front, especially on the high-fractal side. GenTO gives the outermost final front among the four methods. Its extracted final front reaches 1.4567 in fractal dimension and 71.0000 in minimum feature size, and its normalized hypervolume reaches 0.5723. The improvement in Case~2 does not come only from retrieving good structures from the datasets or from sampling repeatedly from a fixed prior. The steering step is needed to move the sampled front further outward in morphology space and to improve both ends of the trade-off together.

\begin{figure}[htbp!]
\centering
\IfFileExists{Figures/SI_F13.pdf}{%
\includegraphics[width=11.56cm]{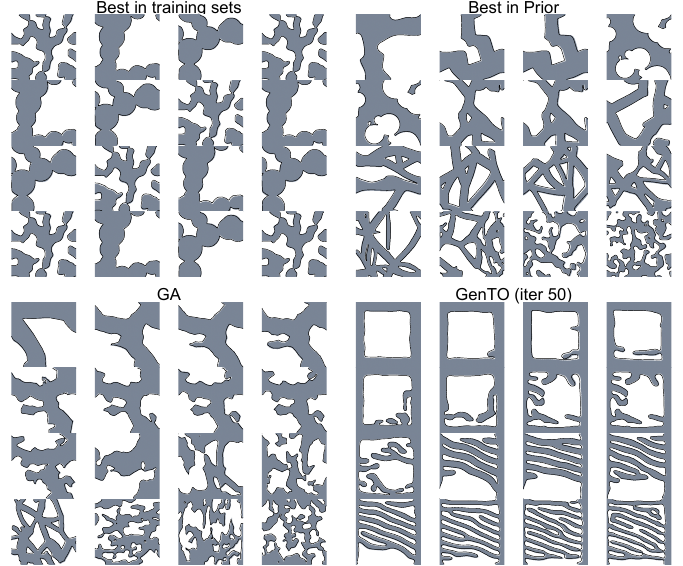}%
}{%
\fbox{\rule{0pt}{3.3in}\rule{0.94\textwidth}{0pt}}%
}
\caption{\textbf{Representative Pareto topologies returned by the four methods for Case~2.} Each panel shows a representative $4\times4$ sample grid extracted from the final Pareto set of the corresponding method.}
\label{fig:SI_F13}
\end{figure}

\runinhead{Representative topology comparison.}
The representative Pareto samples in Supplementary Fig.~\ref{fig:SI_F13} make the same trend easier to interpret in structural form. The training-set baseline gives only a very small number of non-dominated structures, so its trade-off family is narrow and incomplete. The fixed prior produces a broader set of periodic morphologies and already covers both coarse-feature and more complex boundary patterns. GA further expands the coarse-feature side of the trade-off, and many of its representative samples remain relatively thick and well separated. This agrees with the front comparison above: GA is effective in enlarging feature size, but its structural family is still biased toward a relatively limited side of the morphology space.

GenTO shows the broadest overall transition across the trade-off. Within one representative panel, the sampled structures span from large and smooth features to much finer and more irregular boundaries. The steered prior reaches better objective values at a few isolated points while maintaining a broader family of high-performing Pareto designs across the morphology space. The visual difference between the four methods is qualitative as well as quantitative, because the final Pareto sets are associated with different ranges of morphological diversity.

\begin{figure}[htbp!]
\centering
\IfFileExists{Figures/SI_F14.pdf}{%
\includegraphics[width=11.75cm]{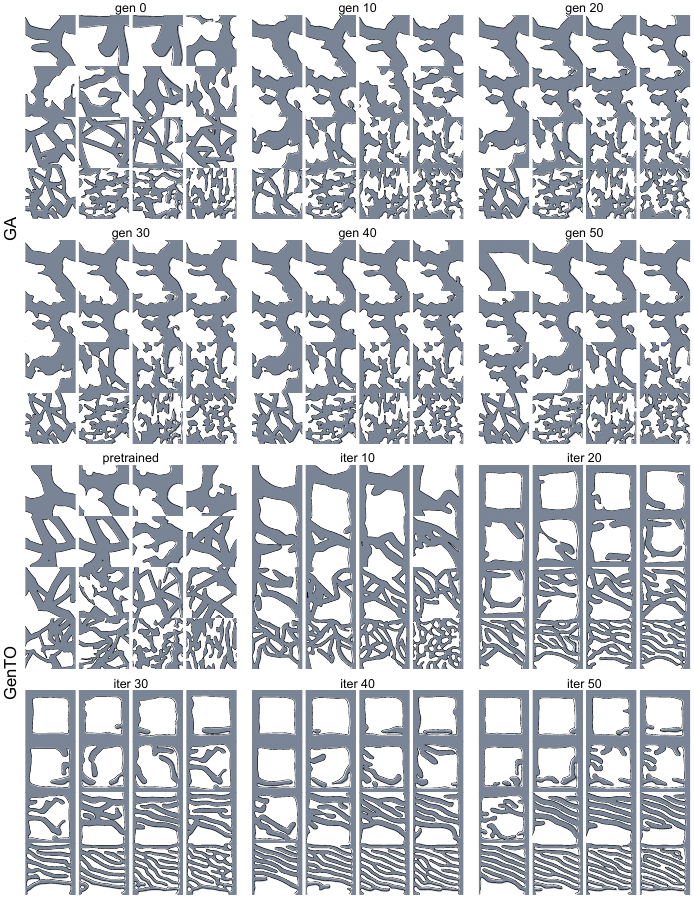}%
}{%
\fbox{\rule{0pt}{2.8in}\rule{0.94\textwidth}{0pt}}%
}
\caption{\textbf{Evolution of representative Pareto sets for GA and GenTO in Case~2.} The panels show selected generations or checkpoints of the two iterative methods.}
\label{fig:SI_F14}
\end{figure}

\runinhead{Evolution of the Pareto search.}
The history comparison in Supplementary Fig.~\ref{fig:SI_F14} further shows the difference between GA and GenTO. Here the history should again be read in a multi-objective sense. A later stage does not simply mean one better scalar value. Instead, it means that the representative set and the frontier behind it are being reshaped across the trade-off. GA improves the population gradually over generations. The representative sets change step by step, and the coarse-feature end of the trade-off appears early. However, the expansion toward more complex high-fractal morphologies is relatively limited, and the population remains concentrated around a narrower structural family.

GenTO shows a different evolution. Already from the pretrained stage to the early checkpoints, the representative set moves away from generic periodic layouts and begins to cover a much broader range of trade-offs. Later checkpoints mainly refine this front and stabilize the sampled morphology family. Unlike the single-objective case, the evolution here is not expected to be monotonic in one scalar score, because improving a Pareto set requires rebalancing two competing objectives. The checkpoint sequence should therefore be read as a redistribution of samples along the trade-off, not as a steady increase of one number.

\runinhead{Discussion of the benchmark results.}
Overall, the four-way comparison again separates several possible sources of improvement. First, GenTO clearly exceeds the trade-off region already present in the sampled training-set baseline, so the method is not simply retrieving a few strong pre-existing samples. Second, GenTO also improves markedly over the fixed-prior baseline, so the gain does not come only from having a reusable pretrained generator. Third, the comparison with GA shows that fixed-prior population search is helpful but still limited. GA gives a stronger coarse-feature frontier than the static references, but it does not reach the same outer boundary in fractal dimension and overall front coverage as GenTO.

The hypervolume statistics in Supplementary Table~\ref{tab:SI_P5_T2} give the same message in numerical form. Although GA reaches a strong coarse-feature end of the front, its overall normalized hypervolume remains much lower than that of GenTO. The final GenTO Pareto set contains both larger-feature and higher-fractal solutions without collapsing toward only one end of the trade-off. The benchmark supports the same main claim as Case~1: the benefit of GenTO comes from steering a reusable topology prior toward a better task-specific region, here in the form of an expanded Pareto front in morphology space.

\subsection{Additional Results of Case 3}
\label{sec:si_case3_additional}
\runinhead{Comparison setup.}
Case~3 differs from Cases~1 and 2 in a fundamental way. The objective is no longer property extremization or morphology control. Instead, it is an inverse-design problem in elasticity space. Each candidate topology is evaluated by the homogenized elasticity tensor, from which the effective Poisson's ratios are obtained for visualization. The reported MAE values use the same constitutive-matrix percentage error as the main figure, i.e. lower MAE means better matching. In addition, this benchmark is split into three target classes, namely periodic, four-fold, and eight-fold targets, so the task is to improve performance while adapting the topology prior to different symmetry-specific elasticity targets.

The same numerical references used in the main text are compared here: GenTO, Best in Prior, Best in training sets, and GA. Best in Prior and Best in training sets are static references. GA and GenTO are iterative methods. For the iterative methods, the final comparison below is reported by the best design reached along the saved history instead of by the last step only, since the matching error is not always monotonic during optimization. This point is especially important for the four-fold target, where the best design may appear early and later checkpoints may move away from it before improving again. For numerical designs, the constitutive-matrix error is reported as $\mathrm{MAE}=100\,\overline{\Delta H}_{\mathrm{abs}}/H_{\max}$, where $\overline{\Delta H}_{\mathrm{abs}}$ is the mean absolute difference over all tensor components and $H_{\max}$ is the maximum absolute component of the target tensor. For the experimental GenTO bars in Fig.~\ref{fig:property_targeted}d, the corresponding value is estimated from the measured Poisson's-ratio response.

\IfFileExists{Figures/SI_F15.pdf}{%
\begin{figure}[h]
\centering
\includegraphics{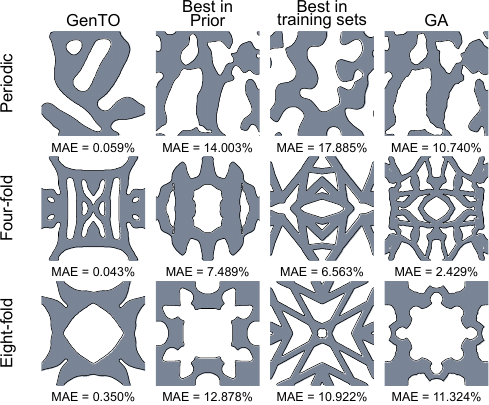}
\caption{Final best-topology comparison for Case~3. Rows correspond to the periodic, four-fold, and eight-fold target tensors, and columns correspond to the compared methods. The values shown under each topology are constitutive-matrix MAE values in percent, so lower values indicate better matching. For GA and GenTO, the reported design is the best one found along the saved optimization history instead of the last state only.}
\label{fig:SI_F15}
\end{figure}
}{}

\begin{table}[h]
\centering
\caption{Quantitative summary of Case~3. Lower constitutive-matrix MAE indicates better agreement with the target response. Values are reported in percent and follow the same error definition as Fig.~\ref{fig:property_targeted}d.}
\label{tab:SI_P5_T3}
\small
\begin{tabular}{lccc}
\toprule
Method & Periodic & Four-fold & Eight-fold \\
\midrule
GenTO                 & 0.059\% & 0.043\% & 0.350\% \\
Best in Prior         & 14.003\% & 7.489\% & 12.878\% \\
Best in training sets & 17.885\% & 6.563\% & 10.922\% \\
GA                    & 10.740\% & 2.429\% & 11.324\% \\
\bottomrule
\end{tabular}
\end{table}

\runinhead{Final comparison across the three target symmetries.}
The final results are summarized in Fig.~\ref{fig:SI_F15} and Table~\ref{tab:SI_P5_T3}. Several clear trends appear. First, the static references already show that the three targets are not equally difficult. For the periodic and eight-fold targets, Best in Prior is slightly better than direct retrieval from the training sets, whereas the four-fold target shows the opposite trend. Thus, the reusable prior alone is helpful but not equally well aligned with all symmetry classes.

GA improves over the two static baselines for the periodic and four-fold targets, showing that fixed-prior population search can already move the designs closer to the target elastic response. However, the eight-fold target remains difficult for GA. By contrast, GenTO gives the best result in all three targets, with MAE values of 0.059\%, 0.043\%, and 0.350\% for the periodic, four-fold, and eight-fold targets, respectively.

The topology patterns in Fig.~\ref{fig:SI_F15} also clarify the nature of this improvement. The static baselines already recover several symmetry-consistent layouts, but they remain limited in how closely their internal member organization matches the target elastic response. GA generates better aligned patterns, especially for the periodic and four-fold cases, yet the resulting layouts are still more irregular and less sharply tuned than those produced by GenTO. The GenTO designs show the clearest emergence of the target-specific structural motifs. In the periodic target, the final pattern becomes highly directional and compliant in a way that matches the target tensor almost exactly. In the four-fold and eight-fold targets, the final designs also show much clearer symmetry-adapted member arrangements than the other methods.

\IfFileExists{Figures/SI_F16.pdf}{%
\begin{figure}[h]
\centering
\includegraphics{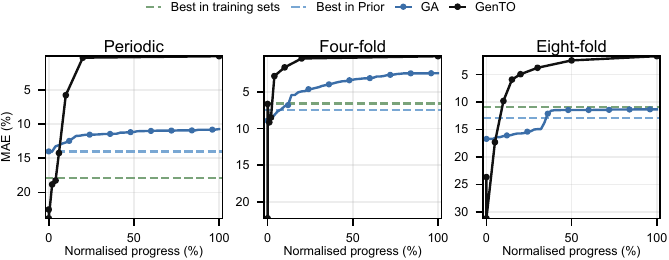}
\caption{History comparison for Case~3. Each panel shows the MAE history for one target symmetry. The horizontal dashed lines denote the Best in training sets and Best in Prior references. The solid curves denote GA and GenTO. Lower MAE indicates better matching.}
\label{fig:SI_F16}
\end{figure}
}{}

\runinhead{Optimization histories of GA and GenTO.}
The optimization histories are shown in Fig.~\ref{fig:SI_F16}. The periodic case is the simplest one to read. GA improves the error gradually over generations and then stabilizes around a relatively narrow range. GenTO decreases the MAE much faster and eventually approaches almost exact tensor matching. The four-fold case is more revealing. GA again improves steadily and reaches a reasonable final error, but GenTO makes a much sharper early move toward the target. This is consistent with the best-over-history result above: once the prior has moved into the right symmetry-specific region, only a few further updates are needed to produce a very accurate match. The eight-fold case is the hardest one. Both iterative methods improve more slowly, and the gap between early and late stages is smaller. Even so, GenTO still maintains a clear advantage over GA throughout the later stages.

These history curves also show a difference from Cases~1 and 2. In the present case, the objective is a target-matching error, so the optimization is not trying to push the design toward one extreme edge of property space. Instead, it is trying to pull the design into a narrow region around the prescribed tensor. The later stages should therefore be read as refinement and realignment, not as simple outward movement. This is why the best checkpoint is not always the last one, especially when the search briefly overshoots or wanders around the target region.

\IfFileExists{Figures/SI_F17.pdf}{%
\begin{figure}[h]
\centering
\includegraphics{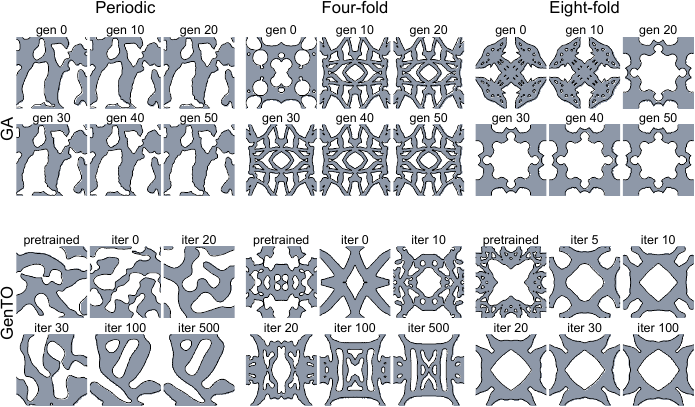}
\caption{Evolution of the best designs in Case~3. The top row shows GA, and the bottom row shows GenTO. Each column corresponds to one target symmetry. Within each block, the displayed sequence shows representative stages from the optimization history.}
\label{fig:SI_F17}
\end{figure}
}{}

\runinhead{Evolution of the best designs.}
The topology histories in Fig.~\ref{fig:SI_F17} make the difference between GA and GenTO more concrete. GA improves through a gradual population-level search. The best design changes step by step, and the main symmetry pattern often becomes visible only after many generations. This behavior is clearest in the periodic target, where the design slowly organizes itself into a more compliant directional layout. The same gradual process also appears in the four-fold and eight-fold targets, although the later changes become smaller as the population begins to stabilize.

GenTO shows a different type of evolution. Since the prior itself is updated, the best design can change more quickly once the checkpoint distribution starts to align with the target tensor. In the periodic and four-fold cases, the early checkpoints already move away from the generic pretrained shapes and toward much more target-specific member organizations. The later checkpoints then mainly refine these structures. In the eight-fold case the progression is slower, but the same pattern is still visible: the design family becomes increasingly symmetry-consistent and mechanically tuned as steering proceeds. Overall, GenTO finds a better final design and reaches the target-specific structural regime more directly than GA.

\runinhead{Discussion of the benchmark results.}
Overall, Case~3 provides a stricter test of reusability than Cases~1 and 2. The goal is no longer to extremize one property or to enlarge a Pareto front, but to hit a prescribed point in elasticity space for three different symmetry classes. The training-set and fixed-prior baselines show that this is not trivial even when the target class is known in advance. GA demonstrates that search in the fixed prior space is useful and can substantially reduce the error, especially for the periodic and four-fold targets. However, its performance remains limited when the target requires a more precise redistribution of structural probability mass.

GenTO gives the clearest evidence of that redistribution. It achieves the best MAE for all three targets, and the gap is especially large in the periodic and four-fold cases. The history plots and topology sequences show that this gain does not come from one lucky late-stage sample. It comes from moving the reusable prior toward a narrower, target-consistent region in elasticity space. From this perspective, Case~3 complements the lessons of Cases~1 and 2. Case~1 showed that steering can move a prior toward different scalar optima. Case~2 showed that steering can expand a Pareto region under competing objectives. Case~3 now shows that the same framework can also support accurate symmetry-specific inverse property matching.

\subsection{Experimental Setup for Case 3}
\label{sec:si_case3_experimental_setup}

This section describes the fabrication and mechanical testing procedure used to experimentally characterize the Case~3 auxetic designs. The experiments were used to obtain the effective Poisson's-ratio response of the printed lattice specimens and to support the comparison shown in Fig.~\ref{fig:property_targeted}d.

\runinhead{Specimen fabrication.}
The negative-Poisson's-ratio specimens were fabricated by stereolithography (SLA) 3D printing using Lasty-702, a green flexural resin supplied by Dongguan Aidi Synthetic Materials Technology Co., Ltd. The printing layer thickness was set to 0.1~mm. Each specimen consisted of a $5\times5$ array of unit cells, with overall dimensions of 100~mm in length, 100~mm in width, and 60~mm in height. The printed specimens of the three models are shown in Supplementary Fig.~\ref{fig:SI_F18}a. The matrix material had a density of 1.18~g~cm$^{-3}$ and a Poisson's ratio of 0.4, as provided by Dongguan Aidi Synthetic Materials Technology Co., Ltd.; its elastic modulus was measured experimentally as 2750~MPa.

\runinhead{Quasi-static compression testing.}
Quasi-static compression tests were performed using an LE5105 electronic universal testing machine manufactured by Lishi (Shanghai) Scientific Instrument Co., Ltd. The tests were conducted under displacement control at a loading rate of 0.5~mm~min$^{-1}$ to obtain the load--displacement responses of the specimens. Before testing, a random speckle pattern was prepared on the specimen surface by spraying a matte white coating followed by a black paint layer, as shown in Supplementary Fig.~\ref{fig:SI_F18}b. During testing, a non-contact three-dimensional full-field strain measurement system (VIC-3D, CSI) was used to continuously acquire speckle images from the specimen surface. The surface full-field strain distribution was measured, and the corresponding displacement and strain fields were obtained by post-processing with the system software.

\runinhead{Poisson's-ratio evaluation.}
To reduce errors caused by local deformation heterogeneity and boundary effects, the central region of each specimen was selected for analysis. This region contained nine unit cells arranged in a $3\times3$ configuration. As illustrated in Supplementary Fig.~\ref{fig:SI_F18}c, measurement points for longitudinal and transverse strains were selected independently. Within the central nine-cell region, six characteristic points were selected for the longitudinal strain, with every two points forming one group and yielding three sets of longitudinal strain data. Another six characteristic points were selected for the transverse strain in the same way, yielding three sets of transverse strain data. The average longitudinal strain $\bar{\varepsilon}_{\mathrm{L}}$ and average transverse strain $\bar{\varepsilon}_{\mathrm{T}}$ were then calculated, and the effective Poisson's ratio $\nu_{\mathrm{eff}}$ of the structure was determined as
\begin{equation}
\nu_{\mathrm{eff}}=-\frac{\bar{\varepsilon}_{\mathrm{T}}}{\bar{\varepsilon}_{\mathrm{L}}}.
\end{equation}

\begin{figure}[htbp!]
\centering
\includegraphics[width=0.99\textwidth]{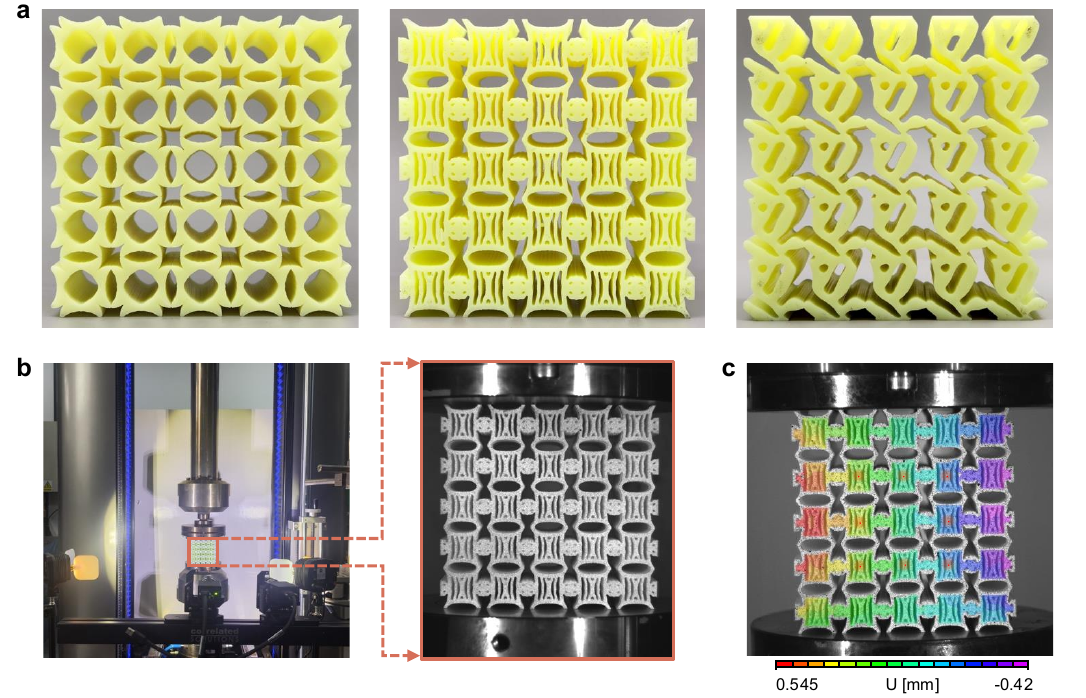}
\caption{\textbf{Experimental characterization of auxetic structures.}
\textbf{a}, SLA-printed auxetic specimens.
\textbf{b}, Quasi-static compression setup with VIC-3D measurement and the speckle-patterned specimen.
\textbf{c}, Representative displacement field and selected central analysis region for strain extraction.}
\label{fig:SI_F18}
\end{figure}

\clearpage
\subsection{Additional Results of Case 4}
\label{sec:si_case4_additional}
\runinhead{Comparison setup.}
Case~4 evaluates function-targeted topology design in the frequency domain. The target is no longer a static property, but a transmission curve that must lie on the prescribed side of a threshold over multiple frequency intervals. The comparison includes GenTO and two static references using the same transmission solver and band-satisfaction score: the best feasible topology found in the 4-fold training sets and the best topology found by sampling the fixed pretrained prior before task-specific steering. This keeps the comparison focused on whether updating the generator distribution improves over both retrieval from existing data and direct use of the unadapted prior.

The two targets are the dual-pass-band configuration, with pass bands at $[1,2]$ and $[4,5]$~kHz and a stop band at $[2,4]$~kHz, and the single-pass-band configuration, with a pass band at $[2,4]$~kHz and stop bands at $[1,2]$ and $[4,5]$~kHz. In both cases, the threshold for classifying a frequency as transmitted or suppressed is $T_{\mathrm{th}}=-10$~dB. The reported score is the weighted band-satisfaction score used in the main text, with weights $0.25$, $0.50$, and $0.25$ for the low-, middle-, and high-frequency regions. For the training-set reference, candidates were screened using the same vertical-support feasibility criterion and a narrow volume-fraction window around the Case~4 setting, and the best candidate was then selected after full frequency-response evaluation. For the fixed-prior reference, the best candidate was selected from the evaluated unsteered prior samples.

\runinhead{Dual-pass-band target.}
Supplementary Fig.~\ref{fig:SI_F19} compares GenTO with the two static references for the dual-pass-band target. The best training-set topology satisfies the low-frequency pass band and the middle stop band, but fails to recover the high-frequency pass band. The best fixed-prior sample shows a similar limitation and does not recover the full dual-pass response. GenTO satisfies all three frequency regions, giving complete agreement with the target band pattern. The additional independent runs show that the same target can also be realized by multiple distinct unit-cell geometries. Some runs retain small mismatches near one pass region, but all strongly suppress the central stop band and achieve substantially higher overall scores than the fixed-prior reference.

\IfFileExists{Figures/SI_F19.pdf}{%
\begin{figure}[htbp!]
\centering
\includegraphics[width=11.74cm]{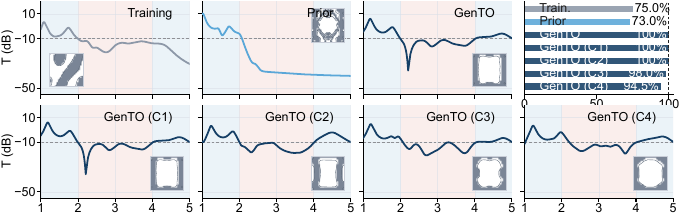}
\caption{\textbf{Static-reference comparison and additional GenTO designs for the dual-pass-band target in Case~4.}
The upper row shows the static-reference comparison. The spectra and inset topologies show, from left to right, the best feasible design in the 4-fold training sets, the best design sampled from the fixed pretrained prior, and the final GenTO design; the rightmost bar chart summarizes the overall weighted band-satisfaction score for these references and the additional GenTO examples. The lower row shows four additional independently optimized GenTO examples, shown from left to right as GenTO (C1)--GenTO (C4). The horizontal axis of each spectrum is frequency in kHz. Blue and red backgrounds mark the target pass and stop regions, respectively, and the dashed line denotes the $-10$~dB threshold.}
\label{fig:SI_F19}
\end{figure}
}{}

\runinhead{Single-pass-band target.}
Supplementary Fig.~\ref{fig:SI_F20} shows the corresponding comparison for the single-pass-band target. The training-set reference and the fixed-prior reference both capture parts of the desired response, but neither consistently satisfies all three frequency regions. In contrast, GenTO reaches complete band satisfaction for the main design. The additional runs are also highly consistent: they transmit the middle frequency band while suppressing the two outer frequency ranges, and their overall scores remain close to one. The successful topologies are not identical. They share the same response class but use different geometric organizations, supporting the interpretation that GenTO steers the prior toward a feasible region of designs instead of simply retrieving one memorized topology.

\IfFileExists{Figures/SI_F20.pdf}{%
\begin{figure}[htbp!]
\centering
\includegraphics[width=11.71cm]{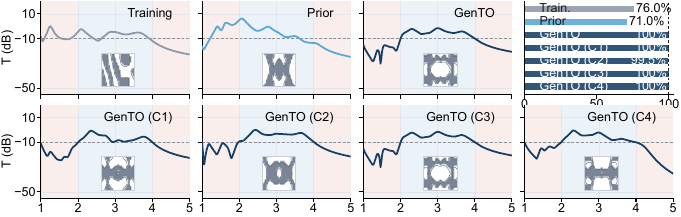}
\caption{\textbf{Static-reference comparison and additional GenTO designs for the single-pass-band target in Case~4.}
The upper row shows the static-reference comparison. The spectra and inset topologies show, from left to right, the best training-set design, the best fixed-prior sample, and the final GenTO design; the rightmost bar chart reports the overall weighted band-satisfaction score for these references and the additional GenTO examples. The lower row shows four additional independently optimized GenTO examples, shown from left to right as GenTO (C1)--GenTO (C4). The horizontal axis of each spectrum is frequency in kHz. Blue and red backgrounds mark the target pass and stop regions, respectively, and the dashed line denotes the $-10$~dB threshold.}
\label{fig:SI_F20}
\end{figure}
}{}

\begin{table}[htbp!]
\centering
\caption{\textbf{Band-satisfaction summary for the Case~4 static references and additional GenTO runs.} Values are percentages of sampled frequencies satisfying the prescribed pass- or stop-band inequality. The overall score is the weighted combination used in the objective, with weights 0.25, 0.50, and 0.25 for the low-, middle-, and high-frequency regions.}
\label{tab:SI_P5_T4}
\scriptsize
\setlength{\tabcolsep}{2.0pt}
\begin{tabular*}{\textwidth}{@{\extracolsep{\fill}}llcccc@{}}
\hline
Target & Example & Low band & Middle band & High band & Overall \\
\hline
Dual pass-band & Best in training sets & 100.0 & 100.0 & 0.0 & 75.0 \\
Dual pass-band & Best in Prior & 100.0 & 96.0 & 0.0 & 73.0 \\
Dual pass-band & GenTO & 100.0 & 100.0 & 100.0 & 100.0 \\
Dual pass-band & GenTO (C1) & 100.0 & 100.0 & 100.0 & 100.0 \\
Dual pass-band & GenTO (C2) & 100.0 & 100.0 & 100.0 & 100.0 \\
Dual pass-band & GenTO (C3) & 100.0 & 97.0 & 98.0 & 98.0 \\
Dual pass-band & GenTO (C4) & 98.0 & 90.0 & 100.0 & 94.5 \\
\hline
Single pass-band & Best in training sets & 32.0 & 86.0 & 100.0 & 76.0 \\
Single pass-band & Best in Prior & 22.0 & 81.0 & 100.0 & 71.0 \\
Single pass-band & GenTO & 100.0 & 100.0 & 100.0 & 100.0 \\
Single pass-band & GenTO (C1) & 100.0 & 100.0 & 100.0 & 100.0 \\
Single pass-band & GenTO (C2) & 98.0 & 100.0 & 100.0 & 99.5 \\
Single pass-band & GenTO (C3) & 100.0 & 100.0 & 100.0 & 100.0 \\
Single pass-band & GenTO (C4) & 100.0 & 100.0 & 100.0 & 100.0 \\
\hline
\end{tabular*}
\end{table}

The dual-pass-band target is generally harder to satisfy than the single-pass-band target. In particular, failures or partial mismatches appear most often in the high-frequency pass band, suggesting that maintaining transmission at higher frequencies while suppressing the middle band imposes a stronger geometric constraint on the unit cell.

\runinhead{Geometry-space PCA analysis.}
The diversity map in the final panel of Fig.~\ref{fig:vibration_transmission} was constructed from geometry descriptors rather than from transmission curves. Each binary unit cell was first converted into an SDF field by subtracting the void-side distance map from the material-side distance map, giving positive values in the material phase and negative values in the void phase. The SDF field was normalized by the image size, downsampled to a compact grid, vectorized, mean-centered, and normalized to unit length. The result is one descriptor vector for each topology,
\[
\mathbf{d}^{(n)}
=
\frac{\operatorname{vec}\!\left(\mathcal{D}(\mathbf{z}^{(n)})\right)-\overline{\operatorname{vec}\!\left(\mathcal{D}(\mathbf{z}^{(n)})\right)}}
{\left\|\operatorname{vec}\!\left(\mathcal{D}(\mathbf{z}^{(n)})\right)-\overline{\operatorname{vec}\!\left(\mathcal{D}(\mathbf{z}^{(n)})\right)}\right\|_2},
\]
where $\mathbf{z}^{(n)}$ is the SDF field of the $n$th topology, $\mathcal{D}(\cdot)$ denotes the downsampling operation, and $\operatorname{vec}(\cdot)$ denotes vectorization. Descriptors were collected from the valid initial samples, the two main optimized designs, and the additional GenTO (C1)--GenTO (C4) designs for both target functions. These descriptors were stacked into a geometry-feature matrix $\mathbf{G}=[(\mathbf{d}^{(1)})^{\mathsf{T}};\ldots;(\mathbf{d}^{(N)})^{\mathsf{T}}]$ and standardized feature-wise to obtain $\widetilde{\mathbf{G}}$. PCA was then computed by applying singular-value decomposition to this standardized descriptor matrix,
\[
\widetilde{\mathbf{G}}=\mathbf{U}\boldsymbol{\Sigma}\mathbf{V}^{\mathsf{T}},\qquad
\mathbf{Y}_{1:2}=\widetilde{\mathbf{G}}\mathbf{V}_{1:2}.
\]
Since $\widetilde{\mathbf{G}}$ contains only SDF-based geometry descriptors, both the principal directions $\mathbf{V}_{1:2}$ and the projected coordinates $\mathbf{Y}_{1:2}$ are purely geometry-based quantities; no transmission curve, target label, or satisfaction score is used in this decomposition. The two columns of $\mathbf{Y}_{1:2}$ were used as the two-dimensional geometry coordinates; the percentages in the axis labels report the explained variance of the corresponding components.

In this map, the grey points and pale ellipse show the initial generated geometry distribution, whereas the colored markers show the optimized designs. The optimized samples are not concentrated in one small region of the SDF geometry space. Instead, high-satisfaction designs for the same target can appear at separated locations, and the two target types occupy partly different regions. The single-pass-band examples are more widely spread, whereas the dual-pass-band examples are more tightly clustered around a similar geometry region. This contrast suggests that the single-pass-band target admits more geometrically distinct high-satisfaction basins, consistent with a more strongly multimodal inverse landscape, while the dual-pass-band target is comparatively more localized. Overall, the map supports the interpretation that Case~4 is a non-convex function-targeted search problem and that GenTO can recover multiple valid geometric basins rather than only refining a single local topology.

\runinhead{Interpretation.}
These supplementary results reinforce the role of Case~4 in the overall study. The objective is non-differentiable, threshold-based, and evaluated through a separate frequency-domain solver, yet the same 4-fold pretrained prior can still be redirected toward high-satisfaction transmission responses. The static references show that neither retrieval from the available training data nor direct sampling from the fixed prior reliably solves the full band-structured objective. GenTO closes this gap by updating the generator itself, and the additional examples show that the resulting success is not restricted to one topology. Case~4 therefore complements the previous static-property and morphology examples by demonstrating that GenTO can also operate as a black-box distribution-steering method for function-targeted spectral design.

\clearpage
\section{Extended Discussions}
\label{sec:si_extended_discussions}

The main text establishes the main empirical claim of GenTO: a pretrained topology prior can be repeatedly adapted to different downstream design tasks through distribution steering. The purpose of the present discussion section is to examine that claim more carefully from four complementary angles. First, we discuss computational efficiency in terms of evaluation budget. Second, we discuss the methodological position of GenTO relative to retrieval, fixed-prior sampling, classical topology optimization, and population-based black-box search. Third, we discuss the modeling choices behind the diffusion prior itself, including the role of SDF representations and the deliberately simple unconditional U-Net used here. Fourth, we reflect on what the four case studies jointly reveal about the scope of the framework. 


\subsection{Efficiency of Distribution Steering}
\label{sec:si_discussion_efficiency}
For the design problems considered here, the main computational cost usually comes from evaluating candidate topologies with task-specific physics models. These evaluations include homogenized thermal or elastic response calculations, geometric-constraint checks, and frequency-domain transmission simulations. Efficiency is therefore discussed using the evaluation budget, defined as the number of candidate topologies submitted to the task-specific objective and constraint evaluators.

To make the comparisons fair, the population-based baselines in this work are run with evaluation budgets of the same order as those used by GenTO (Supplementary Tables~\ref{tab:SI_eval_budget_definition} and \ref{tab:SI_eval_budget}). The Best in training sets and Best in Prior baselines evaluate comparable candidate pools, and the genetic-algorithm baselines use population sizes and generation numbers chosen to give comparable numbers of evaluated candidates. Topology optimization is treated separately because it follows a single design trajectory instead of evaluating an independent candidate population. Under this convention, the comparison tests how effectively each method uses a similar number of physical evaluations instead of which method simply evaluates more candidates.

\begin{table}[htbp!]
\centering
\caption{\textbf{Definition of the evaluation budget for each comparison method.} The evaluation budget counts candidate topologies submitted to the task-specific objective and constraint evaluators. For population-based methods, this count corresponds to the number of evaluated candidates. TO is reported separately because it follows a single trajectory instead of a population of independent candidates.}
\label{tab:SI_eval_budget_definition}
\scriptsize
\setlength{\tabcolsep}{0.8pt}
\begingroup
\renewcommand{\arraystretch}{1.45}
\begin{tabular}{@{}>{\raggedright\arraybackslash}p{2.05cm}>{\raggedright\arraybackslash}p{4.5cm}>{\raggedright\arraybackslash}p{4.25cm}@{}}
\toprule
Method & Budget counted as & Comment \\
\midrule
Best in training sets & Evaluated candidates retrieved from the training datasets & Static retrieval; no update is performed \\
Best in Prior & Evaluated candidates sampled from the fixed pretrained prior & Static generator; no fine-tuning is performed \\
GA & Population size $\times$ generations & Population evolves, but the pretrained prior remains fixed \\
GenTO & Valid candidates evaluated across steering iterations & Evaluated elites are reused to fine-tune the generator \\
TO & Optimization iterations along one design trajectory & Not directly comparable to population-based budgets \\
\bottomrule
\end{tabular}
\endgroup
\end{table}

\begin{table}[htbp!]
\centering
\caption{\textbf{Nominal evaluation budgets used in the four case studies.} The listed values are nominal budgets used to keep population-based comparisons on the same or similar order as GenTO; values are per objective or per target where applicable.}
\label{tab:SI_eval_budget}
\scriptsize
\setlength{\tabcolsep}{1.8pt}
\begin{tabular*}{\textwidth}{@{\extracolsep{\fill}}llllll@{}}
\toprule
Case & GenTO & Training sets & Prior & GA & TO \\
\midrule
Case~1 & $\sim1.3\times10^{4}$ & $\sim1.3\times10^{4}$ & $\sim1.3\times10^{4}$ & $\sim1.3\times10^{4}$ & Single traj. \\
Case~2 & $\sim2.0\times10^{5}$ & $\sim2.0\times10^{5}$ & $\sim2.0\times10^{5}$ & $\sim2.0\times10^{5}$ & Not used \\
Case~3 & $\sim5.1\times10^{4}$ & $\sim5.1\times10^{4}$ & $\sim5.1\times10^{4}$ & $\sim5.1\times10^{4}$ & Not used \\
Case~4 & $\sim5.1\times10^{4}$ & $\sim5.1\times10^{4}$ & $\sim5.1\times10^{4}$ & Not used & Not used \\
\bottomrule
\end{tabular*}
\end{table}

GenTO uses this budget differently from static retrieval, fixed-prior sampling, or genetic search. The evaluated high-performing candidates serve as final designs and are used to update the generator for the next iteration. Later samples are drawn from a progressively more task-adapted distribution, which concentrates physical evaluations in more useful regions of the design space while still preserving multiple candidate solutions.

The distribution update introduces extra computational cost from diffusion sampling and fine-tuning. These costs were not the primary focus of the present study and were not aggressively optimized. In practice, they can be reduced by faster samplers such as DDIM or reduced-step denoising, mixed- or single-precision inference, batched generation, and parallel physics evaluation. The central efficiency argument of GenTO is that a comparable evaluation budget can be turned into a reusable and progressively improved topology generator, even though generation itself is not cost-free.

\subsection{Discussions on the Proposed Method}
\label{sec:si_discussion_method}
Methodologically, GenTO sits between two familiar design paradigms. On one side are classical optimization methods, such as topology optimization, which update a single design field through local sensitivity information. On the other side are generative-design approaches that learn a topology distribution in advance but then often keep that generator fixed, using it either as a conditional inverse model or as a static source of candidate samples. GenTO combines useful aspects of both viewpoints while changing the object of optimization. Like generative design, it starts from a learned structural prior. Like iterative optimization, it repeatedly uses task-specific evaluation to improve performance. However, the update acts on the generator itself, so what evolves is the sampling distribution, not a single design instance.

The distinction is especially important when the design landscape is multimodal, highly non-convex, or only weakly structured by gradients. In such settings, retrieval from a dataset is limited by what is already present, and fixed-prior sampling is limited by the initial concentration of probability mass. Classical topology optimization can still be powerful when sensitivities are reliable, but it typically follows one path and may be sensitive to initialization, regularization, or local traps. A genetic algorithm preserves population diversity more naturally, yet its search still takes place around a fixed representation space unless the generator itself is updated. GenTO addresses these limitations by combining population-level search with representation-level adaptation: high-performing samples are retained and used to reshape the generator that will propose the next population.

Another practical strength of this formulation is its modularity. The same outer loop can be reused once three ingredients are available: a reusable topology prior, a binarization rule compatible with the required material fraction, and a task-dependent evaluator that returns feasibility and score. This makes the framework applicable to differentiable physics objectives as well as thresholded, discontinuous, or experiment-driven objectives for which direct gradient information is unavailable or not trustworthy. From this perspective, GenTO is best understood as a general black-box design engine built on top of a learned structural prior.

The method should not be viewed as uniformly superior to all alternatives in every regime. When the objective is smooth, the feasible set is well behaved, and strong sensitivities are available, classical topology optimization may still remain a highly competitive choice, especially if only one final design is needed. The main comparative advantage of GenTO appears when design knowledge should be reused across tasks, when many high-quality solutions are valuable instead of one isolated optimum, or when the objective structure is too irregular for purely trajectory-based optimization to remain reliable. The results of this work are most naturally interpreted within that methodological niche.

\subsection{Discussions on the Diffusion Model}
\label{sec:si_discussion_diffusion}
The diffusion prior used in this work is intentionally simple, and that simplicity is an important part of the methodological argument. The model is trained on continuous SDF fields instead of on binary images, so the learned representation lives in a smoother geometric space in which interface motion, topology interpolation, and denoising are numerically more stable. This choice is particularly useful for topology data: small geometric shifts produce abrupt changes in binary pixels but much milder changes in the corresponding SDF field. As a result, both prior learning and later task-specific fine-tuning become easier to stabilize. The later sample-wise quantile thresholding step then cleanly restores the prescribed volume fraction for physics evaluation.

The network architecture itself is a lightweight unconditional diffusion U-Net with circular padding and explicit time embeddings. Circular padding is well aligned with periodic unit-cell data because it respects wrap-around continuity at the boundaries. The unconditional formulation is also deliberate. Instead of injecting task labels or target properties into the pretrained model, the prior is learned as a broad reusable description of topology space, and task adaptation is deferred to the steering loop. This separation keeps the offline model reusable and avoids binding the generator too early to a narrow set of predefined conditional interfaces.

Several observations follow from this modeling choice. First, the good performance of the current framework suggests that a very large or heavily conditioned model is not necessary to demonstrate reusable topology steering. Second, the results also indicate that representation choice matters at least as much as network size: the SDF formulation appears to play an important role in enabling smooth migration between topology families and in supporting out-of-distribution generation after steering. Third, the use of symmetry-specific priors is a pragmatic compromise between generality and stability. Training one prior per symmetry class reduces modeling burden and keeps each generative space more coherent, while still allowing reuse across tasks inside that symmetry class.

These advantages come with clear trade-offs. Diffusion sampling remains more expensive than one-shot generation, repeated fine-tuning can narrow the distribution if selection pressure becomes too strong, and the present model does not explicitly control uncertainty or disentangle morphology factors. Future model developments could explore latent diffusion, consistency-style fast samplers, parameter-efficient adaptation, stronger geometry-aware architectures, or hybrid conditional-steering strategies. Nevertheless, the current results suggest that the main conceptual contribution of GenTO does not depend on a highly specialized generator. The reusable-prior idea appears robust to a relatively plain diffusion implementation, which is encouraging for later extensions.

\subsection{Discussions on the Case Studies}
\label{sec:si_discussion_cases}
The four case studies were chosen to probe different meanings of task change within one shared framework. Case~1 changes the optimization direction while keeping the same physics solver and geometric constraints, thereby testing whether the prior can be steered toward two qualitatively different parts of the same conductivity landscape. Case~2 replaces scalar ranking with Pareto-layer selection and asks whether the method can preserve multiple competing morphology directions at once. Case~3 moves the objective from extremization to inverse property matching in elasticity space, and Case~4 shifts further to a frequency-domain function target that is non-differentiable and evaluated through a different numerical pipeline. Together, these tasks span scalar, vector, tensor, and function-valued objectives under one common steering logic.

Seen together, the cases support a more specific interpretation of GenTO. The framework seems most effective because it begins from a topology space that already contains useful geometric regularities and then adapts that space according to the task, not simply because it searches broadly. In Case~1, this produces strong valid solutions even in the irregular minimization landscape. In Case~2, it allows the method to maintain a broader family of non-dominated trade-offs than a fixed-prior search. In Case~3, it supports movement toward narrow target regions in property space, including targets that lie beyond the convex hull of the pretrained samples. In Case~4, it shows that the same mechanism can still function when the score is defined only through black-box spectral evaluation instead of through a conventional static constitutive metric.

The case studies also help clarify what should and should not be inferred from the current results. They do not show that one universal prior can already solve every topology-design problem equally well, nor do they imply that all downstream tasks are equally easy to steer. Rather, they show that a common pretrained topology prior can remain useful across several substantially different design settings, provided that the downstream evaluator and feasibility criteria are formulated carefully. In that sense, the empirical contribution is best read as evidence for reusability across heterogeneous tasks, not as a claim of complete task universality.

\end{document}